# An empirical study of Computing With Words approaches for multi-person and single-person systems

Prashant K Gupta, *Member IEEE*, Pranab Kumar Muhuri, *Member IEEE*

*Abstract-* **Computing with words (CWW) has emerged as a powerful tool for processing the linguistic information, especially the one generated by human beings. Various CWW approaches have emerged since the inception of CWW, such as perceptual computing, extension principle based CWW approach, symbolic method based CWW approach, and 2-tuple based CWW approach. Furthermore, perceptual computing can use interval approach (IA), enhanced interval approach (EIA), or Hao-Mendel approach (HMA), for data processing. There have been numerous works in which HMA was shown to be better at word modelling than EIA, and EIA better than IA. But, a deeper study of these works reveals that HMA captures lesser fuzziness than the EIA or IA. Thus, we feel that EIA is more suited for word modelling in multi-person systems and HMA for single-person systems (as EIA is an improvement over IA). Furthermore, another set of works, compared the performances perceptual computing to the other above said CWW approaches. In all these works, perceptual computing was shown to be better than other CWW approaches. However, none of the works tried to investigate the reason behind this observed better performance of perceptual computing. Also, no comparison has been performed for scenarios where the inputs are differentially weighted. Thus, the aim of this work is to empirically establish that EIA is suitable for multi-person systems and HMA for single-person systems. Another dimension of this work is also to empirically prove that perceptual computing gives better performance than other CWW approaches based on extension principle, symbolic method and 2-tuple especially in scenarios where inputs are differentially weighted.**

## I. Introduction

A key aspect that differentiates computers from the human beings is that computers can process only numeric information. On the other hand, humans can process either numeric or linguistic information, or a combination of these two. Numeric information is precise, whereas linguistic information has an inherent uncertainty. Still, the ability of human cognitive process to tolerate this uncertainty enables them to compute seamlessly using the linguistic information or ‘words’. Zadeh proposed the concept of computing with words (CWW) in 1996 [1], and thus opened up a new frontier for the use of computers to process the linguistic information in a manner similar to the human beings. CWW provides a one-to-one mapping between numeric and linguistic information, and thus in no way means that computers would compute on the linguistic information directly. CWW means that computers would be activated by ‘words’.

Prashant K Gupta and Pranab Kumar Muhuri are with Department of Computer Science, South Asian University, New Delhi, India.
(e-mail: guptaprashant1986@gmail.com, pranabmuhuri@cs.sau.ac.in)

Since the inception of CWW, numerous CWW approaches have been proposed till date. Some of the popular CWW approaches are perceptual computing, extension principle [2] based CWW approach, symbolic method [3] based CWW approach, and 2-tuple linguistic model based CWW approach [4]. The differentiating parameter among all these approaches is the way they represent the uncertainty of linguistic information or word fuzziness. Perceptual computing [6] represents the word uncertainty using the interval type-2 (IT2) fuzzy sets (FSs). The extension principle based CWW approach, and symbolic method based CWW approach represents the word uncertainty using type-1 (T1) FSs [5], and ordinal term sets, respectively. The 2-tuple based CWW approach uses a combination of both the T1 FSs and the ordinal term sets to represent the word uncertainty.

The word uncertainty is generally of two types: intra-person and inter-person. Intra-person uncertainty arises due to the different meanings of the words that same person has over different time instants, whereas the inter-person uncertainty arises due to the different meanings of the words among a group of people. IT2 FSs can capture both intra and inter-person uncertainty. T1 FSs on the other hand can capture only inter-person uncertainty. Thus, utilizing the IT2 FSs for modelling the linguistic information, Mendel proposed the novel CWW approach of perceptual computing.

Perceptual computing has been applied to various fields such as health monitoring of people suffering from heart diseases [8], student strategy evaluation [9], power optimization [14], [15], etc. Perceptual computing uses the interval approach (IA) [10], enhanced interval approach (EIA) [11] or the Hao-Mendel approach (HMA) [12], for the data processing. There have been numerous works which compared the performances of one or more of the IA, EIA and HMA [11-13]. For example, in [11], the authors proposed the EIA and also showed it to give better performance that IA. Authors modified the steps in the data part of EIA by adding more constraints in each step. Also, they modified the height calculation methodology for the lower membership function (LMF) pertaining to interior footprint of uncertainty (FOU), towards the end of FS part.

In [12], authors introduced HMA and compared its performance to the EIA, thereby proving that HMA is better than EIA. The data part of HMA is same as EIA. However, the authors modified the FS part of HMA. In [13], authors presented a thorough comparative analysis of the IA, EIA and HMA. The prime differentiating parameter of HMA in comparison to IA and EIA is that the HMA word FOU models (stored in the codebook) have a height of 1 for both the upper membership function (UMF) and the LMF. On the other hand, for the IA and EIA word FOU models, the UMF is normal, the LMF is generally not normal. Furthermore, a recent work [14] also compared the IA and HMA, in the domain of the power optimization for mobile computing devices. In this work, authors compared the IA and HMA for single-person systems, on the basis of two criteria viz., the generation of unique recommendations and the computational efficiencies. They proved that both the IA and HMA generated same and unique recommendations. Furthermore, they also found that the HMA is faster and computationally more efficient than the IA, though both have the same asymptotic complexity.

The works [8] and [9] involved collecting the data intervals from a group of users for constructing the codebook and thus maybe referred as multi-person systems. On the other hand, [14] involved collecting data intervals form a single user (for codebook construction) and maybe considered as a representation of single-person systems. From the works [8-14], we feel that HMA cannot be universally used as a better word modelling technique than the IA, for both multi and single-person systems. The reason for the same lies in the fact that the FOU plots obtained with HMA are too thin, and therefore capture comparatively very less word fuzziness. The IA FOU plots, on the other hand are too fat (and wide), and thereby capture large amount of word fuzziness. In multi-person systems, opinions of multiple stakeholders assume primacy, thereby making inter-person uncertainty more important than the intra-person uncertainty. However, in single-person systems, the opinion of one person is dominant in the system design, thereby emphasizing the greater importance of intra-person uncertainty in comparison to inter-person uncertainty. Thus, for single-person systems, HMA is more suitable. For multi-person systems, however, EIA, which combines best features of both the IA and HMA, is a better approach.

However, inevitably, IT2 FSs (thus perceptual computing) should be used for word modelling and not the T1 FSs (thus CWW approaches based on them). Mendel also advocated this concept in a number of his works [17], [18]. For example, in the works [8], [9], IA based perceptual computing was shown to give better performance than the CWW approaches based on extension principle, symbolic method and 2-tuple linguistic model, by generating unique recommendations in all the cases (whereas the other CWW approaches failed to do so on multiple occasions). Also, in [7] Herrera and Martinez, established the better performance of 2-tuple based CWW approach over extension principle based CWW approach, and symbolic method based CWW approach, with respect to unique recommendation generation capabilities, using the decision making (DM) scenario pertaining to the purchase of computing systems by an organization. However, it is pertinent to mention that none of these works performed the comparison of CWW approaches, when the inputs are differentially weighted.

Therefore, the objectives of this paper are to present the reader with a holistic comparison of the above said four CWW approaches through an empirical study, as well as establish that EIA is better suited to multi-person systems (compared to IA or HMA), as well as HMA to single-person systems (compared to than the IA or EIA). Furthermore, another dimension of the present work is to prove that perceptual computing is better at word modelling than all three CWW approaches based on extension principle, symbolic method and 2-tuple model, especially when the inputs are differentially weighted.

The organization of the rest of this paper is: Sections 2 and 3 compare the performances of IA, EIA and HMA for multi-person and single-person systems, respectively; Sections 4 and 5 compare the performance of perceptual computing to the extension principle based CWW approach, symbolic method based CWW approach as well as 2-tuple linguistic model based CWW approach for multi-person and single-person systems, respectively; Section 6 discusses the results, and finally Section 7 concludes this paper.

TABLE I

FUZZINESS CAPTURED BY IA, EIA AND HMA AND PERCENTAGE IMPROVEMENTS IN MULTI-PERSON SYSTEMS FOR ASSOCIATED WORDS OF SYSTEM CRITERIA

| Criteria | Associated words of the criteria | Fuzziness | | | | | | | | | % decrease in the mean fuzziness captured by EIA and HMA, compared to IA | |
|---|---|---|---|---|---|---|---|---|---|---|---|---|
| | | IA | | | EIA | | | HMA | | | | |
| | | L[a] | R[b] | M[c] | L[a] | R[b] | M[c] | L[a] | R[b] | M[c] | EIA | HMA |
| Battery (B) life | Very Low (BVL) | 0.09 | 0.61 | 0.35 | 0.00 | 0.46 | 0.23 | 0.00 | 0.00 | 0.00 | 34.29 | 100.00 |
| | Low (BL) | 0.01 | 0.67 | 0.34 | 0.02 | 0.65 | 0.33 | 0.00 | 0.50 | 0.25 | 2.94 | 26.47 |
| | Medium (BM) | 0.00 | 0.83 | 0.41 | 0.00 | 0.81 | 0.40 | 0.05 | 0.62 | 0.33 | 2.44 | 19.51 |
| | High (BH) | 0.03 | 0.66 | 0.34 | 0.04 | 0.64 | 0.34 | 0.00 | 0.56 | 0.28 | 0.00 | 17.65 |
| | Extremely High (BEH) | 0.04 | 0.66 | 0.35 | 0.06 | 0.64 | 0.35 | 0.06 | 0.52 | 0.29 | 0.00 | 17.14 |
| Application (A) Ratings | Very slow (AVS) | 0.04 | 0.66 | 0.35 | 0.09 | 0.61 | 0.35 | 0.00 | 0.63 | 0.32 | 0.00 | 8.57 |
| | Slow (AS) | 0.01 | 0.8 | 0.40 | 0.04 | 0.74 | 0.39 | 0.00 | 0.61 | 0.31 | 2.50 | 22.50 |
| | Moderate (AM) | 0.00 | 0.82 | 0.41 | 0.00 | 0.76 | 0.38 | 0.00 | 0.70 | 0.35 | 7.32 | 14.63 |
| | Fast (AF) | 0.00 | 0.81 | 0.41 | 0.02 | 0.78 | 0.40 | 0.00 | 0.67 | 0.34 | 2.44 | 17.07 |
| | Extremely fast (AEF) | 0.01 | 0.69 | 0.35 | 0.00 | 0.46 | 0.23 | 0.00 | 0.00 | 0.00 | 34.29 | 100.00 |
| Type of application | Absolutely uninteresting (AU) | 0.09 | 0.61 | 0.35 | 0.00 | 0.46 | 0.23 | 0.00 | 0.00 | 0.00 | 34.29 | 100.00 |
| | Somewhat interesting (SI) | 0.00 | 0.82 | 0.41 | 0.01 | 0.80 | 0.41 | 0.00 | 0.71 | 0.35 | 0.00 | 14.63 |
| | Fairly Interesting (FI) | 0.01 | 0.80 | 0.40 | 0.04 | 0.74 | 0.39 | 0.00 | 0.57 | 0.28 | 2.50 | 30.00 |
| | More interesting (MI) | 0.00 | 0.80 | 0.40 | 0.04 | 0.74 | 0.39 | 0.00 | 0.53 | 0.26 | 2.50 | 35.00 |
| | Absolutely interesting (AI) | 0.03 | 0.66 | 0.34 | 0.03 | 0.66 | 0.34 | 0.00 | 0.63 | 0.32 | 0.00 | 5.88 |
| Amount of time spent | Very Little (VL) | 0.09 | 0.61 | 0.35 | 0.00 | 0.84 | 0.42 | 0.00 | 0.45 | 0.22 | -20.00 | 37.14 |
| | Small (S) | 0.01 | 0.80 | 0.40 | 0.00 | 0.49 | 0.25 | 0.00 | 0.45 | 0.22 | 37.50 | 45.00 |
| | Moderate (M) | 0.00 | 0.81 | 0.41 | 0.01 | 0.78 | 0.39 | 0.03 | 0.65 | 0.34 | 4.88 | 17.07 |
| | Large (L) | 0.00 | 0.82 | 0.41 | 0.00 | 0.79 | 0.40 | 0.05 | 0.61 | 0.33 | 2.44 | 19.51 |
| | Very large (VLA) | 0.04 | 0.66 | 0.35 | 0.06 | 0.64 | 0.35 | 0.06 | 0.52 | 0.29 | 0.00 | 17.14 |
| **Average** | | | | | | | | | | | 7.52 | 33.25 |

**[a]Left end point, [b]Right end point, [c]Mean**

TABLE II
FUZZINESS CAPTURED BY IA, EIA AND HMA AND PERCENTAGE IMPROVEMENTS IN SINGLE-PERSON SYSTEMS FOR ASSOCIATED WORDS OF SYSTEM CRITERIA, SATISFACTION RATING AND LINGUISTIC WEIGHTS

| Criteria/ Satisfaction ratings/ Linguistic weights | Associated words of the Criteria/ Satisfaction ratings/ Linguistic weights | Fuzziness | | | | | | | | | % decrease in the mean fuzziness captured by EIA and HMA, compared to IA | |
|---|---|---|---|---|---|---|---|---|---|---|---|---|
| | | IA | | | EIA | | | HMA | | | | |
| | | L[a] | R[b] | M[c] | L[a] | R[b] | M[c] | L[a] | R[b] | M[c] | EIA | HMA |
| Battery (B) life | Very Low (BVL) | 0.12 | 0.55 | 0.33 | 0.12 | 0.55 | 0.33 | 0.05 | 0.18 | 0.11 | 0.00 | 66.67 |
| | Low (BL) | 0.00 | 0.81 | 0.41 | 0.00 | 0.81 | 0.41 | 0.14 | 0.55 | 0.34 | 0.00 | 17.07 |
| | Medium (BM) | 0.00 | 0.83 | 0.42 | 0.00 | 0.83 | 0.42 | 0.12 | 0.61 | 0.36 | 0.00 | 14.29 |
| | High (BH) | 0.00 | 0.82 | 0.41 | 0.00 | 0.82 | 0.41 | 0.16 | 0.56 | 0.36 | 0.00 | 12.2 |
| | Extremely High (BEH) | 0.00 | 0.83 | 0.41 | 0.00 | 0.83 | 0.41 | 0.09 | 0.38 | 0.23 | 0.00 | 43.9 |
| Application (A) Ratings | Very slow (AVS) | 0.04 | 0.76 | 0.4 | 0.04 | 0.75 | 0.39 | 0.07 | 0.28 | 0.18 | 2.50 | 55.00 |
| | Slow (AS) | 0.06 | 0.73 | 0.39 | 0.06 | 0.72 | 0.39 | 0.12 | 0.29 | 0.2 | 0.00 | 48.72 |
| | Moderate (AM) | 0.02 | 0.78 | 0.4 | 0.02 | 0.77 | 0.40 | 0.12 | 0.45 | 0.29 | 0.00 | 27.50 |
| | Fast (AF) | 0.00 | 0.80 | 0.4 | 0.00 | 0.8 | 0.40 | 0.19 | 0.51 | 0.35 | 0.00 | 12.50 |
| | Extremely fast (AEF) | 0.00 | 0.82 | 0.41 | 0.00 | 0.82 | 0.41 | 0.09 | 0.40 | 0.24 | 0.00 | 41.46 |
| Type of application | Absolutely uninteresting (AU) | 0.00 | 0.83 | 0.42 | 0.00 | 0.83 | 0.42 | 0.1 | 0.38 | 0.24 | 0.00 | 42.86 |
| | Somewhat interesting (SI) | 0.00 | 0.82 | 0.41 | 0.00 | 0.82 | 0.41 | 0.14 | 0.58 | 0.36 | 0.00 | 12.20 |
| | Fairly Interesting (FI) | 0.00 | 0.82 | 0.41 | 0.00 | 0.82 | 0.41 | 0.15 | 0.57 | 0.36 | 0.00 | 12.20 |
| | More interesting (MI) | 0.00 | 0.82 | 0.41 | 0.00 | 0.82 | 0.41 | 0.16 | 0.55 | 0.36 | 0.00 | 12.20 |
| | Absolutely interesting (AI) | 0.00 | 0.81 | 0.40 | 0.00 | 0.81 | 0.4 | 0.09 | 0.37 | 0.23 | 0.00 | 42.50 |
| Amount of time spent | Very Little (VL) | 0.00 | 0.82 | 0.41 | 0.00 | 0.82 | 0.41 | 0.06 | 0.29 | 0.18 | 0.00 | 56.10 |
| | Small (S) | 0.00 | 0.82 | 0.41 | 0.00 | 0.82 | 0.41 | 0.13 | 0.58 | 0.36 | 0.00 | 12.20 |
| | Moderate (M) | 0.00 | 0.82 | 0.41 | 0.00 | 0.81 | 0.4 | 0.12 | 0.6 | 0.36 | 2.44 | 12.20 |
| | Large (L) | 0.00 | 0.83 | 0.41 | 0.00 | 0.83 | 0.41 | 0.15 | 0.57 | 0.36 | 0.00 | 12.20 |
| | Very large (VLA) | 0.00 | 0.81 | 0.4 | 0.00 | 0.81 | 0.40 | 0.08 | 0.24 | 0.16 | 0.00 | 60.00 |
| Satisfaction Ratings | Not Satisfied (NS) | 0.00 | 0.83 | 0.42 | 0.00 | 0.83 | 0.42 | 0.09 | 0.38 | 0.23 | 0.00 | 45.24 |
| | Somehow Satisfied (SOS) | 0.00 | 0.82 | 0.41 | 0.00 | 0.82 | 0.41 | 0.17 | 0.55 | 0.36 | 0.00 | 12.20 |
| | Satisfied (SS) | 0.00 | 0.82 | 0.41 | 0.00 | 0.82 | 0.41 | 0.23 | 0.50 | 0.36 | 0.00 | 12.20 |
| | Very Satisfied (VS) | 0.00 | 0.82 | 0.41 | 0.00 | 0.81 | 0.41 | 0.17 | 0.55 | 0.36 | 0.00 | 12.20 |
| | Overly Satisfied (OS) | 0.00 | 0.82 | 0.41 | 0.00 | 0.82 | 0.41 | 0.1 | 0.36 | 0.23 | 0.00 | 43.90 |
| Linguistic Weight | Unimportant (U) | 0.12 | 0.55 | 0.34 | 0.12 | 0.55 | 0.34 | 0.05 | 0.18 | 0.12 | 0.00 | 64.71 |
| | More or Less Unimportant | 0.00 | 0.81 | 0.4 | 0.00 | 0.81 | 0.4 | 0.13 | 0.55 | 0.34 | 0.00 | 15.00 |
| | Important (I) | 0.00 | 0.81 | 0.4 | 0.00 | 0.81 | 0.4 | 0.16 | 0.57 | 0.37 | 0.00 | 7.50 |
| | More or Less Important (MLI) | 0.00 | 0.82 | 0.41 | 0.00 | 0.82 | 0.41 | 0.19 | 0.54 | 0.37 | 0.00 | 9.76 |
| | Very Important (VI) | 0.00 | 0.82 | 0.41 | 0.00 | 0.82 | 0.41 | 0.11 | 0.34 | 0.23 | 0.00 | 43.90 |
| **Average** | | | | | | | | | | | 0.16 | 29.35 |

**[a]Left end point, [b]Right end point, [c]Mean**

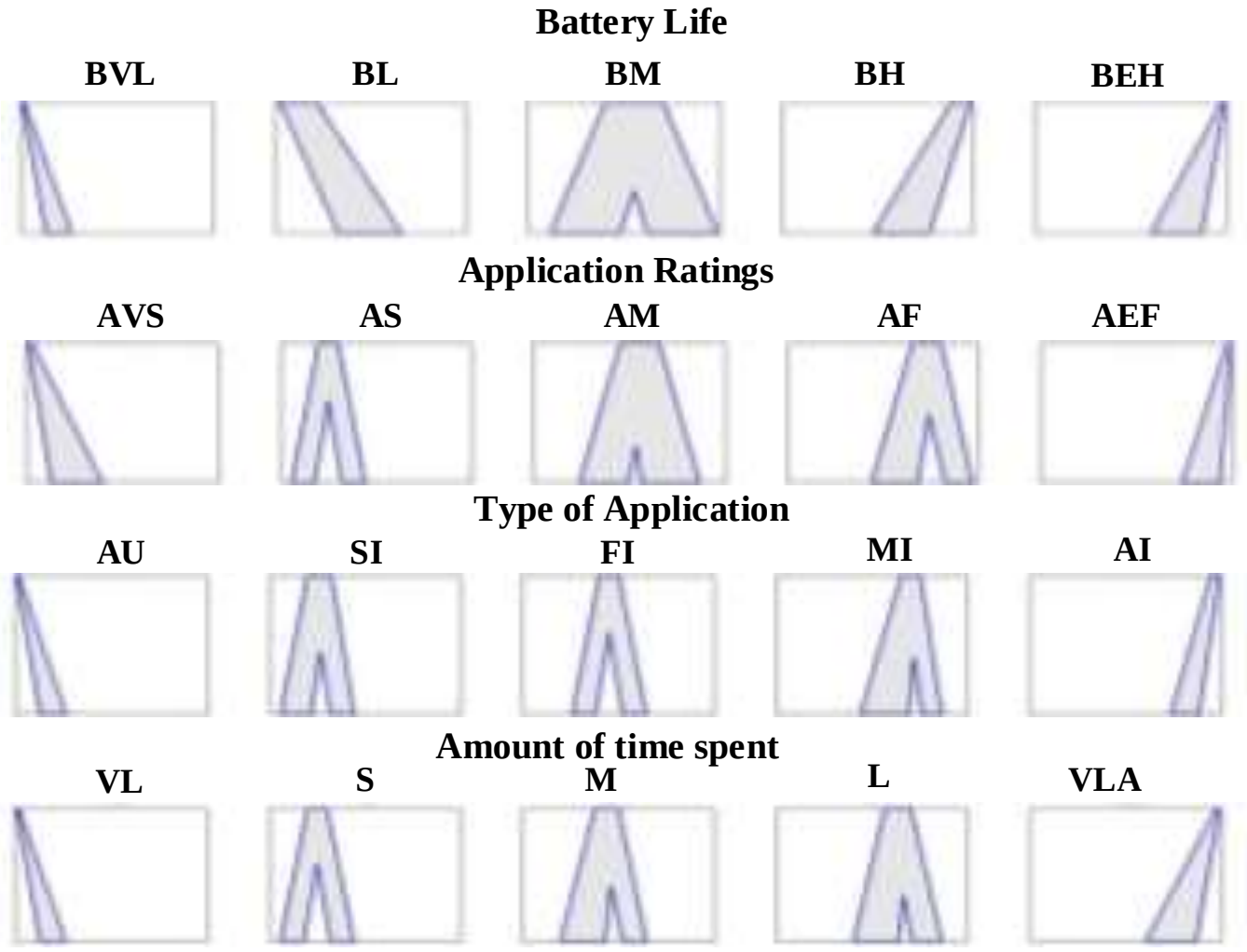

Fig. 1 IA FOU plots for associated words of system criteria for group of users [15].

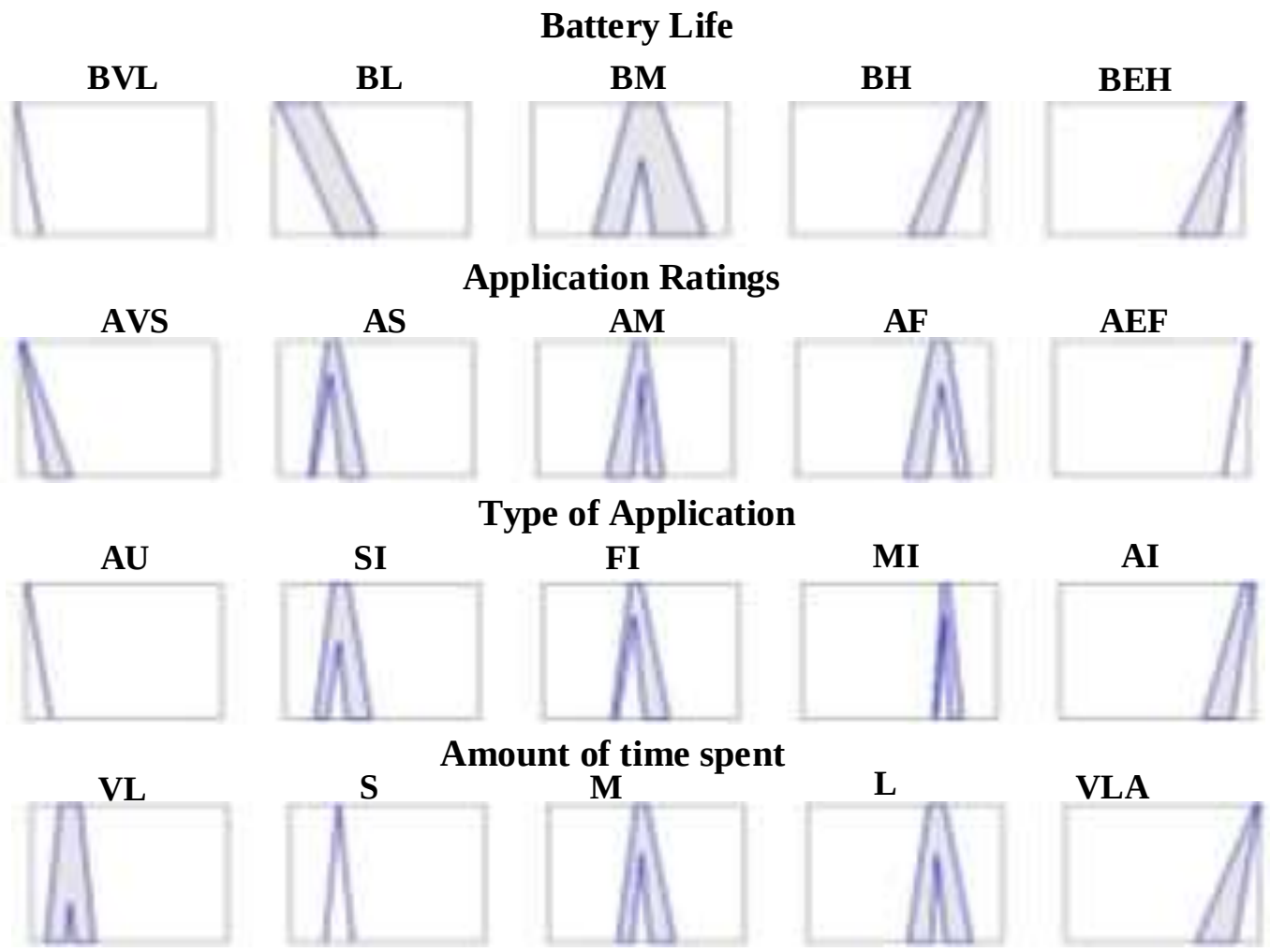

Fig. 2 EIA FOU plots for associated words of system criteria for group of users [15].

## II. MULTI-PERSON SYSTEMS: COMPARISON OF IA, EIA AND HMA PERCEPTUAL COMPUTING TECHNIQUES

In this section, we compare the performances of IA, EIA and HMA for multi-person systems using the data from [15], as a case study. Here, the authors collected data intervals from a group of users to construct the codebook (for various system criteria in order to design a power management policy for battery operated devices). We have designed the codebook by processing the collected data intervals through IA, EIA and HMA. The resulting FOU plots are shown in Fig. 1, Fig. 2 and 3, respectively. Furthermore, the corresponding FOU data are given in the Section SM.I of supplementary materials (SM).

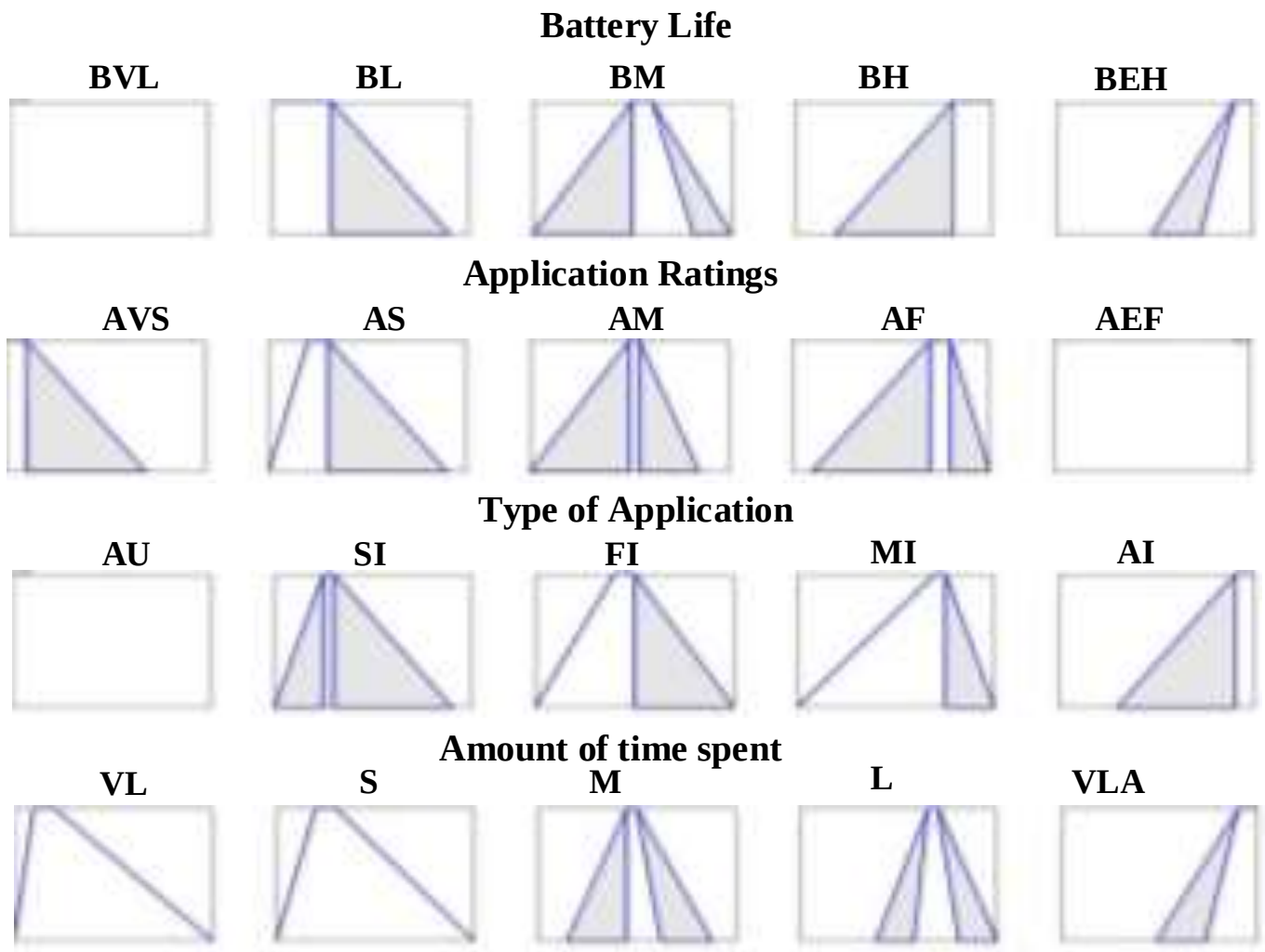

Fig. 3 HMA FOU plots for associated words of system criteria for group of users [15].

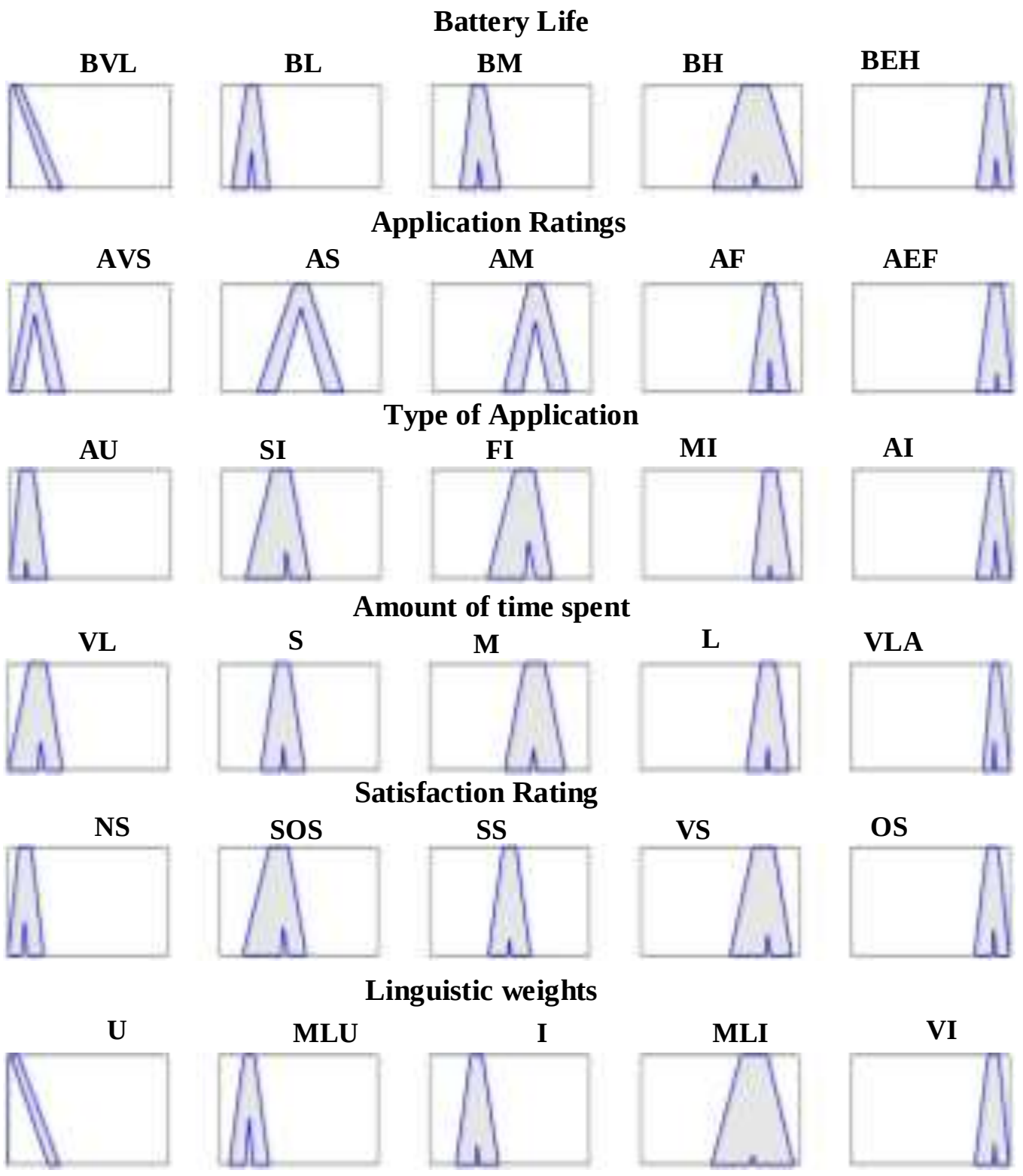

Fig. 4 IA FOU plots for associated words of system criteria, satisfaction ratings and linguistic weights for a user data from [14].

For the FOU plots of Figs. 1, 2 and 3, we calculate the corresponding fuzziness values captured by IA, EIA and HMA (for all the codebook words), and give them in Table I. From Table I, we see that EIA and IA capture almost the comparable amounts of fuzziness, whereas HMA and IA do not. Detailed discussions on the same are given in Section VI.

## III. Single-Person Systems: Comparison of IA, EIA and HMA Perceptual Computing Techniques

In this section, we compare the performances of IA, EIA and HMA for single-person systems using the sample problem of [14], as a case study. Here, the authors asked a single user to provide an interval for the left and right end point, instead of a single precise number, for each word. Then assuming uniform distribution, 50 random numbers are generated in each of the left as well as right interval as: $(L_1, L_2, \ldots, L_{50})$ and $(R_1, R_2, \ldots, R_{50})$, respectively. Then 50 pairs are formed as $(L_i, R_i)$, $i = 1\ to\ 50$, so that each pair becomes a data interval, provided by $i^{th}$ virtual subject. Then using these data intervals, codebook is generated using IA, EIA or HMA. This is also called Person FOU approach.

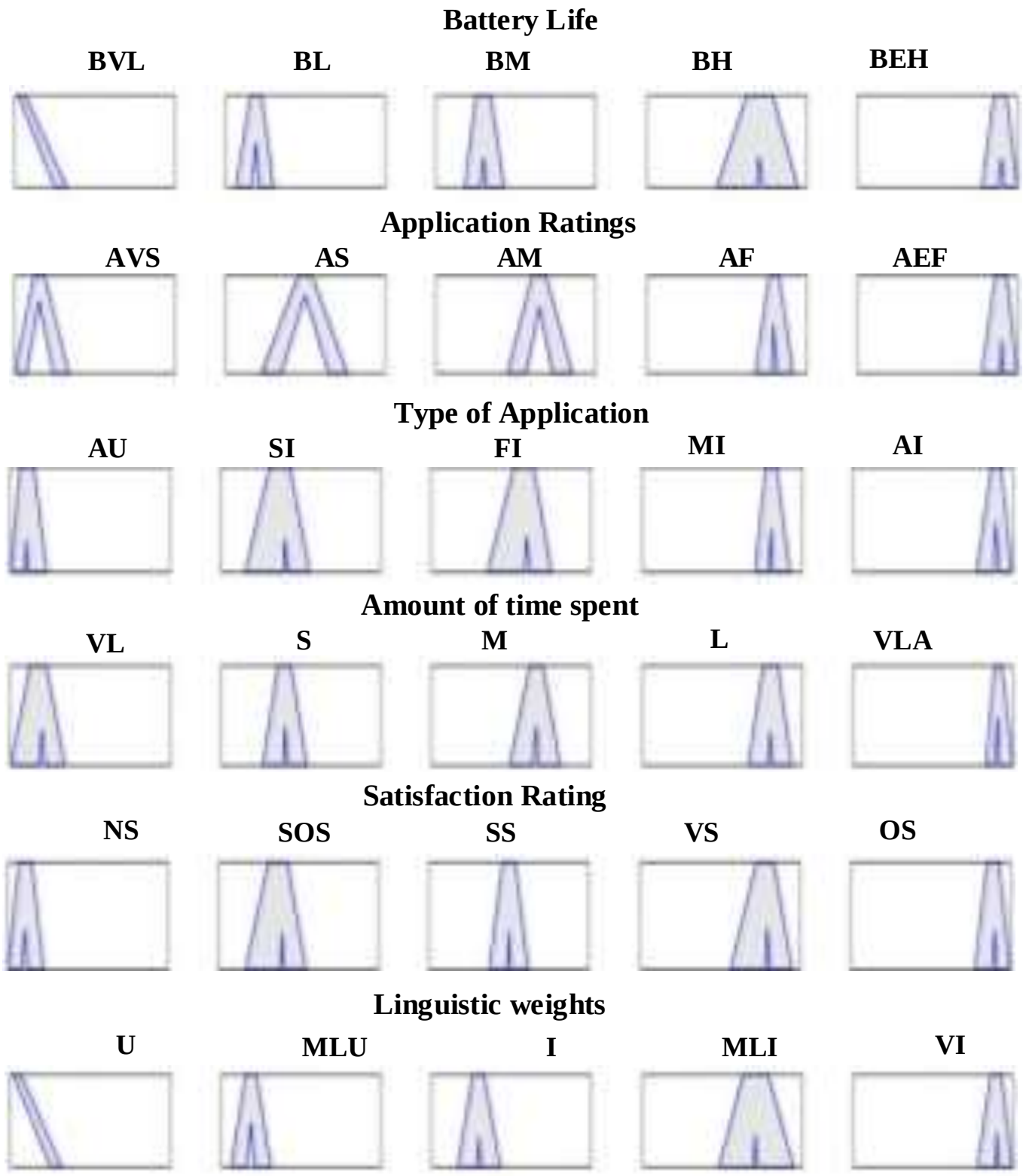

Fig. 5 EIA FOU plots for associated words of system criteria, satisfaction ratings and linguistic weights for a user data from [14].

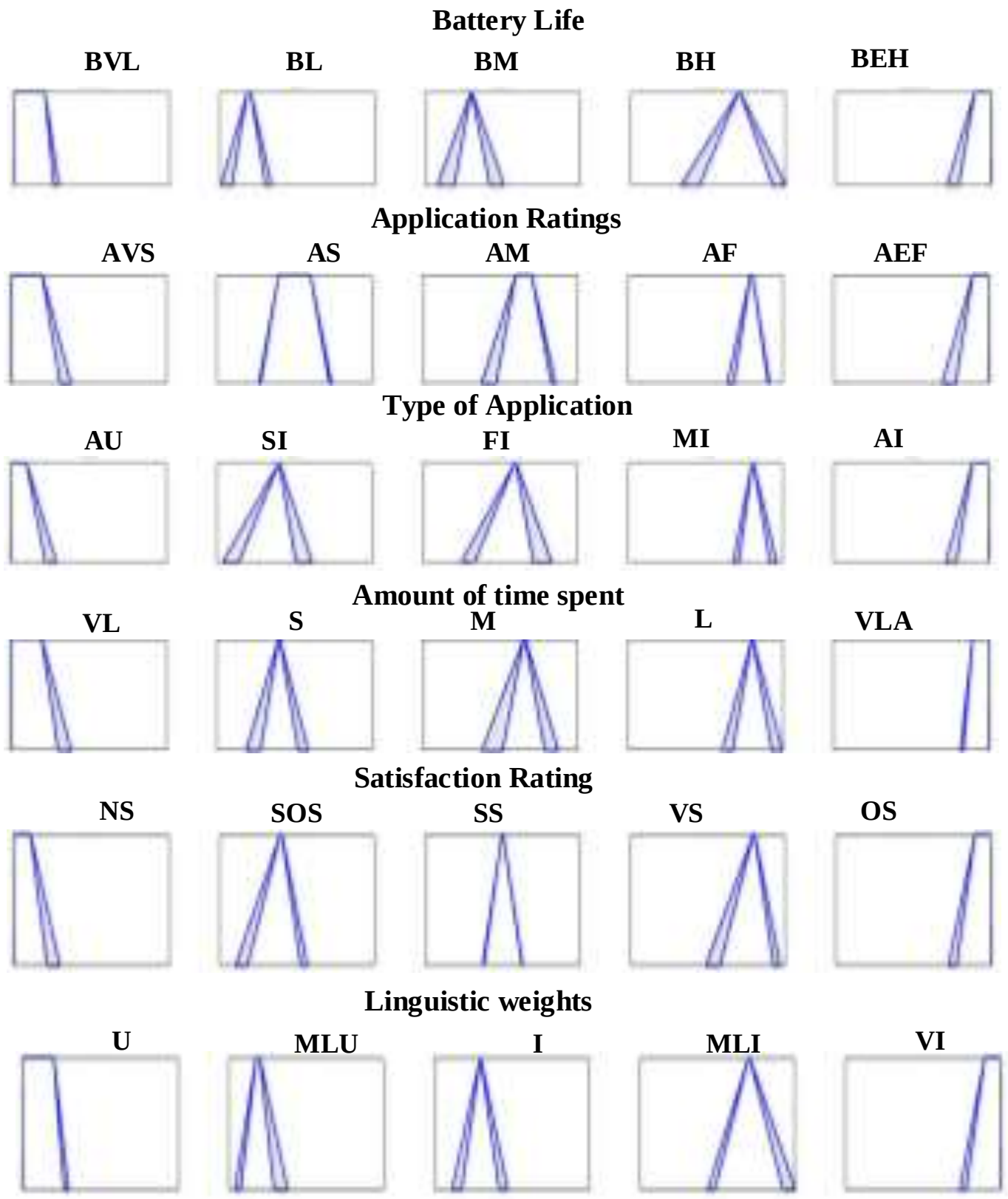

Fig. 6 HMA FOU plots for associated words of system criteria, satisfaction ratings and linguistic weights for a user data from [14].

We provide the FOU plots for the codebook, generated using IA, EIA and HMA, using the data intervals of one user from [14], in Fig. 4, 5 and 6, respectively. The corresponding FOU data are given in the Section SM.II of SM. In the Table II, we have given the data values, corresponding to the fuzziness captured by IA, EIA and HMA for all the codebook words. Detailed discussions on the same are given in Section VI.

## IV. Multi- Person Systems: Comparison of Perceptual Computing To Other CWW Approaches

In this section, we compare the performances of perceptual computing to other CWW approaches based on extension principle, symbolic method and 2-tuple linguistic model, for generating unique recommendations. We have used the sample problem of [15] as a case study, which involved processing the user feedbacks for various system criteria and generating recommendations about user satisfaction aware processor frequencies. The word models generated by perceptual computing (using IA, EIA or HMA) are shown in Figs. 1, 2 and 3. The word models generated by other CWW approaches are discussed in Sections IV.A to IV.C. Then in Section IV.D, we compare the recommendations generated by all the CWW approaches. It is mentioned here that the feedbacks for all the users for various system criteria, used in [15], are given in Section SM.III of SM.

### *A. Extension principle based CWW approach*

The words corresponding to various system criteria in [15], are given in the second column of Table I. Using the extension principle based CWW approach, these words are represented uniformly on the scale of 0 to 1 as the triangular MFs. This is shown in Fig. 7. Each of these triangular MFs are described by three points (corresponding to left, middle and right end of the triangular MF) as tri-tuples and given in Table III.

For illustrating the process of recommending a user satisfaction aware processor frequency, using extension principle based CWW approach, consider that a user rated battery life as high ($BH$), application rating as fast ($AF$), type of application as fairly interesting ($FI$) and amount of time spent as very little ($VL$), in the $i$th frequency of $p$th phase and $g$th game (for details please see [15]). So, the collective preference vector corresponding to the linguistic terms for various criteria of this user's feedback is given by (1) as:

$$User\ feedback = \{BH, AF, FI, VL\} \quad (1)$$

Using Table III, the tri-tuples corresponding to the linguistic terms of (1) are extracted and listed in (2) as:

$$\{(0.5, 0.75, 1), (0.5, 0.75, 1), (0.25, 0.5, 0.75), (0, 0, 0.25)\} (2)$$

When we process the tri-tuples of (20), we get the collective performance vector for the user as:

$$C = \{l_i, m_i, r_i\} = \{0.3125, 0.5, 0.75\} \quad (3)$$

In Fig. 7, we have shown a quantity called the distance vector, which is used to measure the location of collective performance vector of the frequency $Fi$, given in (3).

Table III
Tri-tuple values for linguistic terms of the criteria

| Criteria | Associated linguistic terms of criteria | Tri-tuples for linguistic terms |
|---|---|---|
| Battery (B) life | Very low (BVL) | {0,0,0.25} |
| | Low (BL) | {0,0.25,0.50} |
| | Medium (BM) | {0.25,0.50,0.75} |
| | High (BH) | {0.50,0.75,1} |
| | Extremely high (BEH) | {0.75,1,1} |
| Application (A) Ratings | Very slow (AVS) | {0,0,0.25} |
| | Slow (AS) | {0,0.25,0.50} |
| | Moderate (AM) | {0.25,0.50,0.75} |
| | Fast (AF) | {0.50,0.75,1} |
| | Extremely fast (AEF) | {0.75,1,1} |
| Type of application | Absolutely uninteresting (AU) | {0,0,0.25} |
| | Somewhat interesting (SI) | {0,0.25,0.50} |
| | Fairly interesting (FI) | {0.25,0.50,0.75} |
| | More interesting (MI) | {0.50,0.75,1} |
| | Absolutely interesting (AI) | {0.75,1,1} |
| Amount of time spent | Very little (VL) | {0,0,0.25} |
| | Small (S) | {0,0.25,0.50} |
| | Moderate (M) | {0.25,0.50,0.75} |
| | Large (L) | {0.50,0.75,1} |
| | Very large (VLA) | {0.75,1,1} |

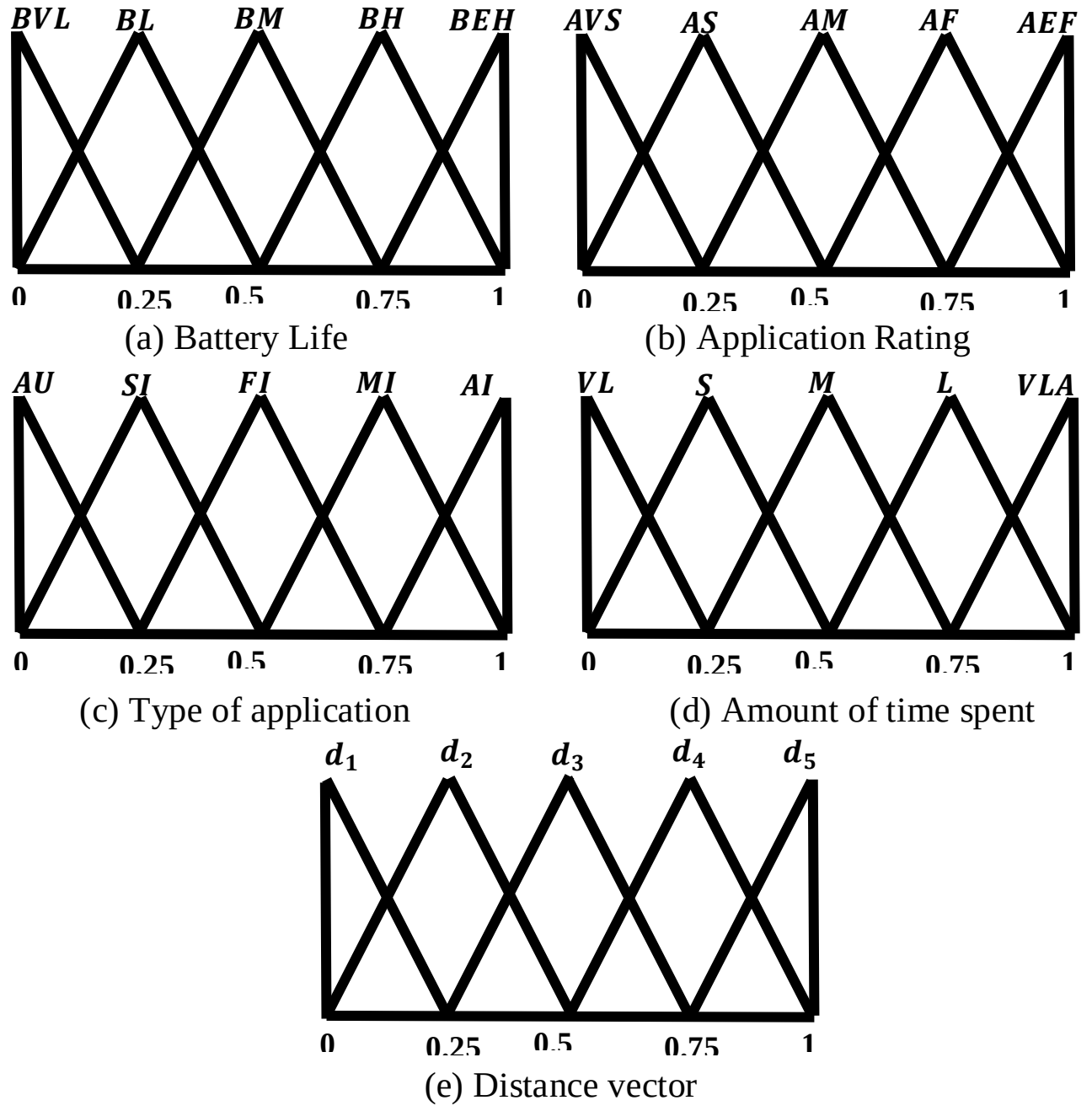

Fig. 7. Triangular MF representation of linguistic terms for criteria and distance vector

Thus, using the concepts of extension principle based CWW approach, each term of the distance vector can be represented in the form of tri-tuple as:

$$d_j = \{l_j, m_j, r_j\}, j = 1, \ldots, 5 \quad (4)$$

In (4), $l_j, m_j$ and $r_j$ are the left, middle and right ends of the triangular MFs, respectively. The distance of the collective performance vector for frequency $Fi$, given in (3), is found from each of the distance term of (4) as:

$$d(Fi, d_j) = \sqrt{P_1(l_i - l_j)^2 + P_2(m_i - m_j)^2 + P_3(r_i - r_j)^2}, \quad j = 1, \ldots, 5 \quad (5)$$

where $P_1, P_2$ and $P_3$ are 0.2, 0.6 and 0.2, respectively. The recommended distance is the one with minimum distance from collective performance vector of the frequency $Fi$. Thus, the values for distances of collective performance vectors corresponding to frequency $Fi$ (given in (3)) from the terms of distance vector shown in Fig. 7 are:

$$\{0.47, 0.26, 0.03, 0.24, 0.45\} \quad (6)$$

Thus, the distance of $d_3$ is minimum. It is therefore closest match to the collective performance vector of (3). Hence, the index recommended for frequency $Fi$ is '3'.

A possible case may arise when multiple distance terms have same distance value from the collective performance vector of frequency $Fi$. In such a scenario, we choose the distance with higher index as the desired recommended distance. The motivation behind such assumption is to maximize the location of frequency's centroid on the information representation scale.

Proceeding similarly, the desired distances for all the six frequencies of training as well as execution phases are found. These are given as:

$$D = \{d(Fi, d_k) | \, F_i \in \{F1, F2, F3, F4, F5, F6\}, d_k \in \{d_1, d_2, d_3, d_4, d_5\}\} \quad (7)$$

Finally, from these six distances given in (7), the recommended frequency is the one with maximum value viz.,

$$F_{reco} = \{Fp | d(Fp, d_k) \geq d(Fi, d_k), i = 1, \ldots, 6, k = 1, \ldots, 5\} \quad (8)$$

However, if two frequencies have the same value of distances, then the lower frequency is the recommended one, as lower the frequency lesser will be the dynamic power dissipation [15].

*B. Symbolic method based CWW approach*

The symbolic method based CWW approach represents the words (corresponding to various system criteria in the second column of Table I) in the form of linguistic term sets, each set corresponding to the a criteria as:

$$Battery\ life: \{s_1: BVL, s_2: BL, s_3: BM, s_4: BH, s_5: BEH\}$$
$$Application\ Ratings: \{s_1: AVS, s_2: AS, s_3: AM, s_4: AF, s_5: AEF\}$$
$$Type\ of\ application: \{s_1: AU, s_2: SI, s_3: FI, s_4: MI, s_5: AI\}$$
$$Amount\ of\ time\ spent: \{s_1: VL, s_2: S, s_3: M, s_4: L, s_5: VLA\} \quad (9)$$

For illustrating the use of symbolic method based CWW approach, for recommending a user satisfaction aware processor frequency, consider again the example of Section IV.A. The user rated battery life as high ($BH$), application rating as ($AF$), type of application as fairly interesting ($FI$) and amount of time spent as very little ($VL$). The indices corresponding to linguistic terms used in feedback of user for various parameters are found from (9), and given in (10) as:

$$User\ feedback = \{s_4: BH, s_4: AF, s_3: FI, s_1: VL\} \quad (10)$$

These linguistic values are first ordered according to the indices of the linguistic terms. Thus, (10) becomes (11) as:

$$User\ feedback = \{s_4, s_4, s_3, s_1\} \quad (11)$$

In [15], all the criteria were equally weighted. Therefore, the weight matrix used in symbolic method based CWW approach is given as:

$$W = \left[w_1 = \frac{1}{4}, w_2 = \frac{1}{4}, w_3 = \frac{1}{4}, w_4 = \frac{1}{4}\right] \quad (12)$$

Thus, for aggregating the user feedback in (11) using the weight matrix of (12), we perform computations as follows:

$$SM^4\left\{\left[\frac{1}{4}, \frac{1}{4}, \frac{1}{4}, \frac{1}{4}\right], [s_4, s_4, s_3, s_1]\right\} = \left(\frac{1}{4} \odot s_4\right) \oplus \left(\frac{3}{4} \odot SM^3\left\{\left[\frac{1}{3}, \frac{1}{3}, \frac{1}{3}\right], [s_4, s_3, s_1]\right\}\right) \quad (13)$$

$$SM^3\left\{\left[\frac{1}{3}, \frac{1}{3}, \frac{1}{3}\right], [s_4, s_3, s_1]\right\} = \left(\frac{1}{3} \odot s_4\right) \oplus \left(\frac{2}{3} \odot SM^2\left\{\left[\frac{1}{2}, \frac{1}{2}\right], [s_3, s_1]\right\}\right) \quad (14)$$

$$SM^2\left\{\left\{\frac{1}{2}, \frac{1}{2}\right\}, \{s_3, s_1\}\right\} = \left(\frac{1}{2} \odot s_3\right) \oplus \left(\frac{1}{2} \odot s_1\right) = s_r \quad (15)$$

In (15), $r = min\left(4, 1 + round(\frac{1}{2} * (3 - 1))\right) = min(4, 2) = 2$. Therefore, the result of (15) is $s_2$. Substituting $s_2$ in (14), we get (16) as:

$$SM^3\left\{\left[\frac{1}{3}, \frac{1}{3}, \frac{1}{3}\right], [s_4, s_3, s_1]\right\} = \left(\frac{1}{3} \odot s_4\right) \oplus \left(\frac{2}{3} \odot s_2\right) = s_r \quad (16)$$

In (16), $r = min\left(4, 2 + round(\frac{1}{3} * (4 - 2))\right) = min(4, 3) = 3$. Therefore, the result of (16) is $s_3$. Substituting $s_3$ in (13), we get (17) as:

$$SM^4\left\{\left[\frac{1}{4}, \frac{1}{4}, \frac{1}{4}, \frac{1}{4}\right], [s_4, s_4, s_3, s_1]\right\} = \left(\frac{1}{4} \odot s_4\right) \oplus \left(\frac{3}{4} \odot s_3\right) = s_r \quad (17)$$

In (17), $r = min\left(4, 3 + round(\frac{1}{4} * (4 - 3))\right) = min(4, 3) = 3$. Therefore, the result of (17) is $s_3$. Therefore, the recommended distance term corresponding to the frequency $Fi$ is $d_3$. Similarly, the recommended distances for all six frequencies of training and execution phases [15] are found. Recommended frequency is the one with maximum recommended distance.

*C. The 2-tuple linguistic model based CWW approach*

We now use the 2-tuple linguistic model based CWW approach for word modelling and processing the feedback of user in example of Section IV.A. Initially, the words (corresponding to various system criteria in the second column of Table I) are represented as linguistic term sets, shown in (9). Using these linguistic term sets, the indices corresponding to the feedback for the criteria of user (from example of Section IV.A) are given in (10). These linguistic values are converted to 2-tuple form by making translation distances 0 for all the terms since each linguistic value is directly drawn from the term set. Therefore, the preference vector becomes:

$$\{(s_4, 0), (s_4, 0), (s_3, 0), (s_1, 0)\} \quad (18)$$

The preference vector of (18) is aggregated as:

$$\beta_{2tp} = \frac{4 + 4 + 3 + 1}{4} = 3 \quad (19)$$

The translation distance is calculated as:

$$\alpha_{2tp} = \beta_{2tp} - round\left(\beta_{2tp}\right) = 3 - round\ (3) = 0 \quad (20)$$

Therefore, the recommended linguistic term is found as:

$$\left(s_{round(\beta_{2tp})}, \alpha_{2tp}\right) = (s_3, 0) = (d_3, 0) \qquad (21)$$

Therefore, the recommended distance term corresponding to the frequency $Fi$ is $d_3$. Similarly, the recommended distance terms for all six frequencies of training and execution phases are found. Recommended frequency is the one with maximum recommended distance.

### *D. Comparison of recommendations generated by Perceptual Computing to other CWW approaches*

Now we compare the performance of perceptual computing against the CWW approaches based on extension principle, symbolic method and 2-tuple linguistic model, pertaining to capability of generating unique recommendations, using the sample problem of [15]. In the experimental setup of [15], authors asked 25 users to play two games viz., Left 4 Dead and Amnesia-the Dark Descent, at variable processor frequencies. The users were asked to provide linguistic feedbacks for the system criteria using the associated words, given in Table I. There were six processor frequencies and two phases (training and execution). The linguistic feedbacks of users were processed to recommend one processor frequency for each phase, game (and the user). The linguistic feedbacks of all the users and games are given in Section SM.III of SM.

The work in [15], used perceptual computing to process the users' linguistic feedbacks and recommend respective processor frequencies, which are listed in Table IV (column 2, 3, 10 and 11). The methodology to process linguistic feedbacks of users, to generate recommendations by CWW approaches based on extension principle, symbolic method and 2-tuple linguistic model, has been discussed in Sections IV.A to IV.C. Thus, the Table IV also lists the recommended frequencies obtained with extension principle based CWW approach, symbolic method based CWW approach and 2-tuple based CWW approach, for both the games viz., Left 4 Dead and Amnesia- the Dark Descent. The frequency values have been bifurcated according to the training (T) and execution (E) phases. Furthermore, it is mentioned here that IA, EIA and HMA, all recommended the same processor frequency.

From Table IV, it can be seen that the CWW approaches based on extension principle, symbolic method and 2-tuple linguistic model fail to generate correct recommendations. For example, consider the data of User 6 in training phase of game Left 4 Dead (See Section SM.II of SM). This user provided following values for the four criteria viz., battery life, application ratings, type of applications and amount of time spent, respectively in frequencies $F1$ and $F4$ as: $\{BH, AF, FI, M\}$ and $\{BH, AM, FI, L\}$, respectively. When we process these values with extension principle based CWW approach, the collective performance vector obtained for both $F1$ and $F4 = \{0.375, 0.625, 0.875\}$. Thus, both end up recommending a distance vector term of $d_3$. When processed with symbolic method based CWW approach, the recommended distance vector term for $F1$ and $F4$ is $d_3$.

Table IV
Recommended frequencies for different CWW approaches

| User No. | Left 4 Dead | | | | | | | | Amnesia- the Dark Descent | | | | | | | |
|---|---|---|---|---|---|---|---|---|---|---|---|---|---|---|---|---|
| | PP[a] | | EP[b] | | SM[c] | | 2-TP[d] | | PP[a] | | EP[b] | | SM[c] | | 2-TP[d] | |
| $T^e/E^f$ → | T | E | T | E | T | E | T | E | T | E | T | E | T | E | T | E |
| 1 | F1 | F6 | F1 | F5 | F1 | F6 | F1 | F6 | F3 | F5 | F1 | F5 | F1 | F5 | F1 | F5 |
| 2 | F2 | F4 | F1 | F4 | F1 | F1 | F2 | F4 | F1 | F1 | F4 | F1 | F4 | F1 | F4 | F1 |
| 3 | F3 | F1 | F3 | F1 | F3 | F1 | F4 | F1 | F6 | F6 | F2 | F1 | F1 | F1 | F2 | F6 |
| 4 | F5 | F6 | F3 | F5 | F1 | F5 | F5 | F6 | F1 | F1 | F4 | F1 | F4 | F1 | F4 | F1 |
| 5 | F5 | F4 | F4 | F4 | F5 | F4 | F5 | F4 | F4 | F1 | F1 | F1 | F1 | F2 | F1 | F2 |
| 6 | F4 | F1 | F1 | F1 | F1 | F1 | F1 | F1 | F1 | F1 | F1 | F1 | F1 | F1 | F6 | F4 |
| 7 | F4 | F6 | F1 | F6 | F2 | F1 | F6 | F6 | F3 | F2 | F1 | F2 | F1 | F2 | F1 | F2 |
| 8 | F3 | F2 | F3 | F1 | F1 | F1 | F3 | F2 | F5 | F1 | F1 | F1 | F1 | F6 | F5 | F6 |
| 9 | F1 | F5 | F1 | F5 | F1 | F5 | F1 | F5 | F3 | F3 | F3 | F3 | F4 | F3 | F4 | F3 |
| 10 | F6 | F1 | F6 | F1 | F1 | F1 | F6 | F1 | F5 | F3 | F1 | F3 | F1 | F5 | F1 | F3 |
| 11 | F1 | F4 | F2 | F3 | F1 | F2 | F2 | F4 | F2 | F5 | F1 | F3 | F1 | F5 | F6 | F3 |
| 12 | F4 | F5 | F3 | F5 | F1 | F5 | F4 | F5 | F2 | F2 | F2 | F1 | F1 | F1 | F2 | F1 |
| 13 | F1 | F3 | F1 | F1 | F1 | F3 | F4 | F5 | F1 | F4 | F1 | F1 | F1 | F1 | F5 | F4 |
| 14 | F1 | F1 | F1 | F1 | F1 | F1 | F1 | F1 | F4 | F2 | F2 | F2 | F2 | F2 | F2 | F2 |
| 15 | F2 | F1 | F1 | F1 | F2 | F1 | F2 | F1 | F1 | F1 | F1 | F1 | F1 | F1 | F1 | F1 |
| 16 | F4 | F4 | F6 | F4 | F1 | F6 | F6 | F6 | F4 | F5 | F6 | F1 | F6 | F1 | F6 | F5 |
| 17 | F3 | F1 | F4 | F1 | F1 | F1 | F4 | F1 | F2 | F1 | F6 | F1 | F1 | F1 | F6 | F1 |
| 18 | F2 | F6 | F2 | F3 | F1 | F6 | F2 | F6 | F1 | F6 | F2 | F1 | F2 | F6 | F3 | F6 |
| 19 | F2 | F5 | F2 | F5 | F2 | F1 | F2 | F5 | F1 | F1 | F4 | F1 | F1 | F5 | F4 | F5 |
| 20 | F3 | F1 | F3 | F1 | F1 | F1 | F3 | F1 | F2 | F3 | F6 | F1 | F6 | F1 | F6 | F3 |
| 21 | F2 | F1 | F2 | F1 | F2 | F1 | F2 | F1 | F1 | F1 | F1 | F1 | F1 | F1 | F1 | F3 |
| 22 | F4 | F1 | F4 | F1 | F1 | F2 | F4 | F1 | F4 | F4 | F4 | F1 | F1 | F1 | F4 | F2 |
| 23 | F3 | F1 | F1 | F1 | F3 | F1 | F3 | F1 | F1 | F1 | F1 | F1 | F1 | F1 | F1 | F1 |
| 24 | F3 | F2 | F3 | F1 | F3 | F1 | F3 | F3 | F1 | F5 | F5 | F5 | F5 | F1 | F6 | F5 |
| 25 | F5 | F6 | F2 | F2 | F1 | F3 | F2 | F3 | F3 | F5 | F6 | F2 | F4 | F1 | F6 | F2 |

[a]Perceptual computing (based on IA, EIA or HMA), [b] Extension principle based CWW approach [c]Symbolic method based CWW approach, [d]2-tuple based CWW approach, [e]Training, [f]Execution

Similarly, when processed with CWW approach based on 2-tuple linguistic model, the recommended distance vector term for both is $(d_4, -0.5)$. However, the mean centroid values obtained for $F1 = 5.99$ and $F4 = 6.30$, when user feedback is processed with perceptual computing. Thus, we can see that perceptual computing can uniquely identify the variations in user 6's feedback for the two processor frequencies. Thus, it can be seen from Table IV, that based on the linguistic feedback of user 6 for various system criteria, the recommended frequency by perceptual computing is $F4$. All the other CWW approaches recommend the frequency as $F1$.

By similar analysis, we can see that CWW approach based on extension principle failed to recommend correct frequency in 48% cases in training phase of game Left 4 dead and 32% cases in execution phase. The corresponding values for CWW approach based on symbolic method are 52% and 40%, whereas for 2-tuple based CWW approach are 24% and 16%. With the game Amnesia the dark descent, CWW approach based on extension principle failed to recommend correct frequency in 68% cases in training phase and 36% cases in execution phase. Corresponding values CWW approach based on symbolic method are 72% and 48%, whereas for 2-tuple linguistic model based CWW approach are 76% and 36%.

## V. Single-Person Systems: Comparison of Person FOU To Other CWW Approaches

In this section, we compare the performances of Person FOU (or perceptual computing for single-person), extension principle based CWW approach, symbolic method based CWW approach, and 2-tuple based CWW approach with the objective of generating unique recommendations, in single-person systems. We have used the sample problem of [14] as a case study, which involved processing the feedback of a single user for various system criteria and generating recommendations about user satisfaction aware processor frequencies. The word models generated by perceptual computing (using IA, EIA or HMA) are shown in Figs. 4, 5 and 6. The word models generated by other CWW approaches are discussed in Sections V.A to V.C. Then in Section V.D, we compare the recommendations generated by all the CWW approaches. It is mentioned here that the feedbacks for all the users for various system criteria, used in [14], are given in Section SM.IV of SM.

*A. Extension principle based CWW approach*

We now use the extension principle based CWW approach for processing the user's linguistic feedback about various system criteria and corresponding linguistic weights (given in Table II). Using these linguistic feedbacks and the weights, we generate recommendations about the user satisfaction aware processor frequency and linguistic satisfaction rating.

*1) Recommended Frequency*

Using the extension principle based CWW approach, the linguistic terms corresponding to the criteria, satisfaction ratings and linguistic weights (from Table II) are represented uniformly on the scale of 0 to 1 as the triangular MFs and are shown in Fig. 8. Each triangular MF is described as a tri-tuple viz., by three points corresponding to left, middle and right end of the triangular MF (Please see Table V).

Consider a user, who rated battery life as high ($BH$), application rating as slow ($AS$), type of application as somewhat interesting ($SI$) and amount of time spent as moderate ($M$), during the training phase of frequency $F1$ in the game Subway surfers. The user assigned a linguistic weight of more or less important ($MLI$) to the battery life, important ($I$) to application ratings, and unimportant ($U$) to each of type of application and amount of time spent. The collective preference vector corresponding to the linguistic terms and weights for various criteria of this user's feedback are given in (22) as:

$$User\ feedback = \{BH, AS, SI, M\}, Weight = \{MLI, I, U, U\} \quad (22)$$

Using Table V, the tri-tuples corresponding to the linguistic terms and weights of (22) are extracted and listed in (23) as:

$$User\ feeback = \{(0.5, 0.75, 1), (0, 0.25, 0.5), (0, 0.25, 0.5), (0.25, 0.5, 0.75)\}$$
$$Weight = \{(0.5, 0.75, 1), (0.25, 0.5, 0.75), (0, 0, 0.25), (0, 0, 0.25)\} \quad (23)$$

Now, we need to perform the weighted aggregation of the linguistic terms corresponding to the criteria as well as linguistic weights. As both these linguistic terms and weights are represented by T1 triangular MFs as shown in Fig. 8, we use the concepts for multiplication of T1 MFs for fuzzy sets (FSs) from [16]. Thus, if we have two T1 MFs for FSs as shown in Fig. 9 with end points as $A = \{a_1, a_2, a_3\}$ and $B = \{b_1, b_2, b_3\}$, then their product is also a T1 triangular MF. The ends of this T1 MF are given by (24) and pictorially shown in Fig. 9 as:

$$A \otimes B = \begin{Bmatrix} min(a_1 \times b_1, a_1 \times b_3, a_3 \times b_1, a_3 \times b_3), \\ a_2 \times b_2, \\ max(a_1 \times b_1, a_1 \times b_3, a_3 \times b_1, a_3 \times b_3) \end{Bmatrix} \quad (24)$$

When we process these tri-tuples of (23) using (24), we get the product vector for the user's feedback in frequency $Fi$ as:

$$C_1 = \begin{Bmatrix} (0.25, 0.5625, 1), (0, 0.125, 0.375), \\ (0, 0, 0.125), (0, 0, 0.1875) \end{Bmatrix} \quad (25)$$

Aggregating the terms of the product vector shown in (25), we get the collective performance vector for the user in frequency $Fi$ as:

$$C = \{l_i, m_i, r_i\} = \{0.06, 0.17, 0.42\} \quad (26)$$

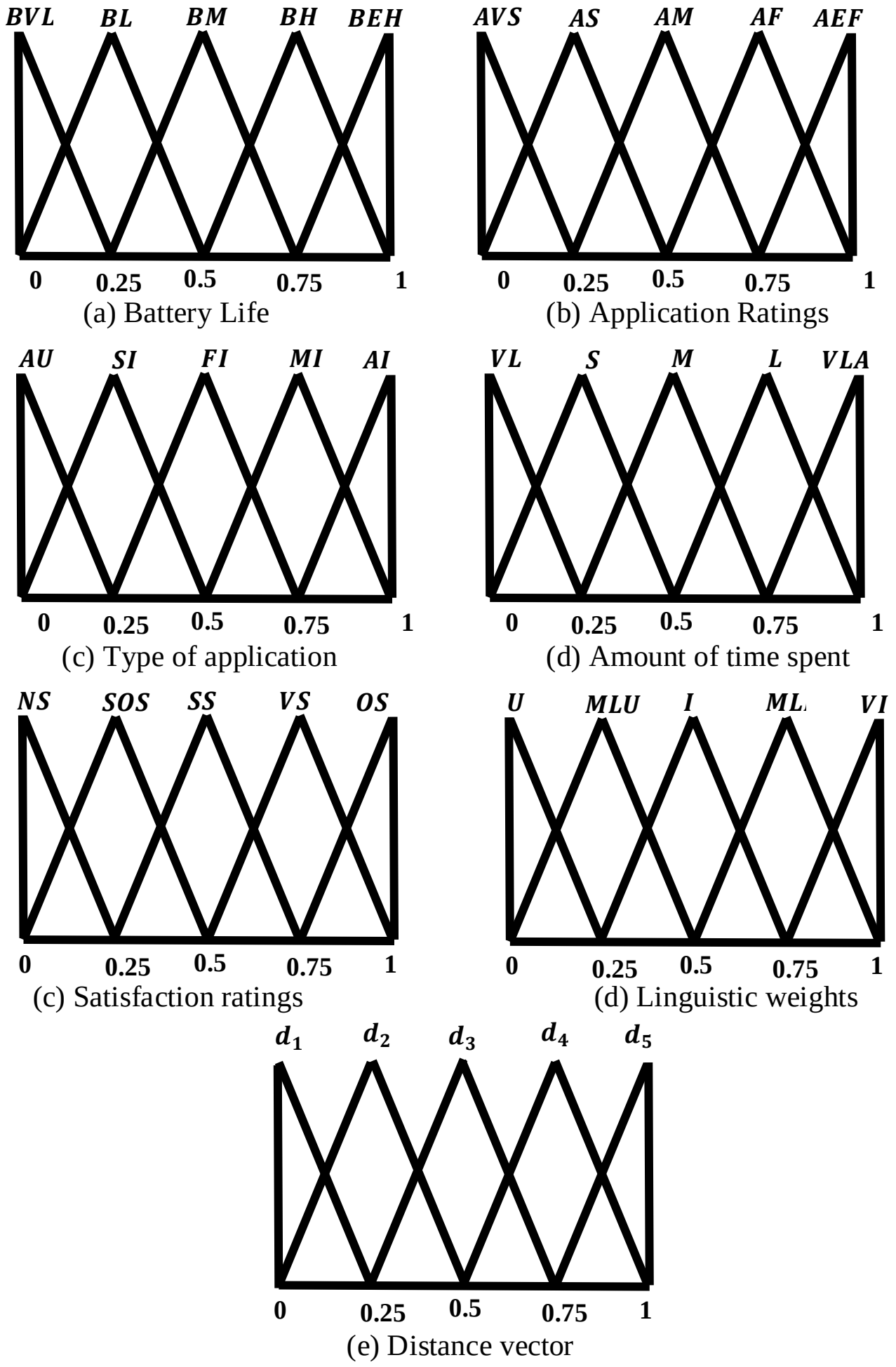

Fig. 8: Triangular MF representation of linguistic terms for criteria, satisfaction ratings, linguistic weights and distance vector

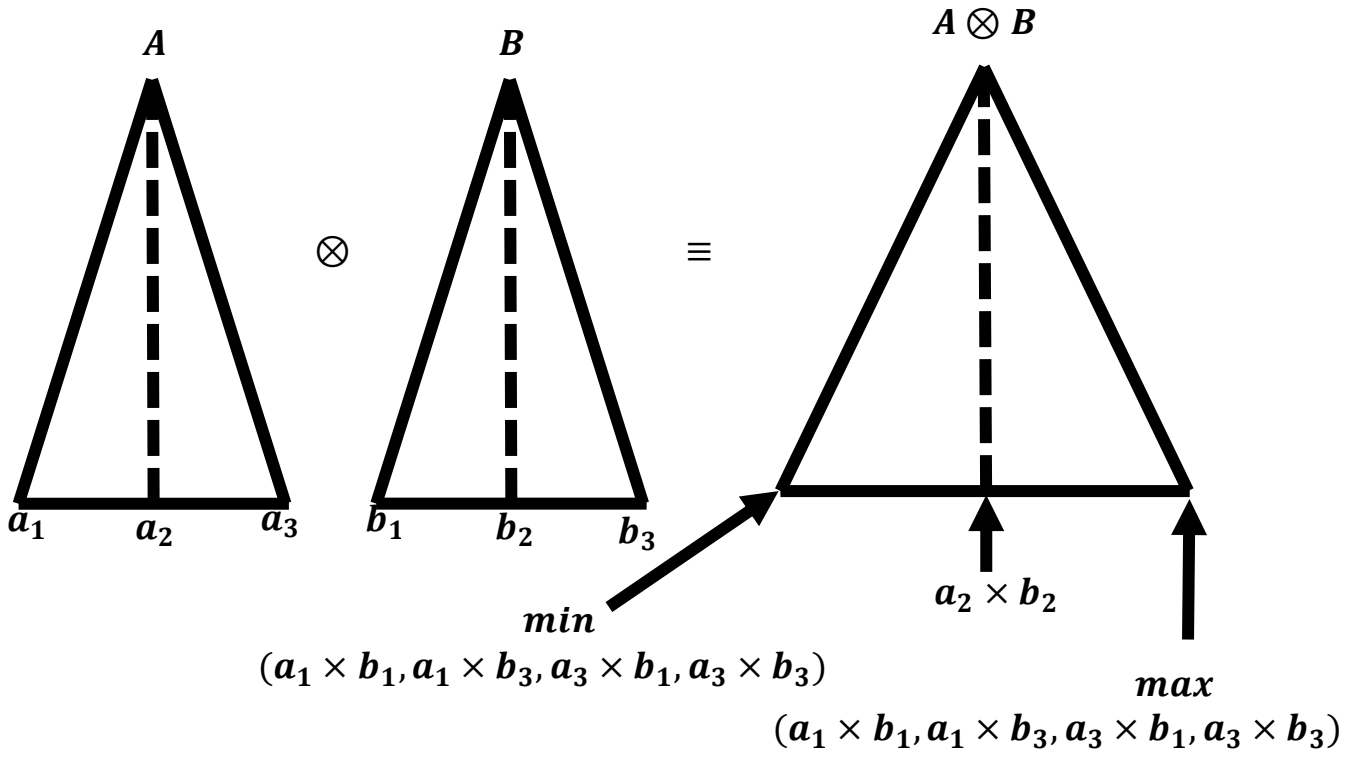

Fig. 9 Product of two triangular type-1 membership functions of fuzzy sets

In Fig. 8, we have shown the quantity called the distance vector, which is used to measure the location of collective performance vector of the frequency $Fi$, given in (26). Using the concepts of extension principle based CWW approach, each term of the distance vector can be represented in the form of tri-tuple as shown in (4). The distance of the collective performance vector for frequency $Fi$, given in (26), is found from each of the distance term of (4) using (5). The recommended distance is the term with minimum distance from collective performance vector of the frequency $Fi$. Thus, the values of distances of collective performance vectors corresponding to frequency $Fi$ (given in (26)) from the terms of distance vector shown in Fig. 8 are:

$$\{0.15, 0.08, 0.31, 0.55, 0.76\} \quad (27)$$

Thus, the distance of $d_2$ is minimum. It is therefore closest match to the collective performance vector of (27). Hence, the index recommended for frequency $F_i$ is '2'.

A possible case may arise when multiple distance terms have same distance value from the collective performance vector of frequency $Fi$. In such a scenario, we choose the distance with higher index as the desired recommended distance. The motivation behind such assumption is to maximize the location of frequency's centroid on the information representation scale. Proceeding similarly, the desired distances for all the six frequencies of training as well as execution phases are found. These are given in (7).

Table V
Tri-tuple values for linguistic terms of the criteria

| Criteria/ Satisfaction Ratings/ | Associated linguistic terms of Criteria/ Satisfaction Ratings/ Linguistic Weights | Tri-tuples for linguistic terms |
|---|---|---|
| Battery (B) life | Very low (BVL) | {0,0,0.25} |
| | Low (BL) | {0,0.25,0.50} |
| | Medium (BM) | {0.25,0.50,0.75} |
| | High (BH) | {0.50,0.75,1} |
| | Extremely high (BEH) | {0.75,1,1} |
| Application (A) ratings | Very slow (AVS) | {0,0,0.25} |
| | Slow (AS) | {0,0.25,0.50} |
| | Moderate (AM) | {0.25,0.50,0.75} |
| | Fast (AF) | {0.50,0.75,1} |
| | Extremely fast (AEF) | {0.75,1,1} |
| Type of application | Absolutely uninteresting (AU) | {0,0,0.25} |
| | Somewhat interesting (SI) | {0,0.25,0.50} |
| | Fairly interesting (FI) | {0.25,0.50,0.75} |
| | More interesting (MI) | {0.50,0.75,1} |
| | Absolutely interesting (AI) | {0.75,1,1} |
| Amount of time spent | Very little (VL) | {0,0,0.25} |
| | Small (S) | {0,0.25,0.50} |
| | Moderate (M) | {0.25,0.50,0.75} |
| | Large (L) | {0.50,0.75,1} |
| | Very large (VLA) | {0.75,1,1} |
| Satisfaction ratings | Not satisfied (NS) | {0,0,0.25} |
| | Somehow satisfied (SOS) | {0,0.25,0.50} |
| | Satisfied (SS) | {0.25,0.50,0.75} |
| | Very satisfied (VS) | {0.50,0.75,1} |
| | Overly satisfied (OS) | {0.75,1,1} |
| Linguistic weights | Unimportant (U) | {0,0,0.25} |
| | More or less unimportant (MLU) | {0,0.25,0.50} |
| | Important (I) | {0.25,0.50,0.75} |
| | More or less important (MLI) | {0.50,0.75,1} |
| | Very important (VI) | {0.75,1,1} |

Finally, from these six distances given in (7), the recommended frequency ($F_{reco}$) is the one with maximum distance value given in (8).

However, if two frequencies have the same value of distances, then the lower frequency is the recommended one, as lower the frequency lesser will be the dynamic power dissipation.

*2) Satisfaction Rating*

For calculating the satisfaction rating corresponding to the recommended frequency ($F_{reco}$) using extension principle based CWW approach, the collective performance vector of the $F_{reco}$ in (8) is compared to the tri-tuples for each of the satisfaction rating's linguistic terms shown in Table V.

Let the collective performance vector of recommended frequency and satisfaction rating is given in tri-tuple form by (28) and (29), respectively as:

$$C_{F_{reco}} = \{l_{reco}, m_{reco}, r_{reco}\} \quad (28)$$

$$s_j = \{l_j, m_j, r_j\}, j = 1, \ldots, 5 \quad (29)$$

In (28), $l_{reco}$, $m_{reco}$ and $r_{reco}$ are the left, middle and right ends of the collective performance vector for recommended frequency represented as T1 MF. (29) give the corresponding values for $j$th satisfaction term. The distances between the quantities given in (28) and (29), are calculated as:

$$d(C_{F_{reco}}, s_j) = \sqrt{P_1(l_{reco} - l_j)^2 + P_2(m_{reco} - m_j)^2 + P_3(r_{reco} - r_j)^2}, \quad j = 1, \ldots, 5 \quad (30)$$

where $P_1, P_2$ and $P_3$ are 0.2, 0.6 and 0.2, respectively. Finally, the recommended satisfaction term is the one with minimum distance, shown in (31) as:

$$s_{reco} = \{s_p | d(C_{F_{reco}}, s_p) \geq d(C_{F_{reco}}, s_k), k = 1, \ldots, 5\} \quad (31)$$

*B. Symbolic method based CWW approach*

We now illustrate the use of symbolic method based CWW approach for processing the user feedback to generate recommendations.

*1) Recommended Frequency*

Initially, the linguistic term sets corresponding to the various criteria, satisfaction ratings and linguistic weights are defined as:

$$Battery\ life: \{s_1: BVL, s_2: BL, s_3: BM, s_4: BH, s_5: BEH\}$$
$$Application\ Ratings: \{s_1: AVS, s_2: AS, s_3: AM, s_4: AF, s_5: AEF\}$$
$$Type\ of\ application: \{s_1: AU, s_2: SI, s_3: FI, s_4: MI, s_5: AI\}$$
$$Amount\ of\ time\ spent: \{s_1: VL, s_2: S, s_3: M, s_4: L, s_5: VLA\}$$
$$Satisfaction\ ratings: \{s_1: NS, s_2: SOS, s_3: SS, s_4: VS, s_5: OS\}$$
$$Linguistic\ weights: \{s_1: U, s_2: MLU, s_3: I, s_4: MLI, s_5: VI\} \quad (32)$$

Consider again the example from Section V.A. The user rated battery life as high ($BH$), application rating as slow ($AS$), type of application as somewhat interesting ($SI$) and amount of time spent as moderate ($M$). The indices corresponding to linguistic terms used in feedback of user for various criteria are given as:

$$User\ feedback = \{s_4: BH, s_2: AS, s_2: SI, s_3: M\} \quad (33)$$

These linguistic values are first ordered according to the indices of the linguistic terms. Thus, (33) becomes (34) as:

Table VI
Recommended frequencies for different CWW approaches

| Game | Subway Surfers | | | | | | | | Asphalt 8: Airborne | | | | | | | | Fruit Ninja | | | | | | | |
|---|---|---|---|---|---|---|---|---|---|---|---|---|---|---|---|---|---|---|---|---|---|---|---|---|
| **User ID** | **PF[a]** | | **EP[b]** | | **SM[c]** | | **2-tuple[d]** | | **PF[a]** | | **EP[b]** | | **SM[c]** | | **2-tuple[d]** | | **PF[a]** | | **EP[b]** | | **SM[c]** | | **2-tuple[d]** | |
| **T[e]/ E[f] →** | **T** | **E** | **T** | **E** | **T** | **E** | **T** | **E** | **T** | **E** | **T** | **E** | **T** | **E** | **T** | **E** | **T** | **E** | **T** | **E** | **T** | **E** | **T** | **E** |
| 1 | F5 | F3 | F5 | F3 | F5 | F1 | F5 | F3 | F6 | F4 | F4 | F2 | F6 | F4 | F6 | F4 | F3 | F3 | F3 | F3 | F3 | F2 | F3 | F3 |
| 2 | F5 | F2 | F5 | F2 | F5 | F2 | F5 | F2 | F5 | F5 | F5 | F4 | F5 | F5 | F5 | F5 | F6 | F5 | F4 | F2 | F5 | F5 | F6 | F5 |
| 3 | F4 | F1 | F1 | F1 | F1 | F1 | F1 | F1 | F6 | F4 | F3 | F2 | F6 | F2 | F6 | F4 | F5 | F2 | F2 | F1 | F3 | F2 | F5 | F2 |
| 4 | F4 | F4 | F5 | F3 | F3 | F3 | F6 | F4 | F6 | F4 | F5 | F3 | F6 | F3 | F6 | F4 | F3 | F3 | F3 | F2 | F5 | F3 | F5 | F3 |
| 5 | F6 | F4 | F5 | F1 | F6 | F1 | F6 | F4 | F1 | F1 | F1 | F1 | F1 | F1 | F1 | F1 | F3 | F3 | F3 | F3 | F3 | F2 | F3 | F3 |
| 6 | F5 | F3 | F5 | F3 | F5 | F1 | F5 | F3 | F6 | F4 | F4 | F2 | F6 | F4 | F6 | F4 | F3 | F3 | F3 | F3 | F3 | F2 | F3 | F3 |
| 7 | F4 | F4 | F5 | F3 | F3 | F3 | F6 | F4 | F6 | F4 | F5 | F3 | F6 | F3 | F6 | F4 | F3 | F3 | F3 | F3 | F5 | F3 | F5 | F3 |
| 8 | F5 | F2 | F5 | F2 | F5 | F2 | F5 | F2 | F5 | F5 | F5 | F4 | F5 | F5 | F5 | F5 | F6 | F5 | F4 | F2 | F5 | F5 | F6 | F5 |
| 9 | F6 | F4 | F5 | F1 | F6 | F1 | F6 | F4 | F1 | F1 | F1 | F1 | F1 | F1 | F1 | F1 | F3 | F3 | F4 | F2 | F5 | F5 | F6 | F5 |
| 10 | F4 | F1 | F1 | F1 | F1 | F1 | F1 | F1 | F6 | F4 | F3 | F2 | F6 | F2 | F6 | F4 | F5 | F2 | F2 | F1 | F3 | F2 | F5 | F2 |
| 11 | F4 | F1 | F1 | F1 | F1 | F1 | F1 | F1 | F6 | F4 | F3 | F2 | F6 | F2 | F6 | F4 | F5 | F2 | F2 | F1 | F3 | F2 | F5 | F2 |
| 12 | F4 | F4 | F5 | F3 | F3 | F3 | F6 | F4 | F6 | F4 | F5 | F3 | F6 | F3 | F6 | F4 | F3 | F3 | F3 | F3 | F5 | F3 | F5 | F3 |
| 13 | F5 | F3 | F5 | F3 | F5 | F1 | F5 | F3 | F6 | F4 | F4 | F2 | F6 | F4 | F6 | F4 | F3 | F3 | F3 | F3 | F3 | F2 | F3 | F3 |
| 14 | F5 | F3 | F2 | F2 | F1 | F1 | F5 | F3 | F6 | F4 | F3 | F3 | F2 | F1 | F6 | F4 | F3 | F3 | F3 | F2 | F1 | F1 | F3 | F3 |
| 15 | F5 | F2 | F3 | F2 | F1 | F1 | F5 | F2 | F5 | F5 | F2 | F2 | F2 | F5 | F5 | F5 | F6 | F5 | F3 | F3 | F1 | F1 | F6 | F5 |
| 16 | F2 | F3 | F2 | F3 | F1 | F1 | F6 | F5 | F1 | F1 | F1 | F1 | F1 | F1 | F1 | F1 | F3 | F2 | F1 | F1 | F5 | F5 | F5 | F2 |
| 17 | F6 | F4 | F4 | F1 | F3 | F1 | F6 | F4 | F6 | F4 | F5 | F1 | F3 | F1 | F6 | F4 | F5 | F3 | F1 | F1 | F1 | F1 | F5 | F3 |
| 18 | F4 | F1 | F1 | F1 | F1 | F3 | F1 | F1 | F6 | F4 | F1 | F2 | F5 | F4 | F6 | F4 | F3 | F2 | F1 | F1 | F5 | F5 | F3 | F2 |
| 19 | F5 | F3 | F2 | F2 | F1 | F1 | F5 | F3 | F6 | F4 | F3 | F3 | F2 | F1 | F6 | F4 | F3 | F3 | F3 | F2 | F1 | F1 | F3 | F3 |
| 20 | F6 | F4 | F4 | F1 | F3 | F1 | F6 | F4 | F6 | F4 | F5 | F1 | F3 | F1 | F6 | F4 | F5 | F3 | F1 | F1 | F1 | F1 | F5 | F3 |
| 21 | F5 | F2 | F3 | F2 | F1 | F1 | F5 | F2 | F5 | F4 | F3 | F4 | F2 | F5 | F5 | F5 | F6 | F5 | F3 | F1 | F1 | F1 | F6 | F5 |
| 22 | F2 | F3 | F1 | F1 | F1 | F1 | F6 | F3 | F1 | F1 | F1 | F1 | F1 | F1 | F1 | F1 | F3 | F2 | F1 | F1 | F1 | F1 | F5 | F3 |
| 23 | F4 | F1 | F1 | F1 | F1 | F3 | F1 | F1 | F6 | F4 | F1 | F2 | F5 | F4 | F6 | F4 | F3 | F2 | F1 | F1 | F5 | F5 | F3 | F2 |
| 24 | F2 | F3 | F1 | F1 | F1 | F2 | F6 | F5 | F1 | F1 | F1 | F1 | F1 | F3 | F1 | F1 | F3 | F2 | F1 | F1 | F1 | F3 | F1 | F1 |
| 25 | F5 | F2 | F3 | F2 | F1 | F1 | F5 | F2 | F5 | F4 | F5 | F4 | F2 | F5 | F4 | F5 | F4 | F5 | F3 | F1 | F1 | F1 | F6 | F5 |

[a]Person FOU (based on IA, EIA or HMA), [b]Extension principle based CWW approach, [c]Symbolic method based CWW approach, [d]2-tuple based CWW approach, [e]Training, [f]Execution

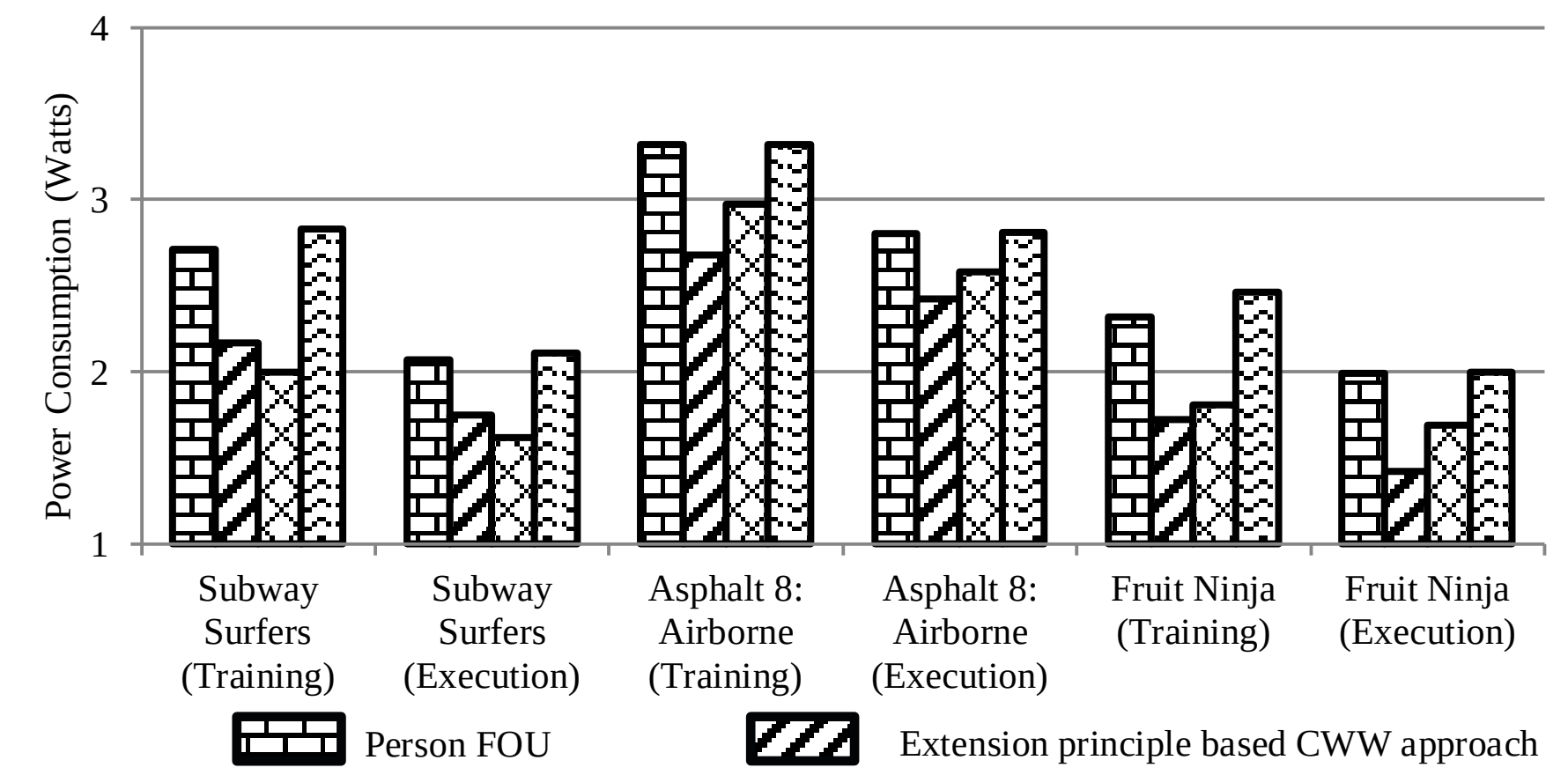

Fig. 10 Average power consumption for the training and execution phases of all the games

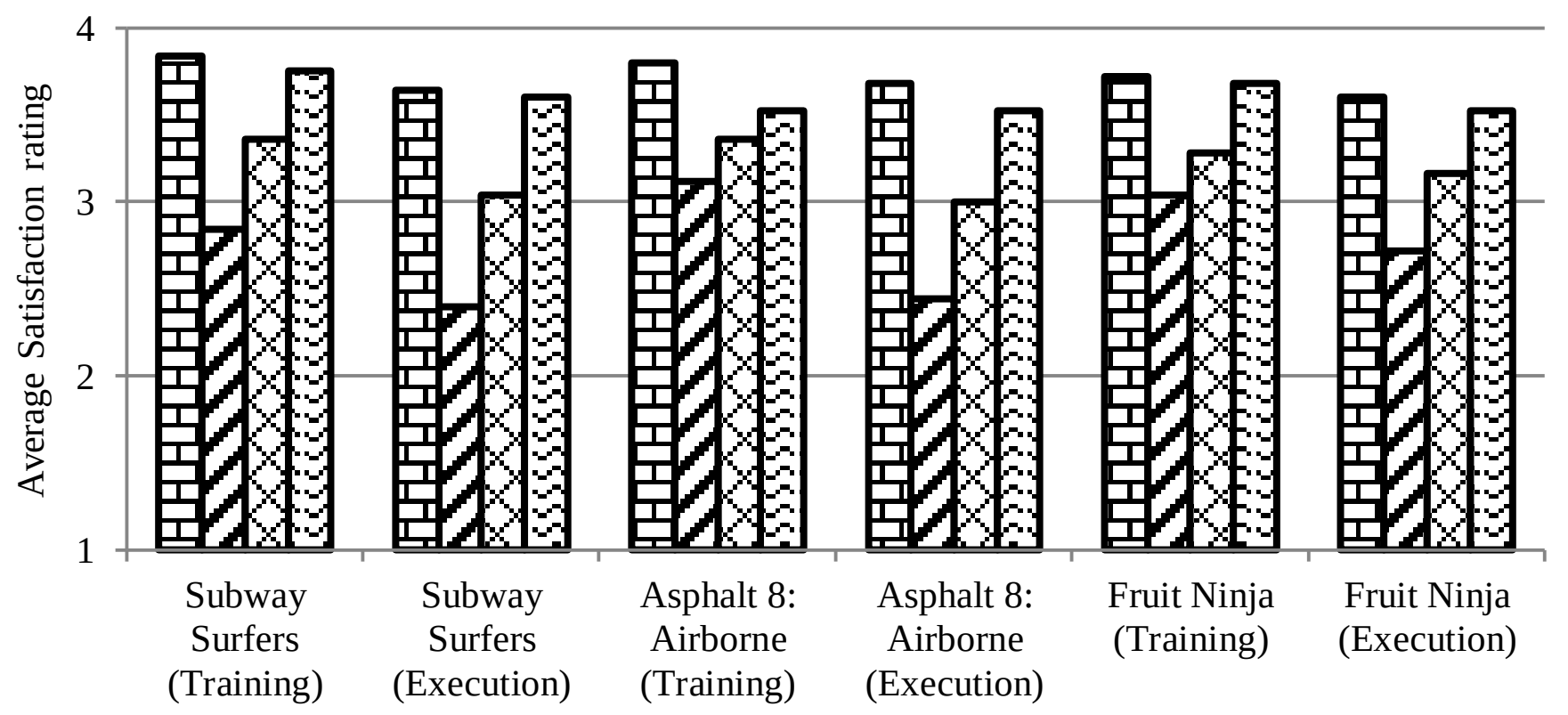

Fig. 11 Average satisfaction rating for training and execution phases of all the games

$$User\ feedback = \{s_4, s_3, s_2, s_2\} \quad (34)$$

The user assigned a linguistic weight of more or less important ($MLI$) to the battery life, important ($I$) to application ratings, and unimportant ($U$) to each of type of application and amount of time spent. Thus, the use the weight matrix as:

$$W = \left[w_1 = \frac{4}{9}, w_2 = \frac{3}{9}, w_3 = \frac{1}{9}, w_4 = \frac{1}{9}\right] \quad (35)$$

For aggregating these values, following computations as performed:

$$SM^4\left\{\left[\frac{4}{9},\frac{3}{9},\frac{1}{9},\frac{1}{9}\right],[s_4,s_3,s_2,s_2]\right\} = \left(\frac{4}{9}\odot s_4\right)\oplus\left(\frac{5}{9}\odot SM^3\left\{\left[\frac{3}{5},\frac{1}{5},\frac{1}{5}\right],[s_3,s_2,s_2]\right\}\right) \quad (36)$$

$$SM^3\left\{\left[\frac{3}{5},\frac{1}{5},\frac{1}{5}\right],[s_3,s_2,s_2]\right\} = \left(\frac{3}{5}\odot s_3\right)\oplus\left(\frac{2}{5}\odot SM^2\left\{\left[\frac{1}{2},\frac{1}{2}\right],[s_2,s_2]\right\}\right) \quad (37)$$

$$SM^2\left\{\left\{\frac{1}{2},\frac{1}{2}\right\},\{s_2,s_2\}\right\} = \left(\frac{1}{2}\odot s_2\right)\oplus\left(\frac{1}{2}\odot s_2\right) = s_r \quad (38)$$

In (38), $r = min\left(4, 2 + round(\frac{1}{2} * (2-2))\right) = min(4,2) = 2$. Therefore, the result of (38) is $s_2$. Substituting $s_2$ in (37), we get (39) as:

$$SM^3\left\{\left[\frac{3}{5},\frac{1}{5},\frac{1}{5}\right],[s_3,s_2,s_2]\right\} = \left(\frac{3}{5}\odot s_3\right)\oplus\left(\frac{2}{5}\odot s_2\right) = s_r \quad (39)$$

In (39), $r = min\left(4, 2 + round(\frac{3}{5} * (3-2))\right) = min(4,3) = 3$. Therefore, the result of (39) is $s_3$. Substituting $s_3$ in (36), we get (40) as:

$$SM^4\left\{\left[\frac{4}{9},\frac{3}{9},\frac{1}{9},\frac{1}{9}\right],[s_4,s_3,s_2,s_2]\right\} = \left(\frac{4}{9}\odot s_4\right)\oplus\left(\frac{5}{9}\odot s_3\right) = s_r (40)$$

In (40), $r = min\left(4, 3 + round(\frac{4}{9} * (4-3))\right) = min(4,3) = 3$. Therefore, the result of (40) is $s_3$. Therefore, the recommended distance term corresponding to the frequency $F_i$ is $d_3$. Similarly, the recommended distances for all six frequencies of training and execution phases are found. Recommended frequency is the one with maximum recommended distance, which is given as:

$$F_{reco} = \{F_p | d_{F_p} \geq d_{F_i}, i = 1, \ldots, 6\} \quad (41)$$

*2) Satisfaction Rating*

We now calculate the satisfaction ratings with the symbolic method based CWW approach. We compare the index of the distance term corresponding to the recommended frequency of (41), to the satisfaction ratings vectors' terms given in (32). The linguistic term from the satisfaction ratings vector which has the same index as that of distance term for recommended frequency, is the corresponding satisfaction rating. This is shown in (42) as:

$$s_{reco} = \{s_k | k = d_{F_{reco}}, k = 1, \ldots, 5\} \quad (42)$$

*C. 2-tuple linguistic model based CWW approach*

Here, we illustrate the use of 2-tuple based CWW approach for processing the user feedback, and generating recommendations.

*1) Recommended Frequency*

Now we will illustrate the use of the 2-tuple based CWW approach for recommending a user-satisfaction aware processor frequency, through the example of Section V.A. The user rated battery life as high ($BH$), application rating as slow ($AS$), type of application as somewhat interesting ($SI$) and amount of time spent as moderate ($M$).

The indices along with various linguistic terms corresponding to the feedback for the criteria of user are given in (33). These linguistic values are converted to 2-tuple form by making translation distances 0 for all the terms since each linguistic value is directly drawn from the term set. Therefore, the preference vector becomes:

$$\{(s_4, 0), (s_2, 0), (s_2, 0), (s_3, 0)\} \quad (43)$$

The user assigned a linguistic weight of more or less important ($MLI$) to the battery life, important ($I$) to application ratings, and unimportant ($U$) to each of type of application and amount of time spent. Thus, the weight matrix is given as:

$$W = [w_1 = 4, w_2 = 3, w_3 = 1, w_4 = 1] \quad (44)$$

Therefore, the user feedback for criteria are aggregated using the weighted average as:

$$\beta_{2tp} = \frac{\sum_{i=1}^{4} I_{s_i} I_{w_i}}{\sum_{i=1}^{4} I_{w_i}} = \frac{(4\times 4) + (2\times 3) + (2\times 1) + (3\times 1)}{(4+3+1+1)} = 3 \quad (45)$$

where $I_{s_i}$ and $I_{w_i}$ are the indices of the satisfaction terms drawn and linguistic weights, from (43) and (44), respectively.

The translation distance is calculated as:

$$\alpha_{2tp} = \beta_{2tp} - round\left(\beta_{2tp}\right) = 3 - round\ (3) = -0 \quad (46)$$

Therefore, the recommended linguistic term is found as:

$$\left(s_{round(\beta_{2tp})}, \alpha_{2tp}\right) = (s_3, 0) = (d_3, 0) \quad (47)$$

Therefore, the recommended distance term corresponding to the frequency $Fi$ is $d_3$. Similarly, the recommended distance terms for all six frequencies of training and execution phases are found. Recommended frequency is the one with maximum recommended distance. This is shown in (41).

*2) Satisfaction Rating*

To calculate the satisfaction ratings with the 2-tuple based CWW approach, we follow the similar approach as the symbolic method based CWW approach. We compare the index of the distance term corresponding to the recommended frequency obtained with 2-tuple based CWW approach (given in (41)), to the satisfaction ratings vectors' terms given in (32).

The linguistic term from the satisfaction ratings vector which has the same index as that of distance term for recommended frequency, is the corresponding satisfaction rating. This is shown in (42).

*D. Comparison of recommendations generated by Person FOU to other CWW approaches*

Now we compare the performance of Person FOU against the CWW approaches based on extension principle, symbolic method and 2-tuple linguistic model, pertaining to capability of generating unique recommendations, using the sample problem of [14]. In the experimental setup of [14], authors processed the linguistic feedbacks of a single user about the perceived system performance for various system criteria, (each criteria with a different linguistic weight), when the user played three games viz., Subway Surfers, Asphalt 8: Airborne, and Fruit Ninja, at variable processor frequencies. There were 25 users in the experimental setup, each of which provided linguistic feedbacks for the system criteria and weights using the associated words, given in Table II. There were six processor frequencies and two phases (training and execution). The linguistic feedbacks of users were processed using Person FOU approach (discussed in Section III) to recommend one processor frequency for each phase, game (and the user). The linguistic feedbacks of all the users and both the games are given in Section SM.IV of SM.

*1) Recommended Frequency*

Table VI lists the recommended frequencies obtained with Person FOU, extension principle based CWW approach, symbolic method based CWW approach and 2-tuple linguistic model based CWW approach, for all the games, bifurcated according to the training (T) and execution (E) phases. It is pertinent to mention that IA, EIA or HMA used in Person FOU, recommended same processor frequency for respective users and games (both phases), and therefore have placed under a single column under respective games.

After a deeper analysis of the user feedbacks (given in SM.IV of SM), we have uncovered that the CWW approaches based on extension principle, symbolic method and 2-tuple linguistic model fail to generate correct recommendations. For example, consider the data of User 22 in the training phase with the game subway surfers (Details in SM.IV of SM). The User provided following linguistic values for battery life, application ratings, type of applications and amount of time spent, respectively in frequencies $F1$, $F2$ and $F6$ as: $F1 = \{BH, AS, SI, M\}$, $F2 = \{BH, AS, FI, M\}$ and $F6 = \{BL, AEF, MI, L\}$, respectively. He/ she assigned the following weights to the battery life, application ratings, type of applications and amount of time spent, respectively as $\{MLI, I, U, U\}$. When we process the linguistic feedbacks of User 22 using Extension principle based CWW approach, the collective performance vector obtained for $F1, F2$ and $F6$ are $\{0.06, 0.17, 0.42\}$, $\{0.06, 0.17, 0.42\}$ and $\{0.05, 0.17, 0.41\}$, respectively. Thus, with all the frequencies, the extension principle based CWW approach recommends a distance term of $d_2$, from the distance vector.

With symbolic method based CWW approach, the recommended distance vector term for $F1, F2$ and $F6$ is $d_3$, $d_3$ and $d_4$, respectively. Similarly, with 2-tuple based CWW approach, the recommended respective distance vector term for $F1, F2$ and $F6$ is $(d_3, 0)$, $(d_3, 0.11)$ and $(d_3, 0.22)$. However, when we process the linguistic feedbacks of user 22 with Person FOU, the mean centroid values obtained are $F1 = 6.08, F2 = 6.31$ and $F6 = 4.59$, respectively. Thus, due to the enhanced capability of Person FOU in effectively capturing the variations in user 22's feedback, it recommends the frequency as $F2$. The CWW approaches based on extension principle, symbolic method and 2-tuple linguistic model, on the other hand are ignorant to the variations in the user's feedback, and end up recommending the frequency as $F1$, $F1$ and $F6$, respectively.

A major disadvantage of ignoring this variation among the user 22's feedbacks for the two frequencies would have been the selection of incorrect processor frequency. A power optimization policy designed using the concepts of CWW approaches based on extension principle, symbolic method and 2-tuple linguistic model would end up choosing $F1$, $F1$ and $F6$, respectively. Scaling the processor frequency at $F1$ would make the user's device slow beyond the point of usability and $F6$ would increase the dynamic power dissipation. In both the cases, user dissatisfaction will increase. The detailed discussions about the effect of incorrect processor frequency selection on power consumption and satisfaction ratings are given in Sections V.D.2 and V.D.3.

Therefore, by similar analysis of other values from Table VI, we can see that CWW approach based on extension principle, symbolic method and 2-tuple failed to recommend correct frequency for the training phase of game Subway surfers in 76%, 72% and 48% cases, respectively. Corresponding values for execution phases are 44%, 80% and 8%, respectively. With the training phase of Asphalt 8: Airborne, incorrect frequency recommendation by the CWW approaches based on extension principle, symbolic method and 2-tuple occurred in 64%, 36% and 4% cases, respectively. The corresponding values in the execution phase are 72%, 56% and 24%. In the training phase of Fruit Ninja, the respective failure rate of the CWW approaches based on extension principle, symbolic method and 2-tuple is 76%, 72% and 20% cases. The corresponding values in the execution phase are 76%, 68% and 12% cases.

*2) Power Consumption*

Table VII gives the values of average power consumed in Watts per second (averaged over time), in a frequency $Fi$, bifurcated by the games Subway Surfers, Asphalt 8: Airborne and Fruit Ninja (Please see [14] for details). Using the frequency values given in Table VI and associated power values from Table VII, the average power consumed across all users for both phases (in Watts) by all the approaches and all three games are presented in Table VIII, as well as Fig. 10.

From the Table VIII we can see that CWW approaches based on Extension principle and symbolic method consume considerably lesser power than the Person FOU, whereas the CWW approach based on 2-tuple linguistic model consumes almost the same power as the Person FOU.

However, we discussed in Section V.D.1 that the CWW approaches based on extension principle, symbolic method and 2-tuple recommend incorrect processor frequencies in a large number of cases. As the power consumption is closely related to the processor frequency [14], [15], therefore, the tendency of CWW approaches based on extension principle,

TABLE VII
AVERAGE POWER CONSUMED (WATTS PER SECOND)

| Frequency ($Fi$) | Subway Surfers | Asphalt 8: Airborne | Fruit Ninja |
|---|---|---|---|
| F1 | 1.41 | 1.92 | 0.97 |
| F2 | 1.84 | 2.32 | 1.64 |
| F3 | 2.18 | 2.85 | 2.01 |
| F4 | 2.57 | 2.99 | 2.36 |
| F5 | 2.71 | 3.15 | 2.5 |
| F6 | 3.65 | 3.85 | 3.26 |

Table VIII
Average system-wide power consumption and improvements

| Phase | Game | Power Consumed (Watts) | | | | % Improvement of Person FOU w.r.t. | | |
|---|---|---|---|---|---|---|---|---|
| | | PF[a] | EP[b] | SM[c] | 2-tuple[d] | EP[b] | SM[c] | 2-tuple[d] |
| Training | Subway Surfers | 2.71 | 2.17 | 2 | 2.83 | -21.67 | -19.67 | 3.33 |
| | Asphalt 8: Airborne | 3.32 | 2.68 | 2.97 | 3.32 | | | |
| | Fruit Ninja | 2.32 | 1.72 | 1.81 | 2.46 | | | |
| Execution | Subway surfers | 2.07 | 1.75 | 1.62 | 2.11 | -19.33 | -15 | 1 |
| | Asphalt 8: Airborne | 2.8 | 2.42 | 2.58 | 2.81 | | | |
| | Fruit Ninja | 1.99 | 1.42 | 1.69 | 2 | | | |

[a]Person FOU (based on IA, EIA or HMA), [b]Extension principle based CWW approach, [c]Symbolic method based CWW approach, [d]2-tuple based CWW approach

Table IX
Average satisfaction rating and improvements for all the games

| Phase | Game | Satisfaction Ratings | | | | % Improvement of Person FOU w.r.t. | | |
|---|---|---|---|---|---|---|---|---|
| | | PF[a] | EP[b] | SM[c] | 2-tuple[d] | EP[b] | SM[c] | 2-tuple[d] |
| Training | Subway Surfers | 3.84 | 2.84 | 3.36 | 3.75 | 26.33 | 13.33 | 3.67 |
| | Asphalt 8: Airborne | 3.8 | 3.12 | 3.36 | 3.52 | | | |
| | Fruit Ninja | 3.72 | 3.04 | 3.28 | 3.68 | | | |
| Execution | Subway surfers | 3.64 | 2.4 | 3.04 | 3.6 | 45 | 19 | 2.67 |
| | Asphalt 8: Airborne | 3.68 | 2.44 | 3.00 | 3.52 | | | |
| | Fruit Ninja | 3.60 | 2.72 | 3.16 | 3.52 | | | |

aPerson FOU (based on IA, EIA or HMA), [b]Extension principle based CWW approach, [c]Symbolic method based CWW approach, [d]2-tuple based CWW approach symbolic method and 2-tuple, to dissipate lesser (or almost the same) power, compared to Person FOU, is not an accurate test of their efficiencies. In fact the power savings achieved by CWW approaches based on extension principle, symbolic method and 2-tuple are accompanied by increase in user dissatisfaction (we discuss this in Section V.D.3).

*3) Satisfaction Rating*

We have calculated the user satisfaction ratings (of individual users) with Person FOU using the methodology given in [14], whereas for the other CWW approaches based on extension principle, symbolic method and 2-tuple, using the methodology given in Sections V.A to V.C. The average satisfaction ratings during the training and the execution phases for all CWW approaches are given in Table IX and depicted pictorially in Fig. 11. The average improvement in the satisfaction ratings achieved by Person FOU during training and execution phases of all the games taken together is 26.33% and 45%, respectively. The corresponding values for the CWW approach based on symbolic method is 13.33% and 19%, respectively, whereas for the 2-tuple based CWW approach is 3.67% and 2.67%, respectively. So, as discussed in Sections V.D.1 (and V.D.2), the reason for higher user satisfaction achieved by Person FOU lies in correct frequency recommendations, which other CWW approaches failed to do.

## VI. Results And Discussion

Here, we discuss some of the core learnings gained from the empirical study, presented in this paper.

### *A. Multi-person systems*

Table I depicts the amount of word fuzziness captured by IA, EIA and HMA, for multi-person systems. It also depicts the percentage improvement of fuzziness captured by IA, compared to EIA and HMA. It can be seen that EIA captures 7.52% lesser fuzziness than the IA, when compared on the basis of mean fuzziness. Corresponding value for HMA is 33.25%. For multi-person systems, as the amount of uncertainty is generally very high, therefore, the word modelling technique used should be able to capture and model it efficiently. EIA improves the methodology for calculating the height of LMF corresponding to interior FOU, in the FS part [11]. Also, it captures almost same amount of fuzziness as the IA. Therefore, we propose that for multi-person systems, EIA is a better word modelling approach than HMA and IA.

### *B. Single-person systems*

Table II depicts the amount of word fuzziness captured by IA, EIA and HMA, for single-person systems. It also depicts the percentage improvement of fuzziness captured by IA, compared to EIA and HMA. It can be seen that EIA captures 0.16% lesser fuzziness than the IA, when compared on the basis of mean fuzziness. Corresponding value for HMA is 29.35%. In this recent work [14], it was shown that HMA is computationally faster and IA (or EIA).

For single-person systems, as the amount of uncertainty is generally not very high. Therefore, we propose that for single-person systems, HMA is a better word modelling approach than IA and EIA.

### *C. Comparative analysis on the basis of variation capturing capabilities by CWW approaches: Insights from single-person systems*

We now compare the capabilities of Person FOU, extension principle based CWW approach, symbolic method based CWW approach and 2-tuple based CWW approach in capturing the variations in the linguistic data processed by the respective approach. It is pertinent to mention that for Person FOU approach, results derived are equally applicable to perceptual computing also.

Returning to the sample from of [14], there were 25users who provided their feedbacks about various system criteria for all the games. Of these 25 users, there were 13 users who provided equal weights to all the criteria. Let these 13 users constitute the Group 1. The other 12 users gave different weights to different criteria, and be called the Group 2. Table X presents the number of cases in which the CWW approaches based on extension principle, symbolic method and 2-tuple linguistic model failed to recommend correct processor frequency. Person FOU on the other hand recommended correct processor frequencies in all the cases. From the Table X, it can be seen that for the Group 1 users, the number of cases where the CWW approaches failed to recommend a correct frequency are a total of 90, when taken for all the games and phases taken together. Whereas the corresponding number of cases for Group 2 users is 137.

Clearly, there are 52.22% more cases of Group 2 users where the CWW approaches failed to recommend a correct frequency as compared to Group 1 cases. Thus, as the variation in the data increases, the uncertainty and correct recommendation generating capability of CWW approaches based on extension principle, symbolic method and 2-tuple decreases. Person FOU (or perceptual computing) on the other hand is able to capture the variations in the data in the best possible way.

Table X
Number of cases where CWW approaches based on extension principle, symbolic method and 2-tuple recommended incorrect processor frequency

| Game | CWW approach | Phase | Number of cases | | Total cases |
|---|---|---|---|---|---|
| | | | Group 1[a] | Group 2[b] | |
| Subway Surfers | Extension Principle based CWW approach | Training | 8 | 11 | 19 |
| | | Execution | 5 | 6 | 11 |
| | Symbolic method based CWW approach | Training | 6 | 12 | 18 |
| | | Execution | 8 | 12 | 20 |
| | 2-tuple based CWW approach | Training | 6 | 6 | 12 |
| | | Execution | 0 | 2 | 2 |
| Asphalt 8: Airborne | Extension Principle based CWW approach | Training | 8 | 8 | 16 |
| | | Execution | 11 | 7 | 18 |
| | Symbolic method based CWW approach | Training | 0 | 9 | 9 |
| | | Execution | 6 | 8 | 14 |
| | 2-tuple based CWW approach | Training | 0 | 1 | 1 |
| | | Execution | 3 | 3 | 6 |
| Fruit Ninja | Extension Principle based CWW approach | Training | 9 | 10 | 19 |
| | | Execution | 7 | 12 | 19 |
| | Symbolic method based CWW approach | Training | 6 | 12 | 18 |
| | | Execution | 5 | 12 | 17 |
| | 2-tuple based CWW approach | Training | 1 | 4 | 5 |
| | | Execution | 1 | 2 | 3 |
| Total | | | 90 | 137 | 227 |

[a]Users who assigned equal weights to all the criteria [b]Users who assigned different weights to all the criteria

Table XI
Average power consumption and Average satisfaction ratings

| Game | CWW approach | Phase | Average Power Consumption (Watts per second) | | Average Satisfaction Ratings | |
|---|---|---|---|---|---|---|
| | | | Group 1[a] | Group 2[b] | Group 1[a] | Group 2[b] |
| Subway Surfers | Person FOU | Training | 2.79 | 2.63 | 3.77 | 3.92 |
| | | Execution | 2.1 | 2.03 | 3.46 | 3.83 |
| | Extension Principle based CWW approach | Training | 2.41 | 1.9 | 3.38 | 2.77 |
| | | Execution | 1.83 | 1.65 | 2.25 | 2.00 |
| | Symbolic method based CWW approach | Training | 2.43 | 1.65 | 2.77 | 3.50 |
| | | Execution | 1.54 | 1.57 | 2.15 | 3.50 |
| | 2-tuple based CWW approach | Training | 2.77 | 2.89 | 3.50 | 3.92 |
| | | Execution | 2.1 | 2.12 | 3.62 | 3.58 |
| Asphalt 8: Airborne | Person FOU | Training | 3.45 | 3.19 | 3.77 | 3.83 |
| | | Execution | 2.85 | 2.74 | 3.77 | 3.58 |
| | Extension Principle based CWW approach | Training | 2.85 | 2.49 | 3.85 | 2.33 |
| | | Execution | 2.48 | 2.35 | 2.77 | 2.08 |
| | Symbolic method based CWW approach | Training | 3.45 | 2.45 | 3.00 | 3.75 |
| | | Execution | 2.66 | 2.48 | 2.38 | 3.75 |
| | 2-tuple based CWW approach | Training | 3.45 | 3.19 | 3.50 | 3.50 |
| | | Execution | 2.85 | 2.76 | 3.54 | 3.50 |
| Fruit Ninja | Person FOU | Training | 2.32 | 2.33 | 3.77 | 3.67 |
| | | Execution | 2.00 | 1.99 | 3.62 | 3.58 |
| | Extension Principle based CWW approach | Training | 2.01 | 1.4 | 3.69 | 2.33 |
| | | Execution | 1.66 | 1.17 | 3.23 | 2.17 |
| | Symbolic method based CWW approach | Training | 2.24 | 1.35 | 2.69 | 3.92 |
| | | Execution | 1.92 | 1.44 | 2.46 | 3.92 |
| | 2-tuple based CWW approach | Training | 2.52 | 2.4 | 3.50 | 3.75 |
| | | Execution | 2.04 | 1.95 | 3.69 | 3.33 |

[a]Users who assigned equal weights to all the criteria [b]Users who assigned different weights to all the criteria

Now let's consider the variations in the power consumption. The average power consumed by both groups of users in all the phases and games is given in the Table XI. From the Table X, it can be seen that the power consumed by the Group 1 users is 58.72 Watts per second across all the games and phases taken together, whereas by Group 2 users is 52.12 Watts per second. Clearly, the group 1 users consume 12.66% more power than the Group 2 users. Furthermore, though the average power consumption by CWW approaches based on extension principle, symbolic method and 2-tuple linguistic model is lesser than the Person FOU, but the reason for the same is the incorrect frequency recommendation.

Now let's consider the variations in the average satisfaction ratings. The average satisfaction rating by both groups of users in all the phases and games is given in the Table XI. From the Table XI, it can be seen that the average satisfaction ratings of the Group 1 users is 3.26 across all the games and phases taken together, whereas by Group 2 users is 3.33. Clearly, the Group 2 users are more satisfied than the Group 1 users. Furthermore, it can be seen from the Table XI that CWW approaches based on extension principle, symbolic method and 2-tuple linguistic model make users less satisfied in comparison to Person FOU.

## VII. Conclusions And Future Work

In this paper, we have presented a thorough empirical comparative performance analysis of various CWW approaches from the context of multi-person and single-person systems. In multi-person systems, we initially compared the performances of IA, EIA and HMA based perceptual computing approaches. We have found that due to overwhelming inter-person uncertainty in comparison to intra-person uncertainty in multi-person systems, EIA is better suited for word modelling in multi-person systems. Then we compared the performances of perceptual computing (based on IA, EIA or HMA), extension principle based CWW approach, symbolic method based CWW approach, and 2-tuple based CWW approach. We found the all the other CWW approaches, except perceptual computing failed to generate unique recommendations on multiple occasions.

In single-person systems, we first of all compared the performances of IA, EIA and HMA based Person FOU. We have found that due to overpowering by intra-person uncertainty in comparison to inter-person uncertainty in single-person systems, HMA is better suited for word modelling in single-person systems. We went ahead and also compared the performances of Person FOU (based on IA, EIA or HMA), extension principle based CWW approach, symbolic method based CWW approach, and 2-tuple based CWW approach. Here again, we found the all the other CWW approaches, except Person FOU failed to generate unique recommendations on multiple occasions.

Another type of comparative performance analysis was made possible by the nature of data in single-person systems. It was possible to divide the data into two categories viz., inputs with equal weights and inputs with differential weights. We have found that the CWW approaches based on extension principle, symbolic method and 2-tuple fail even strongly to generate unique recommendations in scenarios when the inputs become differentially weighted. Though this learning was established from the data of single-person systems, however, the results are equally applicable to multi-person systems also.

In future, we want to develop a hybrid approach which works commonly for both the multi-person and single-person systems.

# SUPPLEMENTARY MATERIALS

## SM-I. FOU DATA FOR MULTI-PERSON SYSTEMS

Here, we give the FOU data values for the words for various system criteria, generated using IA, EIA and HMA, based on the data intervals of a group of subjects.

TABLE SM-I
FOU DATA FOR "BATTERY LIFE"- THE CODEBOOK OBTAINED WITH IA

| Word | UMF | | | | LMF | | | | | Centroid | | Centroid Center |
|---|---|---|---|---|---|---|---|---|---|---|---|---|
| Very Low (BVL) | 0.00 | 0.00 | 0.18 | 2.63 | 0.00 | 0.00 | 0.09 | 1.32 | 1.00 | 0.44 | 0.93 | 0.68 |
| Low (BL) | 0.00 | 0.00 | 2.18 | 6.58 | 0.00 | 0.00 | 0.28 | 3.32 | 1.00 | 1.11 | 2.54 | 1.83 |
| Medium (BM) | 1.17 | 4.00 | 7.00 | 9.83 | 4.79 | 5.5 | 5.5 | 6.21 | 0.32 | 3.23 | 7.77 | 5.50 |
| High (BH) | 4.73 | 8.82 | 10.00 | 10.00 | 7.68 | 9.82 | 10.00 | 10.00 | 1.00 | 8.02 | 9.22 | 8.62 |
| Extremely High (BEH) | 6.05 | 9.72 | 10.00 | 10.00 | 8.68 | 9.91 | 10.00 | 10.00 | 1.00 | 8.53 | 9.56 | 9.04 |

TABLE SM-II
FOU DATA FOR "APPLICATION RATINGS"- THE CODEBOOK OBTAINED WITH IA

| Word | UMF | | | | LMF | | | | | Centroid | | Centroid Center |
|---|---|---|---|---|---|---|---|---|---|---|---|---|
| Very slow (AVS) | 0.00 | 0.00 | 0.28 | 3.95 | 0.00 | 0.00 | 0.09 | 1.32 | 1.00 | 0.44 | 1.47 | 0.96 |
| Slow (AS) | 0.59 | 2.00 | 3.00 | 4.41 | 1.79 | 2.50 | 2.5 | 3.21 | 0.59 | 1.88 | 3.12 | 2.50 |
| Moderate (AM) | 2.38 | 4.5 | 6.50 | 8.62 | 4.9 | 5.32 | 5.32 | 5.6 | 0.26 | 3.62 | 7.29 | 5.46 |
| Fast (AF) | 4.38 | 6.50 | 8.00 | 9.62 | 6.79 | 7.38 | 7.38 | 8.21 | 0.49 | 6.16 | 8.24 | 7.20 |
| Extremely fast (AEF) | 7.37 | 9.73 | 10.00 | 10.00 | 9.34 | 9.95 | 10.00 | 10.00 | 1.00 | 8.95 | 9.78 | 9.36 |

TABLE SM-III
FOU DATA FOR "TYPE OF APPLICATION"- THE CODEBOOK OBTAINED WITH IA

| Word | UMF | | | | LMF | | | | | Centroid | | Centroid Center |
|---|---|---|---|---|---|---|---|---|---|---|---|---|
| Absolutely uninteresting (AU) | 0.00 | 0.00 | 0.18 | 2.63 | 0.00 | 0.00 | 0.09 | 1.32 | 1.00 | 0.44 | 0.93 | 0.68 |
| Somewhat interesting (SI) | 0.59 | 2.00 | 3.25 | 4.41 | 2.19 | 2.67 | 2.67 | 3.21 | 0.45 | 1.79 | 3.38 | 2.58 |
| Fairly Interesting (FI) | 2.59 | 4.00 | 5.00 | 6.41 | 3.79 | 4.5 | 4.5 | 5.21 | 0.59 | 3.88 | 5.12 | 4.50 |
| More interesting (MI) | 4.38 | 6.50 | 7.5 | 8.62 | 6.9 | 7.15 | 7.15 | 7.6 | 0.41 | 5.77 | 7.79 | 6.78 |
| Absolutely interesting (AI) | 7.37 | 9.45 | 10.00 | 10 | 8.84 | 9.91 | 10.00 | 10.00 | 1.00 | 9.02 | 9.61 | 9.31 |

TABLE SM-IV
FOU DATA FOR "AMOUNT OF TIME SPENT"- THE CODEBOOK OBTAINED WITH IA

| Word | UMF | | | | LMF | | | | | Centroid | | Centroid Center |
|---|---|---|---|---|---|---|---|---|---|---|---|---|
| Very Little (VL) | 0.00 | 0.00 | 0.18 | 2.63 | 0.00 | 0.00 | 0.09 | 1.32 | 1.00 | 0.44 | 0.93 | 0.68 |
| Small (S) | 0.59 | 2.00 | 3.00 | 4.41 | 1.79 | 2.50 | 2.50 | 3.21 | 0.59 | 1.88 | 3.12 | 2.50 |
| Moderate (M) | 1.98 | 3.75 | 5.00 | 6.41 | 4.29 | 4.59 | 4.59 | 5.21 | 0.42 | 3.38 | 5.38 | 4.38 |
| Large (L) | 4.02 | 5.65 | 7.00 | 8.62 | 6.40 | 6.60 | 6.60 | 7.10 | 0.34 | 5.23 | 7.60 | 6.41 |
| Very large (VLA) | 6.05 | 9.72 | 10.00 | 10.00 | 8.68 | 9.91 | 10.00 | 10.00 | 1.00 | 8.53 | 9.56 | 9.04 |

TABLE SM-V
FOU DATA FOR "BATTERY LIFE"- THE CODEBOOK OBTAINED WITH EIA

| Word | UMF | | | | LMF | | | | | Centroid | | Centroid Center |
|---|---|---|---|---|---|---|---|---|---|---|---|---|
| Very Low (BVL) | 0.00 | 0.00 | 0.09 | 1.32 | 0.00 | 0.00 | 0.09 | 1.32 | 1.00 | 0.44 | 0.44 | 0.44 |
| Low (BL) | 0.00 | 0.00 | 2.18 | 5.27 | 0.00 | 0.00 | 0.28 | 3.32 | 1.00 | 1.11 | 2.07 | 1.59 |
| Medium (BM) | 3.17 | 5.00 | 6.50 | 8.83 | 4.79 | 5.6 | 5.6 | 6.21 | 0.58 | 4.71 | 6.86 | 5.79 |
| High (BH) | 6.05 | 8.82 | 10.00 | 10.00 | 7.68 | 9.82 | 10.00 | 10.00 | 1.00 | 8.51 | 9.22 | 8.87 |
| Extremely High (BEH) | 6.71 | 9.77 | 10.00 | 10.00 | 8.68 | 9.91 | 10.00 | 10.00 | 1.00 | 8.81 | 9.56 | 9.18 |

TABLE SM-VI
FOU DATA FOR "APPLICATION RATINGS"- THE CODEBOOK OBTAINED WITH EIA

| Word | UMF | | | | LMF | | | | | Centroid | | Centroid Center |
|---|---|---|---|---|---|---|---|---|---|---|---|---|
| Very slow (AVS) | 0.00 | 0.00 | 0.18 | 2.63 | 0.00 | 0.00 | 0.09 | 1.32 | 1.00 | 0.44 | 0.93 | 0.68 |
| Slow (AS) | 1.59 | 2.50 | 3.00 | 4.41 | 1.79 | 2.67 | 2.67 | 3.21 | 0.76 | 2.42 | 3.08 | 2.75 |
| Moderate (AM) | 3.59 | 5.00 | 5.50 | 6.41 | 4.9 | 5.33 | 5.33 | 5.6 | 0.76 | 4.69 | 5.62 | 5.15 |
| Fast (AF) | 5.59 | 7.00 | 7.75 | 8.81 | 6.79 | 7.43 | 7.43 | 8.21 | 0.70 | 6.92 | 7.79 | 7.35 |
| Extremely fast (AEF) | 8.68 | 9.91 | 10.00 | 10.00 | 8.68 | 9.91 | 10.00 | 10.00 | 1.00 | 9.56 | 9.56 | 9.56 |

TABLE SM-VII
FOU DATA FOR "TYPE OF APPLICATION"- THE CODEBOOK OBTAINED WITH EIA

| Word | UMF | | | | LMF | | | | | Centroid | | Centroid Center |
|---|---|---|---|---|---|---|---|---|---|---|---|---|
| Absolutely uninteresting (AU) | 0.00 | 0.00 | 0.09 | 1.32 | 0.00 | 0.00 | 0.09 | 1.32 | 1.00 | 0.44 | 0.44 | 0.44 |
| Somewhat interesting (SI) | 1.59 | 2.50 | 3.25 | 4.41 | 2.19 | 2.80 | 2.80 | 3.21 | 0.58 | 2.4 | 3.34 | 2.87 |
| Fairly Interesting (FI) | 3.59 | 4.50 | 5.00 | 6.41 | 3.79 | 4.67 | 4.67 | 5.21 | 0.76 | 4.42 | 5.08 | 4.75 |
| More interesting (MI) | 6.79 | 7.25 | 7.50 | 8.21 | 6.9 | 7.33 | 7.33 | 7.60 | 0.76 | 7.21 | 7.54 | 7.38 |
| Absolutely interesting (AI) | 7.37 | 9.45 | 10.00 | 10.00 | 8.84 | 9.91 | 10.00 | 10.00 | 1.00 | 9.02 | 9.61 | 9.31 |

TABLE SM-VIII
FOU DATA FOR "AMOUNT OF TIME SPENT"- THE CODEBOOK OBTAINED WITH EIA

| Word | UMF | | | | LMF | | | | | Centroid | | Centroid Center |
|---|---|---|---|---|---|---|---|---|---|---|---|---|
| Very Little (VL) | 0.79 | 1.50 | 2.50 | 3.21 | 1.79 | 2.00 | 2.00 | 2.21 | 0.29 | 1.34 | 2.66 | 2.00 |
| Small (S) | 1.79 | 2.50 | 2.50 | 3.21 | 1.79 | 2.5 | 2.50 | 3.21 | 1.00 | 2.50 | 2.50 | 2.50 |
| Moderate (M) | 3.59 | 4.50 | 5.00 | 6.41 | 4.29 | 4.75 | 4.75 | 5.21 | 0.65 | 4.41 | 5.30 | 4.85 |
| Large (L) | 5.19 | 6.25 | 7.00 | 8.41 | 6.4 | 6.63 | 6.63 | 7.10 | 0.65 | 6.13 | 7.34 | 6.74 |
| Very large (VLA) | 6.71 | 9.77 | 10.00 | 10.00 | 8.68 | 9.91 | 10.00 | 10.00 | 1.00 | 8.81 | 9.56 | 9.18 |

TABLE SM-IX
FOU DATA FOR "BATTERY LIFE"- THE CODEBOOK OBTAINED WITH HMA

| Word | UMF | | | | LMF | | | | | Centroid | | Centroid Center |
|---|---|---|---|---|---|---|---|---|---|---|---|---|
| Very Low (BVL) | 0.00 | 0.00 | 1.00 | 1.00 | 0.00 | 0.00 | 1.00 | 1.00 | 1.00 | 0.50 | 0.50 | 0.50 |
| Low (BL) | 0.00 | 0.00 | 3.00 | 9.00 | 0.00 | 0.00 | 3.00 | 3.00 | 1.00 | 1.50 | 3.28 | 2.39 |
| Medium (BM) | 0.00 | 5.00 | 6.00 | 10.00 | 5.00 | 5.00 | 6.00 | 8.11 | 1.00 | 4.57 | 6.73 | 5.65 |
| High (BH) | 2.00 | 8.00 | 10.00 | 10.00 | 8.00 | 8.00 | 10.00 | 10.00 | 1.00 | 7.10 | 8.99 | 8.04 |
| Extremely High (BEH) | 4.84 | 9.00 | 10.00 | 10.00 | 7.34 | 9.00 | 10.00 | 10.00 | 1.00 | 8.20 | 9.01 | 8.60 |

TABLE SM-X
FOU DATA FOR "APPLICATION RATINGS"- THE CODEBOOK OBTAINED WITH HMA

| Word | UMF | | | | LMF | | | | | Centroid | | Centroid Center |
|---|---|---|---|---|---|---|---|---|---|---|---|---|
| Very slow (AVS) | 0.00 | 0.00 | 1.00 | 7.00 | 0.00 | 0.00 | 1.00 | 1.00 | 1.00 | 0.53 | 2.72 | 1.63 |
| Slow (AS) | 0.00 | 2.00 | 3.00 | 9.00 | 0.00 | 2.00 | 3.00 | 3.00 | 1.00 | 1.89 | 3.84 | 2.87 |
| Moderate (AM) | 0.00 | 5.00 | 5.50 | 8.50 | 5.00 | 5.00 | 5.50 | 5.50 | 1.00 | 3.19 | 6.35 | 4.77 |
| Fast (AF) | 1.00 | 7.00 | 8.00 | 10.00 | 7.00 | 7.00 | 8.00 | 8.00 | 1.00 | 5.21 | 8.11 | 6.66 |
| Extremely fast (AEF) | 9.00 | 9.00 | 10.00 | 10.00 | 9.00 | 9.00 | 10.00 | 10.00 | 1.00 | 9.50 | 9.50 | 9.50 |

TABLE SM-XI
FOU DATA FOR "TYPE OF APPLICATION"- THE CODEBOOK OBTAINED WITH HMA

| Word | UMF | | | | LMF | | | | | Centroid | | Centroid Center |
|---|---|---|---|---|---|---|---|---|---|---|---|---|
| Absolutely uninteresting (AU) | 0.00 | 0.00 | 1.00 | 1.00 | 0.00 | 0.00 | 1.00 | 1.00 | 1.00 | 0.50 | 0.50 | 0.50 |
| Somewhat interesting (SI) | 0.00 | 2.50 | 3.00 | 9.00 | 2.50 | 2.50 | 3.00 | 3.00 | 1.00 | 1.83 | 5.45 | 3.64 |
| Fairly Interesting (FI) | 0.00 | 4.00 | 5.00 | 10.00 | 0.00 | 4.00 | 5.00 | 5.00 | 1.00 | 3.28 | 4.82 | 4.05 |
| More interesting (MI) | 0.00 | 7.00 | 7.50 | 10.00 | 0.00 | 7.00 | 7.50 | 7.50 | 1.00 | 5.01 | 5.79 | 5.40 |
| Absolutely interesting (AI) | 3.00 | 9.00 | 10.00 | 10.00 | 9.00 | 9.00 | 10.00 | 10.00 | 1.00 | 7.28 | 9.47 | 8.37 |

TABLE SM-XII
FOU DATA FOR "AMOUNT OF TIME SPENT"- THE CODEBOOK OBTAINED WITH HMA

| Word | UMF | | | | LMF | | | | | Centroid | | Centroid Center |
|---|---|---|---|---|---|---|---|---|---|---|---|---|
| Very Little (VL) | 0.00 | 1.00 | 2.00 | 10.00 | 0.00 | 1.00 | 2.00 | 10.00 | 1.00 | 3.73 | 3.73 | 3.73 |
| Small (S) | 0.00 | 2.00 | 3.00 | 10.00 | 0.00 | 2.00 | 3.00 | 10.00 | 1.00 | 4.09 | 4.09 | 4.09 |
| Moderate (M) | 1.50 | 4.50 | 5.00 | 8.78 | 4.50 | 4.50 | 5.00 | 6.22 | 1.00 | 4.17 | 5.96 | 5.07 |
| Large (L) | 3.87 | 6.50 | 7.00 | 10.00 | 5.79 | 6.50 | 7.00 | 8.06 | 1.00 | 6.26 | 7.48 | 6.87 |
| Very large (VLA) | 4.84 | 9.00 | 10.00 | 10.00 | 7.34 | 9.00 | 10.00 | 10.00 | 1.00 | 8.20 | 9.01 | 8.60 |

## SM-II. FOU DATA FOR SINGLE-PERSON SYSTEMS

Here, we give the FOU data values for the words for various system criteria, generated using IA, EIA and HMA, based on the data intervals of a single subject.

TABLE SM-XIII
FOU DATA OF USER 22 FOR "BATTERY LIFE" OBTAINED WITH IA- THE CODEBOOK

| Word | UMF | | | | LMF | | | | | Centroid | | Centroid Center |
|---|---|---|---|---|---|---|---|---|---|---|---|---|
| Very Low (BVL) | 0.00 | 0.00 | 0.65 | 3.24 | 0.00 | 0.00 | 0.21 | 2.53 | 1.00 | 0.84 | 1.15 | 0.99 |
| Low (BL) | 0.66 | 1.59 | 2.28 | 3.03 | 1.66 | 1.9 | 1.9 | 2.09 | 0.38 | 1.31 | 2.44 | 1.87 |
| Medium (BM) | 1.81 | 2.53 | 3.34 | 4.29 | 2.81 | 2.96 | 2.96 | 3.12 | 0.27 | 2.3 | 3.72 | 3.01 |
| High (BH) | 4.37 | 6.24 | 7.86 | 9.74 | 6.89 | 7.05 | 7.05 | 7.22 | 0.17 | 5.21 | 8.9 | 7.05 |
| Extremely High (BEH) | 7.78 | 8.55 | 9.31 | 10.0 | 8.84 | 8.98 | 8.98 | 9.18 | 0.31 | 8.33 | 9.51 | 8.92 |

TABLE SM-XIV
FOU DATA OF USER 22 FOR "APPLICATION RATINGS" OBTAINED WITH IA- THE CODEBOOK

| Word | UMF | | | | LMF | | | | | Centroid | | Centroid Center |
|---|---|---|---|---|---|---|---|---|---|---|---|---|
| Very slow (AVS) | 0.01 | 1.21 | 1.85 | 3.37 | 0.73 | 1.54 | 1.54 | 2.41 | 0.72 | 1.2 | 2 | 1.60 |
| Slow (AS) | 2.27 | 4.57 | 5.41 | 7.63 | 3.49 | 4.99 | 4.99 | 6.48 | 0.78 | 4.43 | 5.52 | 4.98 |
| Moderate (AM) | 4.49 | 6.03 | 6.89 | 8.54 | 5.65 | 6.47 | 6.47 | 7.32 | 0.66 | 5.91 | 7.07 | 6.49 |
| Fast (AF) | 6.76 | 7.65 | 8.25 | 9.20 | 7.83 | 7.96 | 7.96 | 8.1 | 0.31 | 7.29 | 8.65 | 7.97 |
| Extremely fast (AEF) | 7.78 | 8.63 | 9.34 | 9.98 | 8.95 | 9.02 | 9.02 | 9.11 | 0.18 | 8.17 | 9.65 | 8.91 |

TABLE SM-XV
FOU DATA OF USER 22 FOR "TYPE OF APPLICATION" OBTAINED WITH IA- THE CODEBOOK

| Word | UMF | | | | LMF | | | | | Centroid | | Centroid Center |
|---|---|---|---|---|---|---|---|---|---|---|---|---|
| Absolutely uninteresting (AU) | 0.03 | 0.62 | 1.47 | 2.3 | 0.9 | 1.00 | 1.00 | 1.08 | 0.18 | 0.36 | 1.88 | 1.12 |
| Somewhat interesting (SI) | 1.54 | 3.18 | 4.32 | 5.55 | 3.94 | 4.04 | 4.04 | 4.34 | 0.26 | 2.48 | 4.88 | 3.68 |
| Fairly Interesting (FI) | 3.52 | 5.17 | 6.40 | 7.52 | 5.82 | 6.03 | 6.03 | 6.49 | 0.35 | 4.7 | 6.73 | 5.72 |
| More interesting (MI) | 6.89 | 7.62 | 8.39 | 9.35 | 7.9 | 7.97 | 7.97 | 8.03 | 0.15 | 7.21 | 8.97 | 8.09 |
| Absolutely interesting (AI) | 7.74 | 8.67 | 9.26 | 9.92 | 8.79 | 8.95 | 8.95 | 9.11 | 0.35 | 8.32 | 9.44 | 8.88 |

TABLE SM-XVI
FOU DATA OF USER 22 FOR "AMOUNT OF TIME SPENT" OBTAINED WITH IA- THE CODEBOOK

| Word | UMF | | | | LMF | | | | | Centroid | | Centroid Center |
|---|---|---|---|---|---|---|---|---|---|---|---|---|
| Very Little (VL) | 0.03 | 1.33 | 2.40 | 3.41 | 1.89 | 2.03 | 2.03 | 2.30 | 0.28 | 0.84 | 2.78 | 1.81 |
| Small (S) | 2.63 | 3.57 | 4.40 | 5.38 | 3.91 | 4.02 | 4.02 | 4.16 | 0.23 | 3.15 | 4.86 | 4.00 |
| Moderate (M) | 4.70 | 5.95 | 7.18 | 8.46 | 6.33 | 6.51 | 6.51 | 6.67 | 0.22 | 5.38 | 7.75 | 6.56 |
| Large (L) | 6.62 | 7.55 | 8.48 | 9.32 | 7.85 | 7.99 | 7.99 | 8.11 | 0.22 | 7.13 | 8.83 | 7.98 |
| Very large (VLA) | 8.31 | 8.87 | 9.32 | 9.99 | 8.89 | 9.01 | 9.01 | 9.06 | 0.28 | 8.62 | 9.6 | 9.11 |

TABLE SM-XVII
FOU DATA OF USER 22 FOR "SATISFACTION RATING" OBTAINED WITH IA- THE CODEBOOK

| Word | UMF | | | | LMF | | | | | Centroid | | Centroid Center |
|---|---|---|---|---|---|---|---|---|---|---|---|---|
| Not Satisfied (NS) | 0.01 | 0.69 | 1.48 | 2.23 | 0.85 | 1.05 | 1.05 | 1.23 | 0.32 | 0.52 | 1.67 | 1.09 |
| Somehow Satisfied (SOS) | 1.49 | 3.14 | 4.32 | 5.46 | 3.9 | 4.01 | 4.01 | 4.31 | 0.26 | 2.44 | 4.81 | 3.63 |
| Satisfied (SS) | 3.69 | 4.6 | 5.41 | 6.81 | 4.92 | 5.00 | 5.00 | 5.58 | 0.17 | 4.60 | 6.41 | 5.50 |
| Very Satisfied (VS) | 5.61 | 7.07 | 8.36 | 9.48 | 7.87 | 7.96 | 7.96 | 8.18 | 0.20 | 6.38 | 8.90 | 7.64 |
| Overly Satisfied (OS) | 7.77 | 8.55 | 9.26 | 9.95 | 8.86 | 8.96 | 8.96 | 9.09 | 0.24 | 8.23 | 9.55 | 8.89 |

Table SM-XVIII
FOU DATA OF USER 22 FOR "LINGUISTIC WEIGHTS" OBTAINED WITH IA- THE CODEBOOK

| Word | UMF | | | | LMF | | | | | Centroid | | Centroid Center |
|---|---|---|---|---|---|---|---|---|---|---|---|---|
| Unimportant (U) | 0.00 | 0.00 | 0.65 | 3.21 | 0.00 | 0.00 | 0.18 | 2.53 | 1.00 | 0.84 | 1.14 | 0.99 |
| More or less unimportant (MLU) | 0.66 | 1.57 | 2.22 | 3.14 | 1.64 | 1.89 | 1.89 | 2.12 | 0.43 | 1.33 | 2.46 | 1.89 |
| Important (I) | 1.67 | 2.6 | 3.31 | 4.32 | 2.91 | 2.99 | 2.99 | 3.09 | 0.20 | 2.11 | 3.87 | 2.99 |
| More or less important (MLI) | 4.46 | 6.31 | 7.95 | 9.61 | 6.93 | 7.04 | 7.04 | 7.13 | 0.11 | 5.07 | 9.02 | 7.04 |
| Very Important (VI) | 7.79 | 8.62 | 9.29 | 9.94 | 8.93 | 9.00 | 9.00 | 9.10 | 0.20 | 8.19 | 9.61 | 8.90 |

TABLE SM-XIX
FOU DATA OF USER 22 FOR "BATTERY LIFE" OBTAINED WITH EIA- THE CODEBOOK

| Word | UMF | | | | LMF | | | | | Centroid | | Centroid Center |
|---|---|---|---|---|---|---|---|---|---|---|---|---|
| Very Low (BVL) | 0.00 | 0.00 | 0.65 | 3.24 | 0.00 | 0.00 | 0.21 | 2.53 | 1.00 | 0.85 | 1.15 | 1.00 |
| Low (BL) | 0.66 | 1.59 | 2.28 | 3.03 | 1.66 | 1.93 | 1.93 | 2.09 | 0.49 | 1.35 | 2.41 | 1.88 |
| Medium (BM) | 1.81 | 2.53 | 3.44 | 4.29 | 2.81 | 3.00 | 3.00 | 3.12 | 0.33 | 2.33 | 3.71 | 3.02 |
| High (BH) | 4.37 | 6.24 | 7.86 | 9.42 | 6.89 | 7.01 | 7.01 | 7.22 | 0.34 | 5.38 | 8.50 | 6.94 |
| Extremely High (BEH) | 7.78 | 8.55 | 9.31 | 10.00 | 8.84 | 8.98 | 8.98 | 9.18 | 0.32 | 8.33 | 9.51 | 8.92 |

TABLE SM-XX
FOU DATA OF USER 22 FOR "APPLICATION RATINGS" OBTAINED WITH EIA- THE CODEBOOK

| Word | UMF | | | | LMF | | | | | Centroid | | Centroid Center |
|---|---|---|---|---|---|---|---|---|---|---|---|---|
| Very slow (AVS) | 0.01 | 1.21 | 1.85 | 3.37 | 0.73 | 1.52 | 1.52 | 2.41 | 0.74 | 1.2 | 1.99 | 1.6 |
| Slow (AS) | 2.27 | 4.57 | 5.41 | 7.63 | 3.49 | 4.98 | 4.98 | 6.48 | 0.8 | 4.43 | 5.51 | 4.97 |
| Moderate (AM) | 4.49 | 6.03 | 6.89 | 8.54 | 5.65 | 6.49 | 6.49 | 7.32 | 0.68 | 5.93 | 7.06 | 6.49 |
| Fast (AF) | 6.76 | 7.65 | 8.25 | 9.15 | 7.83 | 7.91 | 7.91 | 8.10 | 0.5 | 7.36 | 8.54 | 7.95 |
| Extremely fast (AEF) | 7.78 | 8.63 | 9.34 | 9.98 | 8.95 | 9.00 | 9.00 | 9.11 | 0.35 | 8.25 | 9.59 | 8.92 |

TABLE SM-XXI
FOU DATA OF USER 22 FOR "TYPE OF APPLICATION" OBTAINED WITH EIA- THE CODEBOOK

| Word | UMF | | | | LMF | | | | | Centroid | | Centroid Center |
|---|---|---|---|---|---|---|---|---|---|---|---|---|
| Absolutely uninteresting (AU) | 0.03 | 0.62 | 1.47 | 2.3 | 0.9 | 1.01 | 1.01 | 1.08 | 0.31 | 0.42 | 1.81 | 1.11 |
| Somewhat interesting (SI) | 1.54 | 3.06 | 4.32 | 5.55 | 3.94 | 4 | 4 | 4.18 | 0.32 | 2.38 | 4.92 | 3.65 |
| Fairly Interesting (FI) | 3.52 | 5.17 | 6.4 | 7.52 | 5.82 | 5.98 | 5.98 | 6.14 | 0.34 | 4.46 | 6.84 | 5.65 |
| More interesting (MI) | 7.03 | 7.72 | 8.39 | 9.2 | 7.9 | 8.02 | 8.02 | 8.11 | 0.41 | 7.48 | 8.68 | 8.08 |
| Absolutely interesting (AI) | 7.74 | 8.67 | 9.26 | 9.92 | 8.79 | 8.94 | 8.94 | 9.11 | 0.47 | 8.37 | 9.4 | 8.88 |

TABLE SM-XXII
FOU DATA OF USER 22 FOR "AMOUNT OF TIME SPENT" OBTAINED WITH EIA- THE CODEBOOK

| Word | UMF | | | | LMF | | | | | Centroid | | Centroid Center |
|---|---|---|---|---|---|---|---|---|---|---|---|---|
| Very Little (VL) | 0.03 | 1.24 | 2.26 | 3.41 | 1.8 | 1.97 | 1.97 | 2.14 | 0.36 | 0.83 | 2.72 | 1.78 |
| Small (S) | 2.63 | 3.57 | 4.4 | 5.38 | 3.91 | 4.02 | 4.02 | 4.16 | 0.39 | 3.23 | 4.78 | 4 |
| Moderate (M) | 4.91 | 6.17 | 6.97 | 8.06 | 6.33 | 6.59 | 6.59 | 6.67 | 0.4 | 5.67 | 7.35 | 6.51 |
| Large (L) | 6.62 | 7.55 | 8.48 | 9.32 | 7.85 | 7.99 | 7.99 | 8.11 | 0.34 | 7.2 | 8.76 | 7.98 |
| Very large (VLA) | 8.31 | 8.87 | 9.32 | 9.99 | 8.89 | 9.05 | 9.05 | 9.14 | 0.49 | 8.71 | 9.5 | 9.11 |

TABLE SM-XXIII
FOU DATA OF USER 22 FOR "SATISFACTION RATING" OBTAINED WITH EIA- THE CODEBOOK

| Word | UMF | | | | LMF | | | | | Centroid | | Centroid Center |
|---|---|---|---|---|---|---|---|---|---|---|---|---|
| Not Satisfied (NS) | 0.01 | 0.69 | 1.48 | 2.23 | 0.85 | 1.05 | 1.05 | 1.23 | 0.37 | 0.54 | 1.65 | 1.09 |
| Somehow Satisfied (SOS) | 1.7 | 3.12 | 4.32 | 5.46 | 3.9 | 3.95 | 3.95 | 4.09 | 0.35 | 2.46 | 4.87 | 3.66 |
| Satisfied (SS) | 3.78 | 4.61 | 5.41 | 6.21 | 4.92 | 4.99 | 4.99 | 5.08 | 0.36 | 4.26 | 5.74 | 5 |
| Very Satisfied (VS) | 5.65 | 7.27 | 8.36 | 9.48 | 7.86 | 7.94 | 7.94 | 8.09 | 0.39 | 6.51 | 8.84 | 7.67 |
| Overly Satisfied (OS) | 7.77 | 8.55 | 9.26 | 9.95 | 8.86 | 9.01 | 9.01 | 9.09 | 0.37 | 8.29 | 9.5 | 8.9 |

Table SM-XXIV
FOU DATA OF USER 22 FOR "LINGUISTIC WEIGHTS" OBTAINED WITH EIA- THE CODEBOOK

| Word | UMF | | | | LMF | | | | | Centroid | | Centroid Center |
|---|---|---|---|---|---|---|---|---|---|---|---|---|
| Unimportant (U) | 0 | 0 | 0.65 | 3.21 | 0 | 0 | 0.18 | 2.53 | 1 | 0.84 | 1.14 | 0.99 |
| More or less unimportant (MLU) | 0.66 | 1.57 | 2.22 | 3.14 | 1.64 | 1.89 | 1.89 | 2.12 | 0.49 | 1.35 | 2.43 | 1.89 |
| Important (I) | 1.67 | 2.6 | 3.31 | 4.32 | 2.91 | 2.99 | 2.99 | 3.09 | 0.33 | 2.18 | 3.8 | 2.99 |
| More or less important (MLI) | 4.79 | 6.31 | 7.85 | 9.44 | 6.93 | 7.05 | 7.05 | 7.13 | 0.35 | 5.57 | 8.63 | 7.1 |
| Very Important (VI) | 7.79 | 8.62 | 9.29 | 9.94 | 8.93 | 9 | 9 | 9.1 | 0.38 | 8.28 | 9.54 | 8.91 |

TABLE SM-XXV
FOU DATA OF USER 22 FOR "BATTERY LIFE" OBTAINED WITH HMA- THE CODEBOOK

| Word | UMF | | | | LMF | | | | | Centroid | | Centroid Center |
|---|---|---|---|---|---|---|---|---|---|---|---|---|
| Very Low (BVL) | 0.00 | 0.00 | 2.01 | 2.89 | 0.00 | 0.00 | 2.01 | 2.58 | 1.00 | 1.16 | 1.24 | 1.20 |
| Low (BL) | 0.11 | 1.8 | 2.01 | 3.36 | 0.81 | 1.8 | 2.01 | 2.96 | 1.00 | 1.67 | 2.02 | 1.84 |
| Medium (BM) | 0.83 | 2.98 | 3.03 | 5.09 | 1.92 | 2.98 | 3.03 | 4.19 | 1.00 | 2.67 | 3.34 | 3.01 |
| High (BH) | 3.26 | 6.97 | 7.05 | 10.0 | 4.45 | 6.97 | 7.05 | 9.17 | 1.00 | 6.48 | 7.15 | 6.82 |
| Extremely High (BEH) | 7.19 | 8.97 | 10.0 | 10.0 | 7.93 | 8.97 | 10.0 | 10.0 | 1.00 | 8.97 | 9.19 | 9.08 |

TABLE SM-XXVI
FOU DATA OF USER 22 FOR "APPLICATION RATINGS" OBTAINED WITH HMA- THE CODEBOOK

| Word | UMF | | | | LMF | | | | | Centroid | | Centroid Center |
|---|---|---|---|---|---|---|---|---|---|---|---|---|
| Very slow (AVS) | 0.00 | 0.00 | 2.02 | 3.84 | 0.00 | 0.00 | 2.02 | 3.17 | 1.00 | 1.32 | 1.52 | 1.42 |
| Slow (AS) | 2.72 | 3.99 | 6.03 | 7.38 | 2.82 | 3.99 | 6.03 | 7.25 | 1.00 | 5.00 | 5.06 | 5.03 |
| Moderate (AM) | 3.83 | 6.00 | 7.01 | 8.54 | 4.76 | 6.00 | 7.01 | 8.27 | 1.00 | 6.23 | 6.59 | 6.41 |
| Fast (AF) | 6.4 | 7.87 | 8.01 | 9.11 | 6.82 | 7.87 | 8.01 | 9.02 | 1.00 | 7.79 | 7.95 | 7.87 |
| Extremely fast (AEF) | 7.01 | 8.99 | 10.0 | 10.0 | 7.88 | 8.99 | 10.0 | 10.0 | 1.00 | 8.92 | 9.18 | 9.05 |

TABLE SM-XXVII
FOU DATA OF USER 22 FOR "TYPE OF APPLICATION" OBTAINED WITH HMA- THE CODEBOOK

| Word | UMF | | | | LMF | | | | | Centroid | | Centroid Center |
|---|---|---|---|---|---|---|---|---|---|---|---|---|
| Absolutely uninteresting (AU) | 0.00 | 0.00 | 1.02 | 2.92 | 0.00 | 0.00 | 1.02 | 2.23 | 1.00 | 0.85 | 1.06 | 0.96 |
| Somewhat interesting (SI) | 0.39 | 4.00 | 4.03 | 6.15 | 1.42 | 4.00 | 4.03 | 5.09 | 1.00 | 3.17 | 3.86 | 3.51 |
| Fairly Interesting (FI) | 2.52 | 5.95 | 6.02 | 8.3 | 3.36 | 5.95 | 6.02 | 7.13 | 1.00 | 5.22 | 5.88 | 5.55 |
| More interesting (MI) | 6.76 | 7.99 | 8.06 | 9.57 | 7.13 | 7.99 | 8.06 | 9.14 | 1.00 | 7.98 | 8.24 | 8.11 |
| Absolutely interesting (AI) | 7.24 | 8.97 | 10.0 | 10.0 | 7.93 | 8.97 | 10.0 | 10.0 | 1.00 | 8.98 | 9.19 | 9.09 |

TABLE SM-XXVIII
FOU DATA OF USER 22 FOR "AMOUNT OF TIME SPENT" OBTAINED WITH HMA- THE CODEBOOK

| Word | UMF | | | | LMF | | | | | Centroid | | Centroid |
|---|---|---|---|---|---|---|---|---|---|---|---|---|
| Very Little (VL) | 0.00 | 0.00 | 2.00 | 3.87 | 0.00 | 0.00 | 2.00 | 3.05 | 1.00 | 1.29 | 1.52 | 1.40 |
| Small (S) | 1.91 | 3.98 | 4.06 | 5.87 | 2.82 | 3.98 | 4.06 | 5.25 | 1.00 | 3.73 | 4.23 | 3.98 |
| Moderate (M) | 3.84 | 6.52 | 6.61 | 8.71 | 5.16 | 6.52 | 6.61 | 7.85 | 1.00 | 6.09 | 6.81 | 6.45 |
| Large (L) | 6.07 | 7.97 | 8.02 | 9.93 | 6.73 | 7.97 | 8.02 | 9.26 | 1.00 | 7.78 | 8.22 | 8.00 |
| Very large (VLA) | 8.19 | 8.98 | 10.0 | 10.0 | 8.38 | 8.98 | 10.0 | 10.0 | 1.00 | 9.27 | 9.33 | 9.30 |

TABLE SM-XXIX
FOU DATA OF USER 22 FOR "SATISFACTION RATING" OBTAINED WITH HMA- THE CODEBOOK

| Word | UMF | | | | LMF | | | | | Centroid | | Centroid Center |
|---|---|---|---|---|---|---|---|---|---|---|---|---|
| Not Satisfied (NS) | 0.00 | 0.00 | 1.09 | 2.96 | 0.00 | 0.00 | 1.09 | 2.16 | 1.00 | 0.85 | 1.09 | 0.97 |
| Somehow Satisfied (SOS) | 1.05 | 3.92 | 4.01 | 5.65 | 1.81 | 3.92 | 4.01 | 5.24 | 1.00 | 3.42 | 3.81 | 3.62 |
| Satisfied (SS) | 3.72 | 4.96 | 5.02 | 6.78 | 3.81 | 4.96 | 5.02 | 6.74 | 1.00 | 5.48 | 5.53 | 5.50 |
| Very Satisfied (VS) | 4.86 | 7.91 | 8.00 | 9.63 | 5.76 | 7.91 | 8.00 | 9.20 | 1.00 | 7.34 | 7.78 | 7.56 |
| Overly Satisfied (OS) | 7.28 | 8.98 | 10.0 | 10.0 | 7.80 | 8.98 | 10.0 | 10.0 | 1.00 | 9.00 | 9.16 | 9.08 |

TABLE SM-XXX
FOU DATA OF USER 22 FOR "LINGUISTIC WEIGHTS" OBTAINED WITH HMA- THE CODEBOOK

| Word | UMF | | | | LMF | | | | | Centroid | | Centroid Center |
|---|---|---|---|---|---|---|---|---|---|---|---|---|
| Unimportant (U) | 0.00 | 0.00 | 2.00 | 2.94 | 0.00 | 0.00 | 2.00 | 2.66 | 1.00 | 1.18 | 1.25 | 1.22 |
| More or less unimportant (MLU) | 0.42 | 1.78 | 2.01 | 3.8 | 0.79 | 1.78 | 2.01 | 2.95 | 1.00 | 1.76 | 2.15 | 1.95 |
| Important (I) | 1.18 | 2.99 | 3.00 | 4.74 | 1.80 | 2.99 | 3.00 | 4.20 | 1.00 | 2.79 | 3.18 | 2.99 |
| More or less important (MLI) | 4.38 | 7.00 | 7.06 | 10.0 | 4.84 | 7.00 | 7.06 | 9.22 | 1.00 | 6.88 | 7.29 | 7.08 |
| Very Important (VI) | 7.39 | 8.99 | 10.0 | 10.0 | 7.8 | 8.99 | 10.0 | 10.0 | 1.00 | 9.03 | 9.16 | 9.09 |

## SM-III. USER FEEDBACK FOR BOTH GAMES IN MULTI-PERSON SYSTEM

Here, we present the data of all the 25 users for both games for multi-person systems.

TABLE SM-XXXI
FEEDBACK OF 25 USERS FOR GAME LEFT 4 DEAD

| User ID | Criteria | F1 | | F2 | | F3 | | F4 | | F5 | | F6 | |
|---|---|---|---|---|---|---|---|---|---|---|---|---|---|
| | Training (T)/ Execution(E) | T | E | T | E | T | E | T | E | T | E | T | E |
| 1 | Battery Life | BM | BM | BM | BM | BM | BM | BM | BM | BL | BH | BM | BH |
| | Application ratings | AM | AM | AM | AM | AM | AM | AS | AM | AS | AF | AM | AF |
| | Type of application | MI | FI | MI | FI | FI | MI | FI | MI | FI | MI | FI | AI |
| | Amount of time spent | M | S | S | S | M | S | S | S | S | S | S | S |
| 2 | Battery Life | BL | BM | BM | BL | BM | BM | BL | BM | BL | BM | BM | BM |
| | Application ratings | AM | AM | AF | AM | AM | AM | AM | AM | AM | AM | AM | AM |
| | Type of application | SI | MI | SI | FI | SI | MI | MI | MI | FI | MI | FI | MI |
| | Amount of time spent | L | M | L | M | L | M | S | L | L | M | M | M |
| 3 | Battery Life | BL | BM | BL | BL | BM | BL | BM | BM | BM | BM | BM | BM |
| | Application ratings | AS | AF | AS | AM | AF | AF | AF | AF | AS | AF | AM | AF |
| | Type of application | MI | FI | MI | FI | MI | FI | MI | FI | MI | FI | MI | FI |
| | Amount of time spent | S | VL | M | VL | L | VL | S | VL | M | VL | S | VL |
| 4 | Battery Life | BM | BH | BM | BH | BM | BH | BM | BM | BH | BH | BH | BH |
| | Application ratings | AS | AS | AVS | AM | AF | AM | AM | AM | AF | AM | AF | AF |
| | Type of application | AI | FI | AI | MI | MI | SI | AI | MI | AI | AI | AI | AI |
| | Amount of time spent | M | M | M | S | L | S | L | M | L | L | L | L |
| 5 | Battery Life | BM | BL | BM | BL | BM | BL | BM | BL | BL | BL | BL | BL |
| | Application ratings | AVS | AS | AS | AS | AS | AS | AS | AM | AM | AM | AM | AM |
| | Type of application | SI | FI | SI | FI | FI | FI | FI | MI | MI | MI | MI | MI |
| | Amount of time spent | VL | S | VL | S | VL | S | S | S | M | S | M | S |
| 6 | Battery Life | BH | BM | BM | BM | BM | BM | BH | BL | BL | BL | BL | BL |
| | Application ratings | AF | AF | AF | AM | AM | AM | AM | AM | AM | AM | AS | AS |
| | Type of application | FI | FI | FI | FI | FI | FI | FI | FI | FI | FI | FI | FI |
| | Amount of time spent | M | M | M | M | M | M | L | S | M | S | M | S |
| 7 | Battery Life | BM | BM | BM | BM | BM | BM | BL | BM | BL | BM | BL | BM |
| | Application ratings | AS | AVS | AS | AVS | AS | AVS | AM | AVS | AM | AS | AS | AM |
| | Type of application | SI | FI | SI | FI | SI | SI | FI | FI | FI | FI | MI | FI |
| | Amount of time spent | M | VL | L | VL | M | VL | L | VL | M | VL | L | VL |
| 8 | Battery Life | BM | BH | BM | BH | BL | BL | BL | BM | BL | BL | BL | BL |
| | Application ratings | AM | AS | AM | AM | AF | AM | AF | AM | AF | AF | AEF | AF |
| | Type of application | FI | MI | FI | MI | MI | MI | FI | MI | SI | MI | FI | MI |
| | Amount of time spent | L | S | M | S | L | S | M | S | M | S | S | S |
| 9 | Battery Life | BM | BM | BM | BM | BM | BL | BM | BL | BM | BM | BM | BM |
| | Application ratings | AF | AF | AF | AM | AM | AF | AM | AF | AF | AF | AF | AF |
| | Type of application | AI | FI | AI | MI | SI | MI | MI | MI | MI | AI | MI | AI |
| | Amount of time spent | L | M | L | M | M | M | M | M | L | M | L | M |
| 10 | Battery Life | BH | BL | BM | BL | BH | BL | BH | BL | BH | BL | BL | BL |
| | Application ratings | AS | AM | AM | AS | AS | AS | AM | AS | AM | AF | AF | AF |
| | Type of application | SI | FI | FI | SI | MI | MI | SI | SI | FI | MI | MI | MI |
| | Amount of time spent | M | L | S | VL | S | M | M | M | S | S | L | VL |
| 11 | Battery Life | BH | BL | BH | BM | BM | BH | BL | BH | BM | BM | BL | BM |
| | Application ratings | AM | AM | AS | AM | AM | AM | AM | AF | AM | AF | AS | AF |
| | Type of application | SI | SI | MI | AI | FI | AI | MI | AI | MI | MI | MI | MI |
| | Amount of time spent | L | L | L | M | L | M | L | M | M | VLA | M | VLA |
| 12 | Battery Life | BH | BM | BM | BH | BM | BM | BM | BM | BL | BH | BL | BM |
| | Application ratings | AS | AVS | AM | AVS | AM | AS | AM | AS | AS | AM | AS | AS |
| | Type of application | SI | AU | SI | SI | FI | FI | FI | SI | SI | FI | SI | SI |
| | Amount of time spent | VL | VL | VL | VL | VL | VL | S | VL | S | VL | S | VL |
| 13 | Battery Life | BM | BM | BM | BM | BM | BH | BH | BM | BM | BH | BM | BM |
| | Application ratings | AS | AS | AS | AS | AS | AM | AS | AM | AS | AM | AM | AM |
| | Type of application | SI | FI | FI | SI | SI | SI | FI | FI | SI | FI | FI | SI |
| | Amount of time spent | L | S | M | S | S | S | M | S | L | S | M | S |
| 14 | Battery Life | BH | BH | BH | BH | BM | BM | BL | BH | BM | BM | BM | BM |
| | Application ratings | AF | AF | AM | AM | AM | AM | AM | AM | AF | AM | AM | AM |
| | Type of application | MI | MI | FI | MI | FI | FI | SI | FI | MI | FI | FI | FI |
| | Amount of time spent | S | VL | S | VL | S | VL | S | VL | M | VL | M | VL |
| 15 | Battery Life | BH | BH | BH | BH | BH | BH | BH | BH | BH | BH | BH | BH |
| | Application ratings | AS | AS | AM | AM | AS | AM | AM | AM | AS | AS | AS | AM |
| | Type of application | MI | MI | MI | MI | MI | MI | MI | MI | MI | MI | FI | MI |
| | Amount of time spent | L | L | L | S | L | S | L | S | L | S | M | S |
| 16 | Battery Life | BL | BL | BL | BL | BL | BL | BL | BL | BL | BL | BM | BM |
| | Application ratings | AVS | AVS | AVS | AVS | AVS | AVS | AS | AS | AS | AS | AS | AS |
| | Type of application | FI | FI | FI | FI | FI | FI | FI | FI | FI | FI | FI | FI |
| | Amount of time spent | VL | M | VL | M | S | M | S | M | VL | M | S | M |
| 17 | Battery Life | BL | BM | BL | BM | BL | BM | BM | BM | BM | BM | BM | BL |
| | Application ratings | AVS | AS | AVS | AS | AS | AS | AS | AS | AVS | AS | AVS | AS |
| | Type of application | FI | FI | FI | FI | FI | FI | FI | FI | FI | FI | FI | FI |
| | Amount of time spent | VL | VL | VL | VL | S | VL | S | VL | S | VL | M | VL |
| 18 | Battery Life | BH | BH | BH | BH | BH | BH | BH | BH | BM | BH | BL | BH |
| | Application ratings | AEF | AF | AF | AF | AF | AF | AF | AF | AF | AF | AM | AF |
| | Type of application | FI | FI | MI | FI | FI | MI | FI | MI | FI | MI | FI | AI |
| | Amount of time spent | S | S | L | S | L | S | M | S | M | S | M | S |
| 19 | Battery Life | BH | BM | BH | BL | BH | BL | BL | BL | BM | BH | BM | BM |
| | Application ratings | AS | AM | AF | AF | AM | AM | AM | AF | AM | AF | AS | AM |
| | Type of application | MI | MI | MI | MI | FI | SI | MI | FI | FI | MI | FI | SI |
| | Amount of time spent | M | S | VLA | S | L | S | M | S | M | S | M | VL |
| 20 | Battery Life | BH | BM | BH | BM | BH | BM | BM | BM | BH | BM | BM | BL |
| | Application ratings | AS | AM | AVS | AS | AF | AVS | AS | AS | AM | AM | AM | AS |
| | Type of application | FI | MI | FI | MI | MI | MI | MI | MI | FI | MI | FI | MI |
| | Amount of time spent | S | VL | M | VL | S | VL | M | VL | S | VL | M | VL |
| 21 | Battery Life | BH | BM | BH | BM | BM | BM | BM | BM | BM | BM | BM | BM |
| | Application ratings | AF | AM | AF | AM | AM | AS | AF | AS | AM | AM | AM | AM |
| | Type of application | FI | FI | MI | FI | SI | FI | MI | FI | FI | SI | FI | FI |
| | Amount of time spent | VL | S | VLA | S | L | S | L | S | M | S | M | S |
| 22 | Battery Life | BM | BH | BM | BH | BH | BH | BH | BM | BM | BM | BM | BM |
| | Application ratings | AM | AF | AM | AEF | AS | AF | AF | AM | AM | AF | AM | AF |
| | Type of application | FI | MI | FI | FI | SI | MI | MI | FI | FI | FI | MI | MI |
| | Amount of time spent | S | VL | M | VL | VL | VL | S | VL | S | VL | M | VL |
| 23 | Battery Life | BM | BM | BM | BM | BH | BM | BH | BM | BM | BM | BM | BM |
| | Application ratings | AF | AF | AM | AF | AF | AM | AF | AM | AF | AM | AF | AM |
| | Type of application | MI | AI | MI | MI | MI | MI | MI | MI | MI | MI | MI | MI |
| | Amount of time spent | M | S | M | S | L | S | L | S | L | S | M | S |
| 24 | Battery Life | BM | BM | BM | BL | BM | BM | BH | BM | BM | BM | BM | BM |
| | Application ratings | AF | AF | AF | AF | AF | AF | AF | AF | AF | AF | AF | AF |
| | Type of application | FI | MI | MI | MI | MI | MI | FI | MI | FI | MI | FI | MI |
| | Amount of time spent | M | VL | S | S | L | S | S | S | S | S | M | S |
| 25 | Battery Life | BM | BH | BH | BH | BH | BH | BM | BM | BH | BH | BH | BH |
| | Application ratings | AM | AM | AF | AM | AM | AF | AM | AM | AM | AF | AS | AM |
| | Type of application | FI | MI | MI | FI | MI | MI | FI | MI | SI | FI | FI | AI |
| | Amount of time spent | M | S | S | L | S | VLA | M | M | L | M | M | S |

TABLE SM-XXXII
FEEDBACK OF 25 USERS FOR GAME AMNESIA –THE DARK DESCENT

| **User ID** | **Criteria** | **F1** | | **F2** | | **F3** | | **F4** | | **F5** | | **F6** | |
|---|---|---|---|---|---|---|---|---|---|---|---|---|---|
| **Training (T)/ Execution(E)** | | **T** | **E** | **T** | **E** | **T** | **E** | **T** | **E** | **T** | **E** | **T** | **E** |
| 1 | Battery Life | BL | BL | BL | BL | BL | BL | BM | BL | BM | BL | BM | BL |
| | Application ratings | AF | AM | AF | AM | AF | AS | AM | AM | AM | AF | AM | AF |
| | Type of application | FI | SI | FI | SI | FI | SI | FI | SI | SI | SI | SI | SI |
| | Amount of time spent | M | S | M | S | S | S | S | S | M | S | L | S |
| 2 | Battery Life | BM | BM | BL | BL | BL | BM | BM | BL | BM | BL | BM | BL |
| | Application ratings | AM | AM | AM | AM | AM | AM | AM | AM | AM | AM | AM | AM |
| | Type of application | AU | SI | AU | SI | AU | AU | SI | SI | AU | SI | SI | SI |
| | Amount of time spent | S | M | S | M | S | M | M | M | M | M | M | M |
| 3 | Battery Life | BH | BH | BL | BM | BM | BM | BM | BM | BL | BM | BM | BL |
| | Application ratings | AVS | AS | AM | AS | AS | AM | AM | AM | AM | AM | AM | AM |
| | Type of application | SI | SI | SI | AU | FI | AU | AU | SI | AU | FI | SI | AI |
| | Amount of time spent | S | S | M | S | S | S | M | S | S | S | S | S |
| 4 | Battery Life | BM | BH | BM | BH | BM | BH | BH | BH | BM | BH | BL | BH |
| | Application ratings | AM | AF | AS | AM | AM | AM | AF | AF | AM | AF | AF | AS |
| | Type of application | FI | MI | FI | SI | MI | FI | AI | MI | AI | MI | AI | FI |
| | Amount of time spent | M | L | M | L | M | L | L | L | M | M | S | S |
| 5 | Battery Life | BM | BL | BL | BL | BL | BL | BL | BL | BL | BL | BL | BL |
| | Application ratings | AM | AM | AM | AM | AM | AM | AS | AM | AS | AM | AS | AM |
| | Type of application | SI | SI | AU | FI | AU | SI | AU | SI | AU | SI | AU | AU |
| | Amount of time spent | M | M | S | M | VL | M | VL | M | VL | S | VL | S |
| 6 | Battery Life | BM | BL | BM | BL | BL | BM | BM | BM | BL | BM | BM | BL |
| | Application ratings | AM | AF | AM | AM | AM | AM | AM | AF | AM | AM | AM | AM |
| | Type of application | SI | SI | SI | SI | SI | SI | SI | SI | SI | SI | SI | SI |
| | Amount of time spent | M | M | M | M | M | M | M | M | M | M | L | M |
| 7 | Battery Life | BM | BL | BM | BM | BL | BM | BM | BL | BM | BM | BM | BM |
| | Application ratings | AM | AM | AM | AM | AM | AM | AM | AM | AM | AM | AM | AM |
| | Type of application | MI | FI | FI | MI | FI | MI | FI | MI | FI | MI | FI | MI |
| | Amount of time spent | S | VL | S | VL | S | VL | M | VL | S | VL | M | VL |
| 8 | Battery Life | BM | BL | BM | BL | BL | BL | BL | BL | BL | BL | BL | BM |
| | Application ratings | AS | AM | AS | AM | AS | AM | AS | AM | AM | AM | AM | AM |
| | Type of application | SI | SI | SI | SI | SI | SI | AU | SI | FI | SI | FI | SI |
| | Amount of time spent | L | M | S | M | S | M | S | M | L | M | L | M |
| 9 | Battery Life | BM | BM | BH | BM | BH | BH | BH | BH | BH | BM | BH | BM |
| | Application ratings | AM | AF | AM | AF | AF | AF | AF | AF | AF | AF | AF | AF |
| | Type of application | FI | FI | FI | FI | FI | MI | MI | MI | MI | MI | MI | FI |
| | Amount of time spent | M | M | M | M | M | M | M | M | M | M | M | M |
| 10 | Battery Life | BH | BM | BL | BL | BM | BL | BL | BL | BL | BL | BL | BL |
| | Application ratings | AF | AM | AF | AM | AF | AF | AF | AM | AM | AF | AM | AF |
| | Type of application | FI | FI | SI | FI | SI | MI | FI | AI | AU | AI | AU | FI |
| | Amount of time spent | L | S | L | M | L | L | M | VL | VLA | VL | VLA | S |
| 11 | Battery Life | BM | BM | BM | BH | BM | BM | BL | BH | BL | BM | BM | BL |
| | Application ratings | AM | AVS | AM | AS | AVS | AM | AM | AM | AS | AM | AM | AF |
| | Type of application | SI | FI | SI | FI | AU | MI | AU | AU | MI | AI | MI | FI |
| | Amount of time spent | M | M | M | L | L | L | M | M | L | M | M | S |
| 12 | Battery Life | BM | BM | BM | BH | BL | BM | BL | BM | BM | BM | BM | BM |
| | Application ratings | AEF | AEF | AEF | AF | AF | AF | AM | AM | AS | AS | AVS | AS |
| | Type of application | MI | MI | MI | MI | MI | FI | FI | MI | SI | MI | AU | MI |
| | Amount of time spent | S | VL | M | VL | VL | VL | S | VL | M | VL | VL | VL |
| 13 | Battery Life | BM | BM | BM | BM | BM | BM | BM | BM | BM | BM | BM | BM |
| | Application ratings | AM | AM | AM | AM | AM | AM | AM | AF | AM | AF | AM | AF |
| | Type of application | SI | SI | SI | SI | SI | FI | FI | FI | MI | FI | FI | FI |
| | Amount of time spent | L | M | S | M | M | M | M | M | M | M | L | M |
| 14 | Battery Life | BM | BM | BH | BH | BH | BM | BL | BM | BM | BH | BL | BM |
| | Application ratings | AS | AS | AM | AM | AS | AM | AS | AM | AS | AM | AM | AS |
| | Type of application | SI | SI | SI | FI | SI | FI | SI | FI | SI | FI | SI | SI |
| | Amount of time spent | VL | S | S | S | S | S | M | S | S | S | M | S |
| 15 | Battery Life | BH | BH | BH | BM | BH | BM | BH | BL | BH | BM | BH | BM |
| | Application ratings | AM | AM | AM | AM | AS | AS | AM | AVS | AM | AVS | AM | AVS |
| | Type of application | FI | SI | FI | SI | SI | SI | SI | SI | SI | SI | SI | SI |
| | Amount of time spent | L | S | L | S | L | S | M | S | M | S | M | S |
| 16 | Battery Life | BL | BL | BL | BL | BL | BL | BL | BL | BL | BL | BL | BL |
| | Application ratings | AS | AS | AS | AS | AS | AS | AS | AS | AM | AM | AM | AS |
| | Type of application | AU | AU | AU | AU | AU | AU | AU | AU | AU | AU | SI | AU |
| | Amount of time spent | S | S | S | S | VL | S | S | S | S | S | VLA | S |
| 17 | Battery Life | BM | BM | BM | BM | BM | BM | BM | BL | BM | BL | BM | BL |
| | Application ratings | AM | AM | AM | AM | AM | AVS | AM | AVS | AM | AVS | AM | AVS |
| | Type of application | SI | SI | SI | SI | SI | SI | SI | SI | SI | SI | SI | AU |
| | Amount of time spent | VL | VL | VL | VL | VL | VL | VL | VL | VL | VL | S | VL |
| 18 | Battery Life | BH | BH | BH | BH | BH | BH | BM | BH | BM | BH | BM | BH |
| | Application ratings | AF | AF | AF | AF | AF | AM | AF | AM | AF | AF | AF | AF |
| | Type of application | MI | MI | AI | MI | AI | FI | FI | FI | AI | FI | AI | AI |
| | Amount of time spent | VL | S | M | S | L | S | M | S | L | S | L | S |
| 19 | Battery Life | BH | BH | BH | BM | BH | BM | BM | BM | BL | BM | BL | BM |
| | Application ratings | AF | AF | AF | AF | AF | AF | AF | AF | AM | AEF | AF | AEF |
| | Type of application | AU | FI | AU | FI | SI | FI | MI | FI | SI | FI | FI | FI |
| | Amount of time spent | S | S | VL | S | M | S | M | M | M | M | M | M |
| 20 | Battery Life | BH | BM | BH | BM | BH | BM | BH | BL | BM | BM | BH | BM |
| | Application ratings | AS | AS | AM | AM | AF | AF | AF | AF | AF | AF | AF | AF |
| | Type of application | FI | MI | FI | MI | FI | MI | FI | MI | MI | MI | MI | MI |
| | Amount of time spent | VL | VL | S | VL | S | VL | S | VL | S | VL | M | VL |
| 21 | Battery Life | BM | BH | BM | BH | BM | BH | BH | BM | BM | BM | BM | BM |
| | Application ratings | AM | AM | AM | AM | AM | AS | AS | AS | AS | AM | AS | AM |
| | Type of application | MI | SI | FI | SI | FI | MI | FI | MI | SI | FI | SI | FI |
| | Amount of time spent | M | S | S | S | M | S | M | S | M | S | L | S |
| 22 | Battery Life | BM | BL | BM | BM | BH | BM | BH | BM | BM | BL | BM | BL |
| | Application ratings | AM | AF | AM | AM | AM | AS | AF | AS | AM | AS | AS | AS |
| | Type of application | FI | MI | FI | MI | FI | FI | MI | SI | FI | SI | FI | FI |
| | Amount of time spent | M | S | M | M | M | M | S | L | M | L | M | L |
| 23 | Battery Life | BM | BM | BM | BM | BM | BM | BM | BM | BM | BM | BM | BM |
| | Application ratings | AF | AF | AF | AF | AF | AF | AF | AF | AF | AF | AF | AF |
| | Type of application | MI | AI | MI | AI | MI | AI | MI | AI | MI | AI | FI | AI |
| | Amount of time spent | M | L | M | L | M | L | M | L | M | L | M | L |
| 24 | Battery Life | BM | BM | BM | BM | BM | BM | BM | BM | BM | BM | BM | BM |
| | Application ratings | AVS | AVS | AVS | AVS | AVS | AVS | AVS | AVS | AS | AM | AM | AM |
| | Type of application | SI | SI | SI | SI | SI | SI | SI | SI | FI | FI | FI | FI |
| | Amount of time spent | S | VL | S | VL | S | VL | S | VL | M | VL | M | VL |
| 25 | Battery Life | BH | BH | BM | BH | BM | BM | BH | BH | BH | BH | BH | BH |
| | Application ratings | AM | AM | AM | AF | AF | AM | AM | AM | AS | AF | AM | AM |
| | Type of application | FI | FI | FI | FI | AU | SI | AI | FI | FI | MI | AI | FI |

| | | | | | | | | | | | | |
|---|---|---|---|---|---|---|---|---|---|---|---|---|
| Amount of time spent | M | M | S | M | S | M | S | S | M | S | L | S |

## SM-IV. User Feedback For ALL the Games in Single-Person Systems

Here, we present the data of all the 25 users for all three games for single-person systems.

TABLE SM-XXXIII
FEEDBACK OF 25 USERS FOR GAME SUBWAY SURFERS AND LINGUISTIC WEIGHTS

| **User ID** | **Criteria** | **F1** | | **F2** | | **F3** | | **F4** | | **F5** | | **F6** | | **Weights** |
|---|---|---|---|---|---|---|---|---|---|---|---|---|---|---|
| **Training (T)/ Execution(E)** | | **T** | **E** | **T** | **E** | **T** | **E** | **T** | **E** | **T** | **E** | **T** | **E** | |
| 1 | Battery Life | BH | BH | BH | BH | BM | BH | BM | BM | BL | BL | BVL | BL | Equal |
| | Application ratings | AVS | AS | AM | AM | AM | AF | AF | AF | AEF | AF | AEF | AF | |
| | Type of application | AU | SI | SI | FI | FI | MI | SI | FI | MI | SI | FI | SI | |
| | Amount of time spent | S | S | M | M | L | M | L | S | VLA | S | M | S | |
| 2 | Battery Life | BH | BEH | BH | BEH | BH | BM | BM | BM | BM | BL | BL | BVL | Equal |
| | Application ratings | AVS | AVS | AS | AM | AM | AM | AM | AM | AF | AF | AF | AF | |
| | Type of application | AU | SI | SI | MI | FI | FI | SI | MI | MI | FI | SI | FI | |
| | Amount of time spent | VL | S | S | M | M | M | M | S | VLA | S | L | S | |
| 3 | Battery Life | BEH | BH | BEH | BH | BH | BM | BH | BL | BM | BL | BM | BVL | Equal |
| | Application ratings | AS | AM | AS | AM | AM | AM | AF | AF | AF | AF | AF | AF | |
| | Type of application | MI | MI | MI | MI | MI | MI | MI | MI | MI | MI | MI | MI | |
| | Amount of time spent | L | M | M | M | S | M | S | M | VL | M | VL | M | |
| 4 | Battery Life | BH | BH | BH | BH | BH | BH | BH | BM | BM | BM | BM | BL | Equal |
| | Application ratings | AS | AVS | AS | AS | AM | AM | AM | AM | AF | AM | AF | AM | |
| | Type of application | SI | AU | SI | AU | SI | SI | FI | MI | MI | FI | MI | FI | |
| | Amount of time spent | VL | VL | VL | S | S | M | M | M | M | L | L | L | |
| 5 | Battery Life | BH | BH | BH | BH | BM | BH | BM | BM | BL | BM | BL | BM | Equal |
| | Application ratings | AS | AVS | AS | AS | AM | AM | AF | AM | AF | AF | AEF | AF | |
| | Type of application | SI | FI | FI | FI | FI | FI | FI | MI | MI | FI | MI | FI | |
| | Amount of time spent | M | S | M | S | M | S | M | M | L | M | L | M | |
| 6 | Battery Life | BH | BH | BH | BH | BM | BH | BM | BM | BL | BL | BVL | BL | Equal |
| | Application ratings | AVS | AS | AM | AM | AM | AF | AF | AF | AEF | AF | AEF | AF | |
| | Type of application | AU | SI | SI | FI | FI | MI | SI | FI | MI | SI | FI | SI | |
| | Amount of time spent | S | S | M | M | L | M | L | S | VLA | S | M | S | |
| 7 | Battery Life | BH | BH | BH | BH | BH | BH | BH | BM | BM | BM | BM | BL | Equal |
| | Application ratings | AS | AVS | AS | AS | AM | AM | AM | AM | AF | AM | AF | AM | |
| | Type of application | SI | AU | SI | AU | SI | SI | FI | MI | MI | FI | MI | FI | |
| | Amount of time spent | VL | VL | VL | S | S | M | M | M | M | L | L | L | |
| 8 | Battery Life | BH | BEH | BH | BEH | BH | BM | BM | BM | BM | BL | BL | BVL | Equal |
| | Application ratings | AVS | AVS | AS | AM | AM | AM | AM | AM | AF | AF | AF | AF | |
| | Type of application | AU | SI | SI | MI | FI | FI | SI | MI | MI | FI | SI | FI | |
| | Amount of time spent | VL | S | S | M | M | M | M | S | VLA | S | L | S | |
| 9 | Battery Life | BH | BH | BH | BH | BM | BH | BM | BM | BL | BM | BL | BM | Equal |
| | Application ratings | AS | AVS | AS | AS | AM | AM | AF | AM | AF | AF | AEF | AF | |
| | Type of application | SI | FI | FI | FI | FI | FI | FI | MI | MI | FI | MI | FI | |
| | Amount of time spent | M | S | M | S | M | S | M | M | L | M | L | M | |
| 10 | Battery Life | BEH | BH | BEH | BH | BH | BM | BH | BL | BM | BL | BM | BVL | Equal |
| | Application ratings | AS | AM | AS | AM | AM | AM | AF | AF | AF | AF | AF | AF | |
| | Type of application | MI | MI | MI | MI | MI | MI | MI | MI | MI | MI | MI | MI | |
| | Amount of time spent | L | M | M | M | S | M | S | M | VL | M | VL | M | |
| 11 | Battery Life | BEH | BH | BEH | BH | BH | BM | BH | BL | BM | BL | BM | BVL | Equal |
| | Application ratings | AS | AM | AS | AM | AM | AM | AF | AF | AF | AF | AF | AF | |
| | Type of application | MI | MI | MI | MI | MI | MI | MI | MI | MI | MI | MI | MI | |
| | Amount of time spent | L | M | M | M | S | M | S | M | VL | M | VL | M | |
| 12 | Battery Life | BH | BH | BH | BH | BH | BH | BH | BM | BM | BM | BM | BL | Equal |
| | Application ratings | AS | AVS | AS | AS | AM | AM | AM | AM | AF | AM | AF | AM | |
| | Type of application | SI | AU | SI | AU | SI | SI | FI | MI | MI | FI | MI | FI | |
| | Amount of time spent | VL | VL | VL | S | S | M | M | M | M | L | L | L | |
| 13 | Battery Life | BH | BH | BH | BH | BM | BH | BM | BM | BL | BL | BVL | BL | Equal |
| | Application ratings | AVS | AS | AM | AM | AM | AF | AF | AF | AEF | AF | AEF | AF | |
| | Type of application | AU | SI | SI | FI | FI | MI | SI | FI | MI | SI | FI | SI | |
| | Amount of time spent | S | S | M | M | L | M | L | S | VLA | S | M | S | |
| 14 | Battery Life | BH | BH | BH | BH | BM | BH | BM | BM | BL | BL | BVL | BL | VI |
| | Application ratings | AVS | AS | AM | AM | AM | AF | AF | AF | AEF | AF | AEF | AF | VI |
| | Type of application | AU | SI | SI | FI | FI | MI | SI | FI | MI | SI | FI | SI | I |
| | Amount of time spent | S | S | M | M | L | M | L | S | VLA | S | M | S | I |
| 15 | Battery Life | BH | BEH | BH | BEH | BH | BM | BM | BM | BM | BL | BL | BVL | VI |
| | Application ratings | AVS | AVS | AS | AM | AM | AM | AM | AM | AF | AF | AF | AF | VI |
| | Type of application | AU | SI | SI | MI | FI | FI | SI | MI | MI | FI | SI | FI | MLI |
| | Amount of time spent | VL | S | S | M | M | M | M | S | VLA | S | L | S | I |
| 16 | Battery Life | BH | BH | BH | BH | BM | BH | BM | BM | BL | BM | BL | BM | I |
| | Application ratings | AS | AVS | AS | AS | AM | AM | AF | AM | AF | AF | AEF | AF | I |
| | Type of application | SI | FI | FI | FI | FI | FI | FI | MI | MI | FI | MI | FI | U |
| | Amount of time spent | M | S | M | S | M | S | M | M | L | M | L | M | U |
| 17 | Battery Life | BH | BH | BH | BH | BH | BH | BH | BM | BM | BM | BM | BL | VI |
| | Application ratings | AS | AVS | AS | AS | AM | AM | AM | AM | AF | AM | AF | AM | MLI |
| | Type of application | SI | AU | SI | AU | SI | SI | FI | MI | MI | FI | MI | FI | I |
| | Amount of time spent | VL | VL | VL | S | S | M | M | M | M | L | L | L | MLU |
| 18 | Battery Life | BEH | BH | BEH | BH | BH | BM | BH | BL | BM | BL | BM | BVL | MLI |
| | Application ratings | AS | AM | AS | AM | AM | AM | AF | AF | AF | AF | AF | AF | I |
| | Type of application | MI | MI | MI | MI | MI | MI | MI | MI | MI | MI | MI | MI | MLU |
| | Amount of time spent | L | M | M | M | S | M | S | M | VL | M | VL | M | U |
| 19 | Battery Life | BH | BH | BH | BH | BM | BH | BM | BM | BL | BL | BVL | BL | VI |
| | Application ratings | AVS | AS | AM | AM | AM | AF | AF | AF | AEF | AF | AEF | AF | VI |
| | Type of application | AU | SI | SI | FI | FI | MI | SI | FI | MI | SI | FI | SI | I |
| | Amount of time spent | S | S | M | M | L | M | L | S | VLA | S | M | S | I |
| 20 | Battery Life | BH | BH | BH | BH | BH | BH | BH | BM | BM | BM | BM | BL | VI |
| | Application ratings | AS | AVS | AS | AS | AM | AM | AM | AM | AF | AM | AF | AM | MLI |
| | Type of application | SI | AU | SI | AU | SI | SI | FI | MI | MI | FI | MI | FI | I |
| | Amount of time spent | VL | VL | VL | S | S | M | M | M | M | L | L | L | MLU |
| 21 | Battery Life | BH | BEH | BH | BEH | BH | BM | BM | BM | BM | BL | BL | BVL | VI |
| | Application ratings | AVS | AVS | AS | AM | AM | AM | AM | AM | AF | AF | AF | AF | MLI |
| | Type of application | AU | SI | SI | MI | FI | FI | SI | MI | MI | FI | SI | FI | I |
| | Amount of time spent | VL | S | S | M | M | M | M | S | VLA | S | L | S | I |
| 22 | Battery Life | BH | BH | BH | BH | BM | BH | BM | BM | BL | BM | BL | BM | MLI |
| | Application ratings | AS | AVS | AS | AS | AM | AM | AF | AM | AF | AF | AEF | AF | I |
| | Type of application | SI | FI | FI | FI | FI | FI | FI | MI | MI | FI | MI | FI | U |
| | Amount of time spent | M | S | M | S | M | S | M | M | L | M | L | M | U |
| 23 | Battery Life | BEH | BH | BEH | BH | BH | BM | BH | BL | BM | BL | BM | BVL | MLI |
| | Application ratings | AS | AM | AS | AM | AM | AM | AF | AF | AF | AF | AF | AF | I |
| | Type of application | MI | MI | MI | MI | MI | MI | MI | MI | MI | MI | MI | MI | MLU |
| | Amount of time spent | L | M | M | M | S | M | S | M | VL | M | VL | M | U |
| 24 | Battery Life | BH | BH | BH | BH | BM | BH | BM | BM | BL | BM | BL | BM | MLI |
| | Application ratings | AS | AVS | AS | AS | AM | AM | AF | AM | AF | AF | AEF | AF | I |
| | Type of application | SI | FI | FI | FI | FI | FI | FI | MI | MI | FI | MI | FI | MLU |
| | Amount of time spent | M | S | M | S | M | S | M | M | L | M | L | M | MLU |
| 25 | Battery Life | BH | BEH | BH | BEH | BH | BM | BM | BM | BM | BL | BL | BVL | VI |
| | Application ratings | AVS | AVS | AS | AM | AM | AM | AM | AM | AF | AF | AF | AF | MLI |
| | Type of application | AU | SI | SI | MI | FI | FI | SI | MI | MI | FI | SI | FI | I |
| | Amount of time spent | VL | S | S | M | M | M | M | S | VLA | S | L | S | MLU |

TABLE SM-XXXIV

FEEDBACK OF 25 USERS FOR GAME ASPHALT 8: AIRBORNE AND LINGUISTIC WEIGHTS

| User ID | Criteria | F1 | | F2 | | F3 | | F4 | | F5 | | F6 | | Weights |
|---|---|---|---|---|---|---|---|---|---|---|---|---|---|---|
| | Training (T)/ Execution(E) | T | E | T | E | T | E | T | E | T | E | T | E | |
| 1 | Battery Life | BH | BH | BM | BM | BM | BL | BL | BVL | BL | BVL | BL | BVL | Equal |
| | Application ratings | AS | AVS | AS | AM | AM | AF | AF | AEF | AF | AEF | AEF | AEF | |
| | Type of application | FI | AU | FI | SI | FI | FI | MI | MI | MI | MI | MI | MI | |
| | Amount of time spent | VL | VL | S | S | M | M | L | L | L | L | L | L | |
| 2 | Battery Life | BH | BH | BH | BH | BM | BH | BM | BH | BM | BL | BL | BVL | Equal |
| | Application ratings | AS | AVS | AS | AS | AM | AS | AM | AM | AEF | AEF | AEF | AF | |
| | Type of application | FI | SI | FI | FI | MI | FI | FI | MI | AI | AI | AI | AI | |
| | Amount of time spent | VL | S | S | M | M | M | M | L | L | L | L | L | |
| 3 | Battery Life | BM | BM | BM | BM | BM | BM | BL | BM | BL | BL | BL | BL | Equal |
| | Application ratings | AS | AVS | AM | AM | AM | AM | AM | AF | AEF | AF | AEF | AF | |
| | Type of application | SI | SI | SI | FI | MI | FI | MI | FI | FI | FI | MI | FI | |
| | Amount of time spent | M | S | S | M | L | M | L | M | M | M | L | M | |
| 4 | Battery Life | BH | BH | BH | BH | BH | BH | BM | BH | BM | BL | BM | BVL | Equal |
| | Application ratings | AS | AVS | AS | AS | AS | AM | AM | AM | AF | AM | AEF | AF | |
| | Type of application | AU | AU | AU | AU | SI | SI | FI | FI | FI | SI | MI | SI | |
| | Amount of time spent | M | S | M | S | M | S | M | S | L | S | L | S | |
| 5 | Battery Life | BH | BH | BH | BH | BM | BH | BL | BM | BL | BL | BVL | BL | Equal |
| | Application ratings | AF | AF | AF | AF | AS | AS | AM | AM | AM | AM | AM | AM | |
| | Type of application | MI | MI | MI | MI | FI | FI | FI | FI | FI | FI | FI | FI | |
| | Amount of time spent | L | M | L | M | M | M | M | M | M | M | M | M | |
| 6 | Battery Life | BH | BH | BM | BM | BM | BL | BL | BVL | BL | BVL | BL | BVL | Equal |
| | Application ratings | AS | AVS | AS | AM | AM | AF | AF | AEF | AF | AEF | AEF | AEF | |
| | Type of application | FI | AU | FI | SI | FI | FI | MI | MI | MI | MI | MI | MI | |
| | Amount of time spent | VL | VL | S | S | M | M | L | L | L | L | L | L | |
| 7 | Battery Life | BH | BH | BH | BH | BH | BH | BM | BH | BM | BL | BM | BVL | Equal |
| | Application ratings | AS | AVS | AS | AS | AS | AM | AM | AM | AF | AM | AEF | AF | |
| | Type of application | AU | AU | AU | AU | SI | SI | FI | FI | FI | SI | MI | SI | |
| | Amount of time spent | M | S | M | S | M | S | M | S | L | S | L | S | |
| 8 | Battery Life | BH | BH | BH | BH | BM | BH | BM | BH | BM | BL | BL | BVL | Equal |
| | Application ratings | AS | AVS | AS | AS | AM | AS | AM | AM | AEF | AEF | AEF | AF | |
| | Type of application | FI | SI | FI | FI | MI | FI | FI | MI | AI | AI | AI | AI | |
| | Amount of time spent | VL | S | S | M | M | M | M | L | L | L | L | L | |
| 9 | Battery Life | BH | BH | BH | BH | BM | BH | BL | BM | BL | BL | BVL | BL | Equal |
| | Application ratings | AF | AF | AF | AF | AS | AS | AM | AM | AM | AM | AM | AM | |
| | Type of application | MI | MI | MI | MI | FI | FI | FI | FI | FI | FI | FI | FI | |
| | Amount of time spent | L | M | L | M | M | M | M | M | M | M | M | M | |
| 10 | Battery Life | BM | BM | BM | BM | BM | BM | BL | BM | BL | BL | BL | BL | Equal |
| | Application ratings | AS | AVS | AM | AM | AM | AM | AM | AF | AEF | AF | AEF | AF | |
| | Type of application | SI | SI | SI | FI | MI | FI | MI | FI | FI | FI | MI | FI | |
| | Amount of time spent | M | S | S | M | L | M | L | M | M | M | L | M | |
| 11 | Battery Life | BM | BM | BM | BM | BM | BM | BL | BM | BL | BL | BL | BL | Equal |
| | Application ratings | AS | AVS | AM | AM | AM | AM | AM | AF | AEF | AF | AEF | AF | |
| | Type of application | SI | SI | SI | FI | MI | FI | MI | FI | FI | FI | MI | FI | |
| | Amount of time spent | M | S | S | M | L | M | L | M | M | M | L | M | |
| 12 | Battery Life | BH | BH | BH | BH | BH | BH | BM | BH | BM | BL | BM | BVL | Equal |
| | Application ratings | AS | AVS | AS | AS | AS | AM | AM | AM | AF | AM | AEF | AF | |
| | Type of application | AU | AU | AU | AU | SI | SI | FI | FI | FI | SI | MI | SI | |
| | Amount of time spent | M | S | M | S | M | S | M | S | L | S | L | S | |
| 13 | Battery Life | BH | BH | BM | BM | BM | BL | BL | BVL | BL | BVL | BL | BVL | Equal |
| | Application ratings | AS | AVS | AS | AM | AM | AF | AF | AEF | AF | AEF | AEF | AEF | |
| | Type of application | FI | AU | FI | SI | FI | FI | MI | MI | MI | MI | MI | MI | |
| | Amount of time spent | VL | VL | S | S | M | M | L | L | L | L | L | L | |
| 14 | Battery Life | BH | BH | BM | BM | BM | BL | BL | BVL | BL | BVL | BL | BVL | VI |
| | Application ratings | AS | AVS | AS | AM | AM | AF | AF | AEF | AF | AEF | AEF | AEF | VI |
| | Type of application | FI | AU | FI | SI | FI | FI | MI | MI | MI | MI | MI | MI | I |
| | Amount of time spent | VL | VL | S | S | M | M | L | L | L | L | L | L | I |
| 15 | Battery Life | BH | BH | BH | BH | BM | BH | BM | BH | BM | BL | BL | BVL | VI |
| | Application ratings | AS | AVS | AS | AS | AM | AS | AM | AM | AEF | AEF | AEF | AF | VI |
| | Type of application | FI | SI | FI | FI | MI | FI | FI | MI | AI | AI | AI | AI | MLI |
| | Amount of time spent | VL | S | S | M | M | M | M | L | L | L | L | L | I |
| 16 | Battery Life | BH | BH | BH | BH | BM | BH | BL | BM | BL | BL | BVL | BL | I |
| | Application ratings | AF | AF | AF | AF | AS | AS | AM | AM | AM | AM | AM | AM | I |
| | Type of application | MI | MI | MI | MI | FI | FI | FI | FI | FI | FI | FI | FI | U |
| | Amount of time spent | L | M | L | M | M | M | M | M | M | M | M | M | U |
| 17 | Battery Life | BH | BH | BH | BH | BH | BH | BM | BH | BM | BL | BM | BVL | VI |
| | Application ratings | AS | AVS | AS | AS | AS | AM | AM | AM | AF | AM | AEF | AF | MLI |
| | Type of application | AU | AU | AU | AU | SI | SI | FI | FI | FI | SI | MI | SI | I |
| | Amount of time spent | M | S | M | S | M | S | M | S | L | S | L | S | MLU |
| 18 | Battery Life | BM | BM | BM | BM | BM | BM | BL | BM | BL | BL | BL | BL | MLI |
| | Application ratings | AS | AVS | AM | AM | AM | AM | AM | AF | AEF | AF | AEF | AF | I |
| | Type of application | SI | SI | SI | FI | MI | FI | MI | FI | FI | FI | MI | FI | MLU |
| | Amount of time spent | M | S | S | M | L | M | L | M | M | M | L | M | U |
| 19 | Battery Life | BH | BH | BM | BM | BM | BL | BL | BVL | BL | BVL | BL | BVL | VI |
| | Application ratings | AS | AVS | AS | AM | AM | AF | AF | AEF | AF | AEF | AEF | AEF | VI |
| | Type of application | FI | AU | FI | SI | FI | FI | MI | MI | MI | MI | MI | MI | I |
| | Amount of time spent | VL | VL | S | S | M | M | L | L | L | L | L | L | I |
| 20 | Battery Life | BH | BH | BH | BH | BH | BH | BM | BH | BM | BL | BM | BVL | VI |
| | Application ratings | AS | AVS | AS | AS | AS | AM | AM | AM | AF | AM | AEF | AF | MLI |
| | Type of application | AU | AU | AU | AU | SI | SI | FI | FI | FI | SI | MI | SI | I |
| | Amount of time spent | M | S | M | S | M | S | M | S | L | S | L | S | MLU |
| 21 | Battery Life | BH | BH | BH | BH | BM | BH | BM | BH | BM | BL | BL | BVL | VI |
| | Application ratings | AS | AVS | AS | AS | AM | AS | AM | AM | AEF | AEF | AEF | AF | MLI |
| | Type of application | FI | SI | FI | FI | MI | FI | FI | MI | AI | AI | AI | AI | I |
| | Amount of time spent | VL | S | S | M | M | M | M | L | L | L | L | L | I |
| 22 | Battery Life | BH | BH | BH | BH | BM | BH | BL | BM | BL | BL | BVL | BL | MLI |
| | Application ratings | AF | AF | AF | AF | AS | AS | AM | AM | AM | AM | AM | AM | I |
| | Type of application | MI | MI | MI | MI | FI | FI | FI | FI | FI | FI | FI | FI | U |
| | Amount of time spent | L | M | L | M | M | M | M | M | M | M | M | M | U |
| 23 | Battery Life | BM | BM | BM | BM | BM | BM | BL | BM | BL | BL | BL | BL | MLI |
| | Application ratings | AS | AVS | AM | AM | AM | AM | AM | AF | AEF | AF | AEF | AF | I |
| | Type of application | SI | SI | SI | FI | MI | FI | MI | FI | FI | FI | MI | FI | MLU |
| | Amount of time spent | M | S | S | M | L | M | L | M | M | M | L | M | U |
| 24 | Battery Life | BH | BH | BH | BH | BM | BH | BL | BM | BL | BL | BVL | BL | MLI |
| | Application ratings | AF | AF | AF | AF | AS | AS | AM | AM | AM | AM | AM | AM | I |
| | Type of application | MI | MI | MI | MI | FI | FI | FI | FI | FI | FI | FI | FI | MLU |
| | Amount of time spent | L | M | L | M | M | M | M | M | M | M | M | M | MLU |
| 25 | Battery Life | BH | BH | BH | BH | BM | BH | BM | BH | BM | BL | BL | BVL | VI |
| | Application ratings | AS | AVS | AS | AS | AM | AS | AM | AM | AEF | AEF | AEF | AF | MLI |
| | Type of application | FI | SI | FI | FI | MI | FI | FI | MI | AI | AI | AI | AI | I |
| | Amount of time spent | VL | S | S | M | M | M | M | L | L | L | L | L | MLU |

TABLE SM-XXXV
FEEDBACK OF 25 USERS FOR GAME FRUIT NINJA AND LINGUISTIC WEIGHTS

| **User ID** | **Criteria** | **F1** | | **F2** | | **F3** | | **F4** | | **F5** | | **F6** | | **Weights** |
|---|---|---|---|---|---|---|---|---|---|---|---|---|---|---|
| **Training (T)/ Execution(E)** | | **T** | **E** | **T** | **E** | **T** | **E** | **T** | **E** | **T** | **E** | **T** | **E** | |
| 1 | Battery Life | BEH | BH | BM | BM | BL | BL | BL | BL | BL | BVL | BL | BVL | Equal |
| | Application ratings | AVS | AVS | AS | AM | AF | AF | AF | AF | AF | AF | AF | AF | |
| | Type of application | AU | AU | FI | FI | MI | MI | SI | MI | SI | MI | MI | MI | |
| | Amount of time spent | VL | VL | M | M | L | L | M | M | L | L | L | L | |
| 2 | Battery Life | BH | BEH | BH | BH | BH | BM | BM | BL | BL | BL | BL | BL | Equal |
| | Application ratings | AS | AVS | AS | AS | AM | AM | AF | AF | AEF | AEF | AEF | AEF | |
| | Type of application | SI | AU | SI | SI | FI | FI | MI | MI | MI | MI | MI | MI | |
| | Amount of time spent | M | S | M | S | M | S | L | S | L | S | VLA | S | |
| 3 | Battery Life | BH | BH | BH | BH | BH | BH | BH | BH | BM | BM | BM | BM | Equal |
| | Application ratings | AF | AF | AF | AF | AF | AF | AF | AF | AEF | AEF | AEF | AEF | |
| | Type of application | FI | FI | MI | MI | MI | MI | MI | MI | MI | MI | MI | MI | |
| | Amount of time spent | S | M | M | L | L | L | L | L | VLA | L | VLA | L | |
| 4 | Battery Life | BEH | BH | BM | BM | BL | BL | BL | BL | BL | BVL | BL | BVL | Equal |
| | Application ratings | AVS | AVS | AS | AM | AF | AF | AF | AF | AEF | AF | AEF | AF | |
| | Type of application | AU | AU | SI | SI | MI | MI | MI | MI | MI | MI | MI | MI | |
| | Amount of time spent | VL | S | M | S | L | S | L | S | L | S | L | S | |
| 5 | Battery Life | BEH | BH | BM | BM | BL | BL | BL | BL | BL | BVL | BL | BVL | Equal |
| | Application ratings | AVS | AVS | AS | AM | AF | AF | AF | AF | AF | AF | AF | AF | |
| | Type of application | AU | AU | FI | FI | MI | MI | SI | MI | SI | MI | MI | MI | |
| | Amount of time spent | VL | VL | M | M | L | L | M | M | L | L | L | L | |
| 6 | Battery Life | BEH | BH | BM | BM | BL | BL | BL | BL | BL | BVL | BL | BVL | Equal |
| | Application ratings | AVS | AVS | AS | AM | AF | AF | AF | AF | AF | AF | AF | AF | |
| | Type of application | AU | AU | FI | FI | MI | MI | SI | MI | SI | MI | MI | MI | |
| | Amount of time spent | VL | VL | M | M | L | L | M | M | L | L | L | L | |
| 7 | Battery Life | BEH | BH | BM | BM | BL | BL | BL | BL | BL | BVL | BL | BVL | Equal |
| | Application ratings | AVS | AVS | AS | AM | AF | AF | AF | AF | AEF | AF | AEF | AF | |
| | Type of application | AU | AU | SI | SI | MI | MI | MI | MI | MI | MI | MI | MI | |
| | Amount of time spent | VL | VL | S | S | L | L | L | L | L | L | L | L | |
| 8 | Battery Life | BH | BEH | BH | BH | BH | BM | BM | BL | BL | BL | BL | BL | Equal |
| | Application ratings | AS | AVS | AS | AS | AM | AM | AF | AF | AEF | AEF | AEF | AEF | |
| | Type of application | SI | AU | SI | SI | FI | FI | MI | MI | MI | MI | MI | MI | |
| | Amount of time spent | M | S | M | S | M | S | L | S | L | S | VLA | S | |
| 9 | Battery Life | BH | BEH | BH | BH | BH | BM | BM | BL | BL | BL | BL | BL | Equal |
| | Application ratings | AS | AVS | AS | AS | AM | AM | AF | AF | AEF | AEF | AEF | AEF | |
| | Type of application | SI | AU | SI | SI | FI | FI | MI | MI | MI | MI | MI | MI | |
| | Amount of time spent | M | S | M | S | M | S | L | S | L | S | VLA | S | |
| | Battery Life | BH | BH | BH | BH | BH | BH | BH | BH | BM | BM | BM | BM | |
| 10 | Application ratings | AF | AF | AF | AF | AF | AF | AF | AF | AEF | AEF | AEF | AEF | Equal |
| | Type of application | FI | FI | MI | MI | MI | MI | MI | MI | MI | MI | MI | MI | |
| | Amount of time spent | S | M | M | L | L | L | L | L | VLA | L | VLA | L | |
| | Battery Life | BH | BH | BH | BH | BH | BH | BH | BH | BM | BM | BM | BM | |
| 11 | Application ratings | AF | AF | AF | AF | AF | AF | AF | AF | AEF | AEF | AEF | AEF | Equal |
| | Type of application | FI | FI | MI | MI | MI | MI | MI | MI | MI | MI | MI | MI | |
| | Amount of time spent | S | M | M | L | L | L | L | L | VLA | L | VLA | L | |
| | Battery Life | BEH | BH | BM | BM | BL | BL | BL | BL | BL | BVL | BL | BVL | |
| 12 | Application ratings | AVS | AVS | AS | AM | AF | AF | AF | AF | AEF | AF | AEF | AF | Equal |
| | Type of application | AU | AU | SI | SI | MI | MI | MI | MI | MI | MI | MI | MI | |
| | Amount of time spent | VL | VL | S | S | L | L | L | L | L | L | L | L | |
| | Battery Life | BEH | BH | BM | BM | BL | BL | BL | BL | BL | BVL | BL | BVL | |
| 13 | Application ratings | AVS | AVS | AS | AM | AF | AF | AF | AF | AF | AF | AF | AF | Equal |
| | Type of application | AU | AU | FI | FI | MI | MI | SI | MI | SI | MI | MI | MI | |
| | Amount of time spent | VL | VL | M | M | L | L | M | M | L | L | L | L | |
| | Battery Life | BEH | BH | BM | BM | BL | BL | BL | BL | BL | BVL | BL | BVL | VI |
| 14 | Application ratings | AVS | AVS | AS | AM | AF | AF | AF | AF | AF | AF | AF | AF | VI |
| | Type of application | AU | AU | FI | FI | MI | MI | SI | MI | SI | MI | MI | MI | I |
| | Amount of time spent | VL | VL | M | M | L | L | M | M | L | L | L | L | I |
| | Battery Life | BH | BEH | BH | BH | BH | BM | BM | BL | BL | BL | BL | BL | VI |
| 15 | Application ratings | AS | AVS | AS | AS | AM | AM | AF | AF | AEF | AEF | AEF | AEF | VI |
| | Type of application | SI | AU | SI | SI | FI | FI | MI | MI | MI | MI | MI | MI | MLI |
| | Amount of time spent | M | S | M | S | M | S | L | S | L | S | VLA | S | I |
| | Battery Life | BH | BH | BH | BH | BH | BH | BH | BH | BM | BM | BM | BM | I |
| 16 | Application ratings | AF | AF | AF | AF | AF | AF | AF | AF | AEF | AEF | AEF | AEF | I |
| | Type of application | FI | FI | MI | MI | MI | MI | MI | MI | MI | MI | MI | MI | U |
| | Amount of time spent | S | M | M | L | L | L | L | L | VLA | L | VLA | L | U |
| | Battery Life | BEH | BH | BM | BM | BL | BL | BL | BL | BL | BVL | BL | BVL | VI |
| 17 | Application ratings | AVS | AVS | AS | AM | AF | AF | AF | AF | AEF | AF | AEF | AF | MLI |
| | Type of application | AU | AU | SI | SI | MI | MI | MI | MI | MI | MI | MI | MI | I |
| | Amount of time spent | VL | VL | S | S | L | L | L | L | L | L | L | L | MLU |
| | Battery Life | BH | BH | BH | BH | BH | BH | BH | BH | BM | BM | BM | BM | MLI |
| 18 | Application ratings | AF | AF | AF | AF | AF | AF | AF | AF | AEF | AEF | AEF | AEF | I |
| | Type of application | FI | FI | MI | MI | MI | MI | MI | MI | MI | MI | MI | MI | MLU |
| | Amount of time spent | S | M | M | L | L | L | L | L | VLA | L | VLA | L | U |
| | Battery Life | BEH | BH | BM | BM | BL | BL | BL | BL | BL | BVL | BL | BVL | VI |
| 19 | Application ratings | AVS | AVS | AS | AM | AF | AF | AF | AF | AF | AF | AF | AF | VI |
| | Type of application | AU | AU | FI | FI | MI | MI | SI | MI | SI | MI | MI | MI | I |
| | Amount of time spent | VL | VL | M | M | L | L | M | M | L | L | L | L | I |
| | Battery Life | BEH | BH | BM | BM | BL | BL | BL | BL | BL | BVL | BL | BVL | VI |
| 20 | Application ratings | AVS | AVS | AS | AM | AF | AF | AF | AF | AEF | AF | AEF | AF | MLI |
| | Type of application | AU | AU | SI | SI | MI | MI | MI | MI | MI | MI | MI | MI | I |
| | Amount of time spent | VL | VL | S | S | L | L | L | L | L | L | L | L | MLU |
| | Battery Life | BH | BEH | BH | BH | BH | BM | BM | BL | BL | BL | BL | BL | VI |
| 21 | Application ratings | AS | AVS | AS | AS | AM | AM | AF | AF | AEF | AEF | AEF | AEF | MLI |
| | Type of application | SI | AU | SI | SI | FI | FI | MI | MI | MI | MI | MI | MI | I |
| | Amount of time spent | M | S | M | S | M | S | L | S | L | S | VLA | S | I |
| | Battery Life | BEH | BH | BM | BM | BL | BL | BL | BL | BL | BVL | BL | BVL | MLI |
| 22 | Application ratings | AVS | AVS | AS | AM | AF | AF | AF | AF | AEF | AF | AEF | AF | I |
| | Type of application | AU | AU | SI | SI | MI | MI | MI | MI | MI | MI | MI | MI | U |
| | Amount of time spent | VL | S | M | S | L | S | L | S | L | S | L | S | U |
| | Battery Life | BH | BH | BH | BH | BH | BH | BH | BH | BM | BM | BM | BM | MLI |
| 23 | Application ratings | AF | AF | AF | AF | AF | AF | AF | AF | AEF | AEF | AEF | AEF | I |
| | Type of application | FI | FI | MI | MI | MI | MI | MI | MI | MI | MI | MI | MI | MLU |
| | Amount of time spent | S | M | M | L | L | L | L | L | VLA | L | VLA | L | U |
| | Battery Life | BH | BH | BH | BH | BM | BH | BL | BM | BL | BL | BVL | BL | MLI |
| 24 | Application ratings | AF | AF | AF | AF | AS | AS | AM | AM | AM | AM | AM | AM | I |
| | Type of application | MI | MI | MI | MI | FI | FI | FI | FI | FI | FI | FI | FI | MLU |
| | Amount of time spent | L | M | L | M | M | M | M | M | M | M | M | M | MLU |
| | Battery Life | BH | BEH | BH | BH | BH | BM | BM | BL | BL | BL | BL | BL | VI |
| 25 | Application ratings | AS | AVS | AS | AS | AM | AM | AF | AF | AEF | AEF | AEF | AEF | MLI |
| | Type of application | SI | AU | SI | SI | FI | FI | MI | MI | MI | MI | MI | MI | I |

| Amount of time spent | M | S | M | S | M | S | L | S | L | S | VLA | S | MLU |
|---|---|---|---|---|---|---|---|---|---|---|---|---|---|